\documentclass{article}

\usepackage{arxiv}

\usepackage[utf8]{inputenc} 
\usepackage[T1]{fontenc}    
\usepackage{lmodern}        
\usepackage{hyperref}       
\usepackage{url}            
\usepackage{booktabs}       
\usepackage{amsfonts}       
\usepackage{nicefrac}       
\usepackage{microtype}      
\usepackage{graphicx}
\usepackage{authblk}

\title{Comparison of static and dynamic random forests models for EHR data in the presence of competing risks: predicting central line-associated bloodstream infection}

\author[1]{\small Elena ALBU}
\author[1]{\small Shan GAO}
\author[2]{\small Pieter STIJNEN}
\author[3]{\small Frank RADEMAKERS}
\author[4]{\small Christel JANSSENS}
\author[1, 5]{\small Veerle COSSEY}
\author[6]{\small Yves DEBAVEYE}
\author[1, 7, 8]{\small Laure WYNANTS}
\author[1, 7, 9]{\small Ben VAN CALSTER}

\affil[1]{\footnotesize Department of Development \& Regeneration, KU Leuven}
\affil[2]{\footnotesize Management Information Reporting Department, University Hospitals Leuven, Belgium}
\affil[3]{\footnotesize Faculty of Medicine, KU Leuven, Belgium}
\affil[4]{\footnotesize Vascular Access Specialty Team, University Hospitals Leuven, Belgium}
\affil[5]{\footnotesize Department of Infection Control and Prevention, University Hospitals Leuven, Belgium}
\affil[6]{\footnotesize Department of Cellular and Molecular Medicine, University Hospitals Leuven, Belgium}
\affil[7]{\footnotesize Leuven Unit for Health Technology Assessment Research (LUHTAR), KU Leuven}
\affil[8]{\footnotesize School for Public Health and Primary Care, Maastricht University, Maastricht, the Netherlands}
\affil[9]{\footnotesize Department of Biomedical Data Sciences, Leiden University Medical Center, Leiden}

\providecommand{\tightlist}{%
  \setlength{\itemsep}{0pt}\setlength{\parskip}{0pt}}

\usepackage{longtable,booktabs,array}
\usepackage{calc} 
\usepackage{etoolbox}
\makeatletter
\patchcmd\longtable{\par}{\if@noskipsec\mbox{}\fi\par}{}{}
\makeatother
\IfFileExists{footnotehyper.sty}{\usepackage{footnotehyper}}{\usepackage{footnote}}
\makesavenoteenv{longtable}

\NewDocumentCommand\citeproctext{}{}

\makeatletter
 \let\@cite@ofmt\@firstofone
 \def\@biblabel#1{}
 \def\@cite#1#2{{#1\if@tempswa , #2\fi}}
\makeatother
\newlength{\cslhangindent}
\newlength{\csllabelwidth}
\newenvironment{CSLReferences}[2] 
 {\begin{list}{}{%
  \setlength{\itemindent}{0pt}
  \setlength{\leftmargin}{0pt}
  \setlength{\parsep}{0pt}
  \ifodd #1
   \setlength{\leftmargin}{\cslhangindent}
   \setlength{\itemindent}{-1\cslhangindent}
  \fi
  \setlength{\itemsep}{#2\baselineskip}}}
 {\end{list}}
\usepackage{calc}

\usepackage{booktabs}
\usepackage{longtable}
\usepackage{array}
\usepackage{multirow}
\usepackage{wrapfig}
\usepackage{float}
\usepackage{colortbl}
\usepackage{pdflscape}
\usepackage{tabu}
\usepackage{threeparttable}
\usepackage{threeparttablex}
\usepackage[normalem]{ulem}
\usepackage{makecell}
\usepackage{xcolor}
\begin{document}
\maketitle

\begin{abstract}
Prognostic outcomes related to hospital admissions typically do not suffer from censoring, and can be modeled either categorically or as time-to-event. Competing events are common but often ignored. We compared the performance of random forest (RF) models to predict the risk of central line-associated bloodstream infections (CLABSI) using different outcome operationalizations. We included data from 27478 admissions to the University Hospitals Leuven, covering 30862 catheter episodes (970 CLABSI, 1466 deaths and 28426 discharges) to build static and dynamic RF models for binary (CLABSI vs no CLABSI), multinomial (CLABSI, discharge, death or no event), survival (time to CLABSI) and competing risks (time to CLABSI, discharge or death) outcomes to predict the 7-day CLABSI risk. We evaluated model performance across 100 train/test splits. Performance of binary, multinomial and competing risks models was similar: AUROC was 0.74 for baseline predictions, rose to 0.78 for predictions at day 5 in the catheter episode, and decreased thereafter. Survival models overestimated the risk of CLABSI (E:O ratios between 1.2 and 1.6), and had AUROCs about 0.01 lower than other models. Binary and multinomial models had lowest computation times. Models including multiple outcome events (multinomial and competing risks) display a different internal structure compared to binary and survival models. In the absence of censoring, complex modelling choices do not considerably improve the predictive performance compared to a binary model for CLABSI prediction in our studied settings. Survival models censoring the competing events at their time of occurrence should be avoided.
\end{abstract}


\section{Background and Significance}\label{background-and-significance}

Electronic Health Records (EHR) are commonly used to develop prediction models using statistical or machine learning (ML) methods. Many models utilizing EHR data focus on acute clinical events during hospital or ICU admissions, such as sepsis (Moor et al. (2021), Fleuren et al. (2020), Yan, Gustad, and Nytrø (2022), Deng et al. (2022)), ventilator-associated pneumonia (Frondelius et al. 2023) and acute kidney injury (Hodgson et al. 2017). These models vary with regards to the prediction time point: static (e.g.: early detection at admission) or dynamic models (predictions are updated at different times during the patient follow up).
Models using EHR data are often built against surveillance event definitions. In contrast to outcomes recorded after patient discharge (such as ICD codes), the time on the patient timeline when the clinical event of interest has occurred is typically known whenever surveillance event definitions are used. This makes time-to-event models, also known as survival models, an attractive modeling choice. The outcome of interest is typically a specific event (e.g., sepsis), but competing events may preclude the occurrence of the event of interest (e.g.: being discharged in good health conditions or dying of other causes). Competing events are regularly ignored during model building, which may be detrimental to the models' predictive performance according to statistical literature (Austin, Lee, and Fine 2016).

A particularity of EHR data is that patients are normally followed up until discharge; therefore in-hospital outcomes based on EHR data are not subject to loss to follow up and the outcome can be modeled either categorically (i.e., as binary or multinomial) or using time-to-event approaches. While survival or competing risks models are the preferred methods when censoring is present, the absence of censoring does not render these models invalid; it only simplifies the settings in which these models operate.

Random forests (RFs) (Breiman 2001) are ensemble machine learning models, initially proposed for regression and classification. They have been extended for survival (Ishwaran et al. 2008) and competing risks (Ishwaran et al. 2014) settings, with adapted split rules for the survival outcome (survival difference between the left and right daughter nodes) or competing risks outcome (cause-specific cumulative hazard function or cumulative incidence function).

\section{Objective}\label{objective}

We perform a methodological comparison of random forest models built against different outcome types (binary, multinomial, survival and competing risks) for predicting central line-associated bloodstream infection (CLABSI). CLABSI is a hospital acquired infection defined as a bloodstream infection in patients with a central line that is not related to an infection at another site. CLABSI prediction models are built primarily on EHR data and most models use a binary outcome without a fixed time horizon (CLABSI event at any time during admission) or survival models not accounting for competing risks (Gao et al. 2023). Most models are static, making predictions at a fixed moment in time (usually at catheter placement), without updating the predictions throughout the patient's hospitalization.

The question we aim to answer is whether incorporating more information in the outcome (exact event times for survival and competing risks; additional events in multinomial and competing risks models) leads to improved performance compared to the binary model to predict the 7 day risk of CLABSI. We build both static and dynamic models and consider discharge, catheter removal and patient death as competing events that preclude the occurrence of the event of interest.

\section{Materials and Methods}\label{materials-and-methods}

\subsection{Study design and participants}\label{study-design-and-participants}

Patient data are extracted from the EHR system of the University Hospitals Leuven for hospital admissions in the period January 2012 - December 2013. We included patient admissions with registration of one or multiple central lines: centrally inserted central catheter (CICC), tunneled cuffed and non-cuffed central venous catheter, port catheter (Totally Implanted Vascular Access Devices - TIVAD), peripherally inserted central catheter (PICC) and dialysis catheter. The terms ``central line'' and ``catheter'' will be used interchangeably. Patients with only a dialysis catheter (and no other catheter type) were included only if they were admitted to ICU (due to the data extraction constraint). Because the neonatology department did not record catheters in the EHR system before October 2013, 260 neonates admissions have been excluded. The dataset consists of 27478 hospital admissions after excluding neonates.

The following levels of the outcome are considered:

\begin{itemize}
\tightlist
\item
  \textbf{CLABSI}: any laboratory-confirmed bloodstream infection (LC-BSI) during hospital admission for a patient with central line or within 48 hours after the central line removal that is not present in the first 48 hours after admission, that is not a secondary infection, is not a skin contamination and is not a mucosal barrier injury LC-BSI. The CLABSI definition has been calculated retrospectively based on the extracted data following the Sciensano definition published in 2019 (Duysburgh 2019).
\end{itemize}

\begin{itemize}
\item
  \textbf{Discharge}: Hospital discharge or 48 hours after catheter removal, whichever happens first. According to the Sciensano definition, the patient remains at risk of CLABSI for 48 hours after catheter removal.
\item
  \textbf{Death}: Either the first contact with palliative care during admission, transfer to palliative care or patient death, whichever happens first. Patients stop being closely monitored in palliative care and predictions on this ward are not actionable.
\end{itemize}

\begin{figure}
\includegraphics[width=400px]{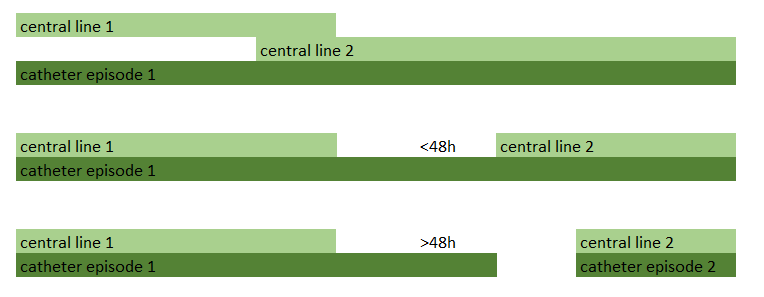} \caption{Catheter episodes}\label{fig:catheter-episodes}
\end{figure}

Patient admissions are split in catheter episodes. A catheter episode starts at catheter placement or at the registration of the first catheter observation during that admission (e.g.: in case of admission to hospital with a long-term catheter) and ends when no catheter observation is made for 48 hours or when discharge, death or CLABSI occurs. In case of multiple concomitant catheters (e.g.: admission with a long-term catheter and placement of another central venous catheter during hospitalization), these are grouped in the same catheter episode if they are overlapping or if there is a time gap of less than 48 hours between removal of one catheter and placement of a new one (Figure \ref{fig:catheter-episodes}).

A catheter episode is further split in landmarks (LMs). The exact time of the first catheter observation is considered landmark 0 (LM0). Subsequent landmarks are created every 24 hours (LM1 is 24 hours after LM0, LM2 48 hours after LM0, and so on up to LM30); each LM corresponds to a catheter-day in the catheter episode. Predictions are made every 24 hours. The landmark dataset contains in rows each LM and in columns the features which may have varying values on different landmarks. An excerpt of the dataset is presented in Supplementary Material 1.

The dataset consists of 27478 admissions, 30862 catheter episodes with complete follow up for all event types: 970 CLABSI, 1466 deaths and 28426 discharges. There are in total 227928 landmarks.

\subsection{Prediction horizon and outcome types}\label{prediction-horizon-and-outcome-types}

The risk of CLABSI in any of the following 7 days is predicted at each landmark. We will represent the outcome in four different ways:

\begin{itemize}
\tightlist
\item
  Binary: CLABSI vs.~no CLABSI within 7 days.
\item
  Multinomial: CLABSI, death, discharge and no event within 7 days.
\item
  Survival: status 1 if CLABSI occurs and 0 (censored) if no CLABSI occurs; the event time is the time of CLABSI occurrence or the censoring time.
\item
  Competing risks (CR): CLABSI, death and discharge with their event times.
\end{itemize}

\subsection{Features}\label{features}

The dataset consists of 302 baseline and time-varying features comprising admission and demographics, medication, laboratory test results, comorbidities, vital signs and catheter registrations. The features have been reviewed by three clinical experts (an infection preventionist, a clinical nurse specialist in vascular access and a critical care physician) and assessed as important or unimportant for CLABSI prediction. Out of the complete set of features, 21 have been selected for the model building, based on the clinical review as well as their inclusion in other CLABSI prediction studies (Gao et al. 2023). The features included in the model and the patient characteristics at baseline are presented in Table \ref{tab:table-1}. Full features descriptions are available in Supplementary Material 2.

\begin{table}

\caption{\label{tab:table-1}Patient characteristics at baseline using the features included in the models (n = 30862 catheter episodes); mean (sd) for continuous variables, before missing data imputation; n (\%) for categorical variables. Patients can have two different catheter types simultaneously with different locations, therefore catheter type and location have been coded as binary and not categorical features with multiple categories.}
\centering
\resizebox{\linewidth}{!}{
\begin{tabular}[t]{>{\raggedright\arraybackslash}p{8em}>{\raggedright\arraybackslash}p{12em}lllll}
\toprule
Feature name & Description & Statistic & Total & CLABSI & Death & Discharge\\
\midrule
 &  &  & n = 30862 & n = 970 & n = 1466 & n = 28426\\
CVC & Catheter type CICC & 0, n(\%) & 17264 (55.9) & 340 (35.1) & 872 (59.5) & 16052 (56.5)\\
 &  & 1, n(\%) & 13598 (44.1) & 630 (64.9) & 594 (40.5) & 12374 (43.5)\\
Port catheter & Catheter type TIVAD & 0, n(\%) & 16855 (54.6) & 768 (79.2) & 744 (50.8) & 15343 (54.0)\\
 &  & 1, n(\%) & 14007 (45.4) & 202 (20.8) & 722 (49.2) & 13083 (46.0)\\
\addlinespace
Tunneled CVC & Catheter type tc-CICC or t-CICC & 0, n(\%) & 29489 (95.6) & 870 (89.7) & 1426 (97.3) & 27193 (95.7)\\
 &  & 1, n(\%) & 1373 (4.4) & 100 (10.3) & 40 (2.7) & 1233 (4.3)\\
PICC & Catheter type PICC & 0, n(\%) & 29579 (95.8) & 941 (97.0) & 1380 (94.1) & 27258 (95.9)\\
 &  & 1, n(\%) & 1283 (4.2) & 29 (3.0) & 86 (5.9) & 1168 (4.1)\\
Jugular & Catheter location jugular & 0, n(\%) & 14104 (45.7) & 581 (59.9) & 592 (40.4) & 12931 (45.5)\\
\addlinespace
 &  & 1, n(\%) & 16758 (54.3) & 389 (40.1) & 874 (59.6) & 15495 (54.5)\\
Subclavian & Catheter location subclavian & 0, n(\%) & 18793 (60.9) & 448 (46.2) & 1026 (70.0) & 17319 (60.9)\\
 &  & 1, n(\%) & 12069 (39.1) & 522 (53.8) & 440 (30.0) & 11107 (39.1)\\
TPN & Total parenteral nutrition (TPN) order in previous 7 days & 0, n(\%) & 28494 (92.3) & 686 (70.7) & 1292 (88.1) & 26516 (93.3)\\
 &  & 1, n(\%) & 2368 (7.7) & 284 (29.3) & 174 (11.9) & 1910 (6.7)\\
\addlinespace
AB & Antibacterials order in previous 7 days & 0, n(\%) & 13060 (42.3) & 279 (28.8) & 563 (38.4) & 12218 (43.0)\\
 &  & 1, n(\%) & 17802 (57.7) & 691 (71.2) & 903 (61.6) & 16208 (57.0)\\
Chemotherapy & Antineoplastic agents order in previous 7 days & 0, n(\%) & 24720 (80.1) & 897 (92.5) & 1410 (96.2) & 22413 (78.8)\\
 &  & 1, n(\%) & 6142 (19.9) & 73 (7.5) & 56 (3.8) & 6013 (21.2)\\
CLABSI history & CLABSI history & 0, n(\%) & 30304 (98.2) & 940 (96.9) & 1434 (97.8) & 27930 (98.3)\\
\addlinespace
 &  & 1, n(\%) & 558 (1.8) & 30 (3.1) & 32 (2.2) & 496 (1.7)\\
Tumor history & Tumor history & 0, n(\%) & 14078 (45.6) & 598 (61.6) & 634 (43.2) & 12846 (45.2)\\
 &  & 1, n(\%) & 16784 (54.4) & 372 (38.4) & 832 (56.8) & 15580 (54.8)\\
Temperature & Max. temperature in last 24 hours [°C] & Mean (SD) & 36.8 (0.8) & 37.0 (0.9) & 37.0 (0.9) & 36.8 (0.8)\\
Systolic BP & Last systolic blood pressure in last 24 hours [mmHg] & Mean (SD) & 125.8 (22.3) & 123.0 (23.5) & 121.3 (25.7) & 126.1 (22.1)\\
\addlinespace
WBC & White blood cell count in last 24 hours [10**9/L] & Mean (SD) & 9.4 (8.4) & 10.4 (12.1) & 11.7 (13.5) & 9.2 (7.7)\\
Lymphoma history & Lymphoma history & 0, n(\%) & 29480 (95.5) & 928 (95.7) & 1387 (94.6) & 27165 (95.6)\\
 &  & 1, n(\%) & 1382 (4.5) & 42 (4.3) & 79 (5.4) & 1261 (4.4)\\
Transplant history & Transplant history & 0, n(\%) & 29355 (95.1) & 904 (93.2) & 1391 (94.9) & 27060 (95.2)\\
 &  & 1, n(\%) & 1507 (4.9) & 66 (6.8) & 75 (5.1) & 1366 (4.8)\\
\addlinespace
Other infection than BSI & Other infection than BSI in previous 17 days & 0, n(\%) & 27768 (90.0) & 847 (87.3) & 1252 (85.4) & 25669 (90.3)\\
 &  & 1, n(\%) & 3094 (10.0) & 123 (12.7) & 214 (14.6) & 2757 (9.7)\\
CRP & C-Reactive Protein Unit in last 24 hours [mg/L] & Mean (SD) & 48.7 (71.9) & 59.3 (84.0) & 95.3 (96.6) & 45.1 (68.1)\\
Admission source Home & Admission source: home & 0, n(\%) & 3728 (12.3) & 240 (24.9) & 344 (23.7) & 3144 (11.2)\\
 &  & 1, n(\%) & 26637 (87.7) & 725 (75.1) & 1107 (76.3) & 24805 (88.8)\\
\addlinespace
MV & Mechanical ventilation in last 24 hours & 0, n(\%) & 29352 (95.1) & 872 (89.9) & 1253 (85.5) & 27227 (95.8)\\
 &  & 1, n(\%) & 1510 (4.9) & 98 (10.1) & 213 (14.5) & 1199 (4.2)\\
ICU & Patient currently in ICU (Intensive Care Unit) ward & 0, n(\%) & 25807 (83.6) & 683 (70.4) & 973 (66.4) & 24151 (85.0)\\
 &  & 1, n(\%) & 5055 (16.4) & 287 (29.6) & 493 (33.6) & 4275 (15.0)\\
Days to event & Days to event & Mean (SD) & 7.3 (9.4) & 13.0 (14.6) & 9.8 (13.6) & 7.0 (8.8)\\
\bottomrule
\end{tabular}}
\end{table}

\subsection{Train / test split}\label{train-test-split}

One hundred random train/test splits are generated on the landmark dataset, keeping two thirds of the hospital admissions for training and one third for test, so that a full admission (with all its catheter episodes and all its landmarks) falls entirely in train set or entirely in the test set. Baseline datasets have then been generated by filtering the dynamic datasets for LM0. Missing data have been imputed using a combination of mean/mode imputation, normal value imputation and the missForestPredict algorithm (Albu 2023) (Supplementary Material 4). The average number and proportion of events in the baseline and dynamic training and test sets are presented in Table \ref{tab:table-n-p-events}. Cumulative incidence functions (CIF) for all events are presented in Supplementary Material 3.

\begin{table}

\caption{\label{tab:table-n-p-events}Average number and average proportion of events in train and test sets; for baseline datasets the averaging is done over all catheter episodes; for dynamic datasets the averaging is done over all landmarks}
\centering
\resizebox{\linewidth}{!}{
\begin{tabular}[t]{lllllll}
\toprule
Baseline/Dynamic & Horizon & Train/Test & CLABSI, n (\%) & Death, n (\%) & Discharge, n (\%) & No event until horizon, n (\%)\\
\midrule
Baseline & Any & Train & 646 (3.1\%) & 977 (4.7\%) & 18947 (92.1\%) & 0 (0\%) by definition\\
Baseline & Any & Test & 324 (3.1\%) & 489 (4.8\%) & 9479 (92.1\%) & 0 (0\%) by definition\\
Baseline & 7 days & Train & 269 (1.3\%) & 558 (2.7\%) & 13039 (63.4\%) & 6704 (32.6\%)\\
Baseline & 7 days & Test & 135 (1.3\%) & 279 (2.7\%) & 6522 (63.4\%) & 3356 (32.6\%)\\
Dynamic & Any & Train & 8700 (5.4\%) & 10050 (6.2\%) & 143264 (88.4\%) & 0 (0\%) by definition\\
\addlinespace
Dynamic & Any & Test & 4376 (5.4\%) & 5078 (6.3\%) & 71716 (88.4\%) & 0 (0\%) by definition\\
Dynamic & 7 days & Train & 1152 (0.7\%) & 1905 (1.2\%) & 48327 (29.8\%) & 110631 (68.3\%)\\
Dynamic & 7 days & Test & 584 (0.7\%) & 945 (1.2\%) & 24176 (29.8\%) & 55464 (68.3\%)\\
\bottomrule
\end{tabular}}
\end{table}

\subsection{Static and dynamic model building using random forests}\label{static-and-dynamic-model-building-using-random-forests}

On each imputed baseline and dynamic training set, RF models are trained for the different outcome types (binary, multinomial, survival and competing risks). Baseline (static) models are built on the baseline datasets (only LM0) and dynamic models are built on the landmark dataset. The dynamic models are built considering each landmark in each catheter episode an independent observation; the landmark number is included in the dynamic model, allowing the dynamic models to account for time effects as well as interactions with time.

Random forest models are built using the randomforestSRC package (Ishwaran, Kogalur, and Kogalur 2023). Each tree is built on an ``in-bag'' (a sample of the training set) and evaluated on an ``out-of-bag'' (the remaining observations in the train set that are not in-bag) for hyperparameter tuning. The in-bags are created by sampling complete admissions with all their catheter episodes, so that an admission falls completely in-bag or completely out-of-bag. Sampling is done with replacement (bootstraps of size equal to the number of catheter episodes) for baseline models and without replacement (subsamples with tuned subsample size) for dynamic models; we consider that when data are large enough (Probst, Wright, and Boulesteix 2019) we can build the trees on subsamples from the data; because of the limited number of events in the baseline data, we consider bootstraps a more appropriate strategy. The number of trees is fixed to 1000 trees, considering this as large enough and unnecessary to tune (Probst, Wright, and Boulesteix 2019). We thus create 1000 in-bags for each of the 100 training sets. Because randomforestSRC package imposes the limitation that all in-bags for the 1000 trees must have the same size and sampling by admission id will not result in equal in-bag sizes as not all admissions have an equal numbers of catheter episodes (and landmarks for the dynamic models), we further adjusted the in-bags to the minimum size (minsize) of all the 1000 inbags for all trees by randomly sampling out some catheter episodes and/or landmarks, which will consequently fall out-of-bag. This procedure had negligible effects on model performance (Supplementary Material 5).

\begin{figure}
\includegraphics[width=500px]{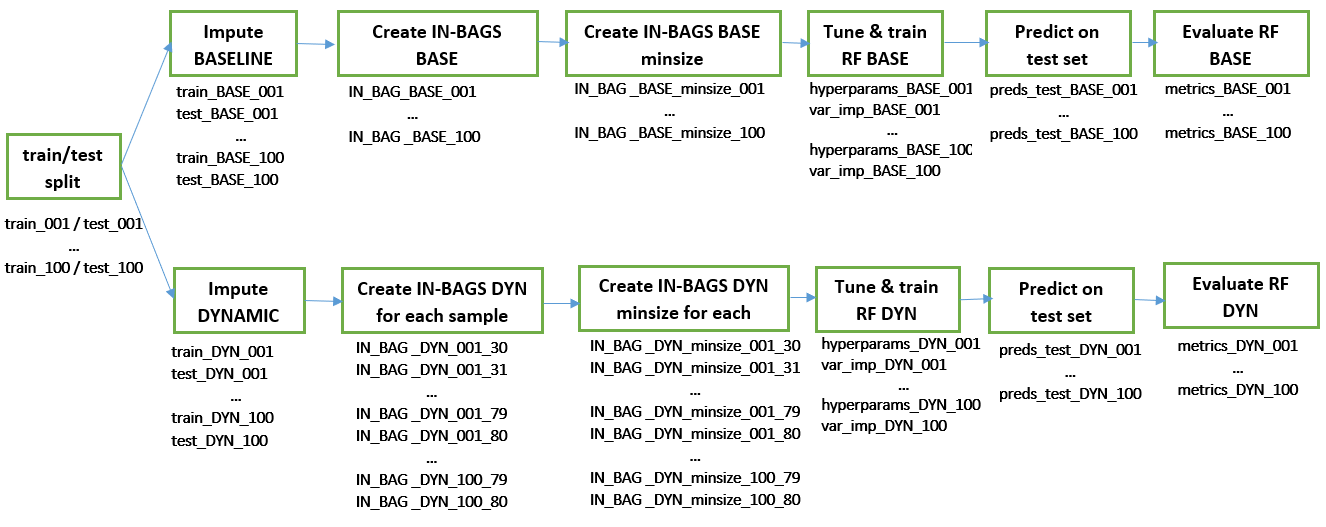} \caption{Model building pipeline}\label{fig:pipeline}
\end{figure}

The tuned hyperparameters are the number of variables selected at each split (mtry) and the minimum size of a terminal node (nodesize) for both baseline and dynamic models. Additionally for dynamic models, the subsample size is tuned. Model tuning is based on the out-of-bag binary logloss for next 7 days prediction to allow all models to perform at their best for the outcome of interest. Model based optimization tuning is performed using the mlrMBO R package (Bischl et al. 2017). (details in the Supplementary Material 6). The best hyperparameters are chosen and a final RF model is built using these hyperparameters. The hyperparameter values used to build the final model, as well as a variable importance measure (the minimum depth of the maximal subtree (Ishwaran et al. 2021)) are saved. The runtimes for model tuning, final model building and prediction on the test set are logged. The pipeline is schematically presented in Figure \ref{fig:pipeline}.

Survival and competing risks models are unfeasible to train on a large dataset when using all event times. We have manipulated the event times for all events (CLABSI, death and discharge) to speed up computations by discretizing the event times and applying administrative censoring (Houwelingen and Putter 2011). Additionally, for survival and competing risks models we compare different censoring, split rules or weighting of events in the split rule, as described in Table \ref{tab:table-surv-CR}.

\begin{table}

\caption{\label{tab:table-surv-CR}Additional options for survival and competing risks models}
\centering
\begin{tabular}[t]{>{\raggedright\arraybackslash}p{6em}>{\raggedright\arraybackslash}p{30em}>{\raggedright\arraybackslash}p{6em}}
\toprule
Option & Description & Models\\
\midrule
Discretized event times & The split statistics are calculated for all event times present in the dataset which proves computationally inefficient. To considerably speed up computations, event times have been discretized using ceiling(t), e.g.: an event happening at day 6.1 is considered to have happened at time 7. & Survival and competing risks models\\
Administrative censoring & We have applied artificial censoring as suggested by Van Houwelingen and Putter by considering catheter episodes with events after a specific time t as censored. Two options are used for administrative censoring: (1) at day 7 (the time of interest for the predictions) or (2) at day 30 (92\% of the events happen in the first 30 days after the start of the catheter episode; we consider it close to no administrative censoring). & Survival and competing risks models\\
Censoring & For survival models we use two types of censoring for the competing events (death and discharge): (1) censoring at event time, e.g.: a discharge happening at day 5 is censored at time 5; (2) censoring at day 7 all competing events that happened before day 7, e.g.: a discharge happening at time 5 is censored at time 7, keeping in the risk set catheter episodes with competing events until our time of interest, as done in Fine-Gray models & Survival models\\
Split rule & The two split rules available for competing risks models are compared: (1) "logrank", based on the difference in cause-specific hazard function and advised when the interest lies in a specific event and (2) "logrankCR" based on the difference in event-specific cumulative incidence and advised for prediction settings. & Competing risks models\\
Cause & When splitting a node in a competing risks model, the split-statistics weight by default all events equally. We compare: (1) the default equal weighting of events with (2) custom weighting (specified in parameter “cause”) using weight 1 for CLABSI and 0 for the other events (death and discharge outcome levels will not be used in determining tree splits). & Competing risks models\\
\bottomrule
\end{tabular}
\end{table}

The overview of all models is presented in Table \ref{tab:table-models}. We compare 14 baseline models and 8 dynamic models. We chose not to run the dynamic models with administrative censoring at day 30 because of increased computation time.

\begin{table}

\caption{\label{tab:table-models}Models}
\centering
\resizebox{\linewidth}{!}{
\begin{tabular}[t]{lll>{\raggedright\arraybackslash}p{16em}ll}
\toprule
Model name & Outcome type & Splitrule & Other hyperparameters / options & Baseline model & Dynamic model\\
\midrule
bin & Binary & gini &  & YES & YES\\
multinom & Multinomial & gini &  & YES & YES\\
surv7d & Survival & logrank & Administrative censoring at day 7; Censoring death and discharge at event time & YES & YES\\
surv7d\_cens7 & Survival & logrank & Administrative censoring at day 7; Censoring death and discharge at time 7 & YES & YES\\
surv30d & Survival & logrank & Administrative censoring at day 30; Censoring death and discharge at event time & YES & NO\\
\addlinespace
surv30d\_cens7 & Survival & logrank & Administrative censoring at day 30; Censoring death and discharge at time 7 & YES & NO\\
CR7d\_LRCR\_c\_1 & Competing risks & logrankCR & cause = 1 (CLABSI); Administrative censoring at day 7 & YES & YES\\
CR7d\_LR\_c\_1 & Competing risks & logrank & cause = 1 (CLABSI); Administrative censoring at day 7 & YES & YES\\
CR7d\_LRCR\_c\_all & Competing risks & logrankCR & cause = default (all events have equal weights); Administrative censoring at day 7 & YES & YES\\
CR7d\_LR\_c\_all & Competing risks & logrank & cause = default (all events have equal weights); Administrative censoring at day 7 & YES & YES\\
\addlinespace
CR30d\_LRCR\_c\_1 & Competing risks & logrankCR & cause = 1 (CLABSI); Administrative censoring at day 30 & YES & NO\\
CR30d\_LR\_c\_1 & Competing risks & logrank & cause = 1 (CLABSI); Administrative censoring at day 30 & YES & NO\\
CR30d\_LRCR\_c\_all & Competing risks & logrankCR & cause = default (all events have equal weights); Administrative censoring at day 30 & YES & NO\\
CR30d\_LR\_c\_all & Competing risks & logrank & cause = default (all events have equal weights); Administrative censoring at day 30 & YES & NO\\
\bottomrule
\end{tabular}}
\end{table}

Models tuning, building and predictions have been run on a high-performance computing cluster using 36 cores and 128 GB RAM. R version 4.2.1 and randomForestSRC 3.2.2 have been used to build the models.

\subsection{Model evaluation}\label{model-evaluation}

Predictions are made on the test sets for each of the model types. We retain the predicted risks of developing CLABSI in any of the next 7 days. We assume CLABSI risk within 7 days is the only outcome of clinical interest and the users of the model are not interested in predictions with other prediction horizons, nor are they interested in prediction of additional events (death or discharge). Survival models yield as predictions the survival probability for different horizons. We retain \(1 - p(survival\ upto\ day\ 7)\). Competing risks models predict the cumulative incidence function conditional on predictors and we retain the predicted CIF of CLABSI at day 7.

Following metrics are evaluated on each test set: AUPRC (Area Under the Precision Recall Curve), AUROC (Area Under the ROC curve), BSS (Brier Skill Score), E:O ratio (the mean of predicted risks divided by the mean of observed binary events), calibration slope (calculated by regressing the true binary outcome on the logit of the predicted risks (Van Calster et al. 2019)) and ECI (Estimated Calibration Index; the mean squared difference between the predicted probabilities and the predicted probabilities obtained with a loess fit of the observed outcome on the predicted risks, multiplied by 100 (Van Hoorde et al. 2015)). For dynamic models, time-dependent metrics (metrics calculated at each LM) are presented. The median and interquartile range (IQR) for each metric over the 100 test sets are compared.

\subsection{Code availability}\label{code-availability}

The code used for model tuning, prediction and model evaluation is available at: \url{https://github.com/sibipx/CLABSI_compare_RFSRC_models}

\section{Results}\label{results}

\subsection{Predictive performance for baseline models}\label{predictive-performance-for-baseline-models}

\begin{figure}
\centering
\includegraphics{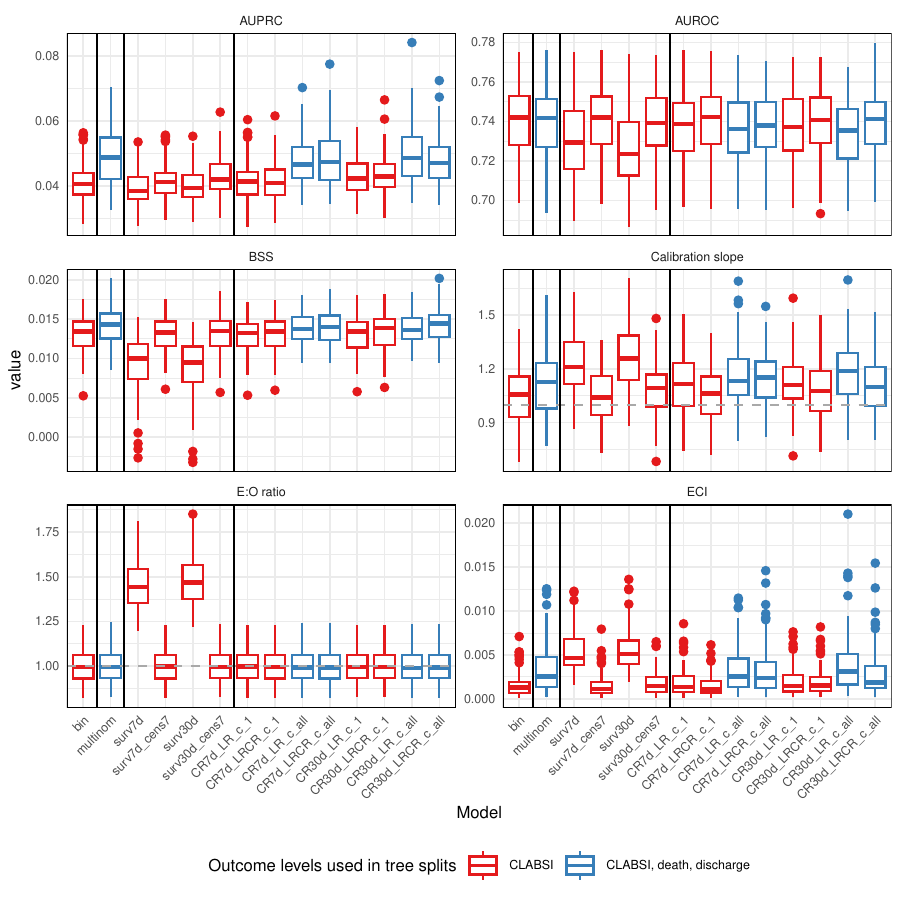}
\caption{\label{fig:baseline-cutoff-independent}Prediction performance for baseline models. From left to right, separated by vertical lines: binary outcome model, multinomial outcome model, survival models, competing risk models. Models that consider all outcome classes to determine splits are displayed in blue, models that only consider CLABSI to determine splits are displayed in red.}
\end{figure}

Figure \ref{fig:baseline-cutoff-independent} shows the baseline models performance. Survival models with competing events censored at their occurrence time (surv7d and surv30d) show overestimated predicted risks (median E:O ratio of 1.44 and 1.47 respectively, compared to values between 0.99 and 1.00 for the other models) and poorer discrimination (median AUROC 0.729 for surv7d and 0.724 for surv30d, compared to AUROC between 0.735 and 0.742 for the other models). However, if individuals experiencing competing risks are kept in the risk set by censoring the event at the time of interest for predictions (surv7d\_cens7 and surv30d\_cens7 models), the performance becomes comparable to other models: median E:O ratio of 1.00 and AUROC of 0.742 and 0.739 respectively.

\begin{figure}
\centering
\includegraphics{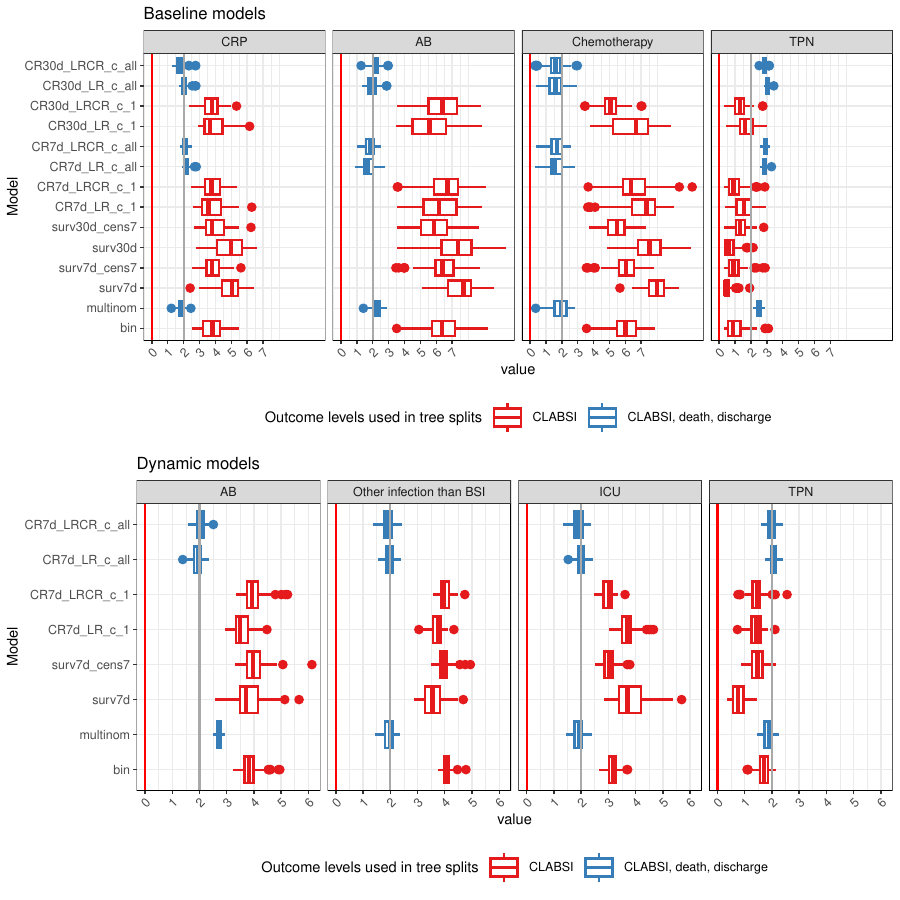}
\caption{\label{fig:split-depth-boxplots}Feature minimal depth for baseline and dynamic models. Includes only a subset of `important' features for which the median minimal depth is less than 2 in at least one model type. Lower minimal depths indicate more important variables. The minimal depth for all features is included in Supplementary Material 7. Models that consider all outcome classes to determine splits are in blue, models that only consider CLABSI to determine splits are in red.}
\end{figure}

Models using all outcome levels (multinomial, CR7d\_LRCR\_c\_all, CR7d\_LR\_c\_all, CR30d\_LRCR\_c\_all, CR30d\_LR\_c\_all) show slightly higher AUPRC, despite slight miscalibration visible in the ECI, but with no noticeable difference in AUROC. These models venture into predicting higher risks than the other models and these larger predictions lead to miscalibration (calibration curves in Supplementary Material 7 and prediction plots in Supplementary Material 10) but better discrimination. Despite small differences in performance metrics, the internal structure of these models is different. Figure (\ref{fig:split-depth-boxplots}) presents the minimal depth of the maximal subtree (Ishwaran et al. 2021), which is the depth in a tree on which the first split is made on a variable \(v\), averaged over all trees in the forest. The lowest possible value is 0 (root node split). Chemotherapy, antibiotics and CRP features are selected at first splits for models using all outcome levels, while binary and survival models favor TPN at early splits. These models also differ in hyperparameter values and predicted risk distributions (Supplementary Material 7).

The choice for administrative censoring for survival and CR model (day 7 or day 30) or the splitrule used by CR models (logrank or logrankCR) did not yield any visible difference.

\subsection{Predictive performance for dynamic models}\label{predictive-performance-for-dynamic-models}

\begin{figure}
\centering
\includegraphics{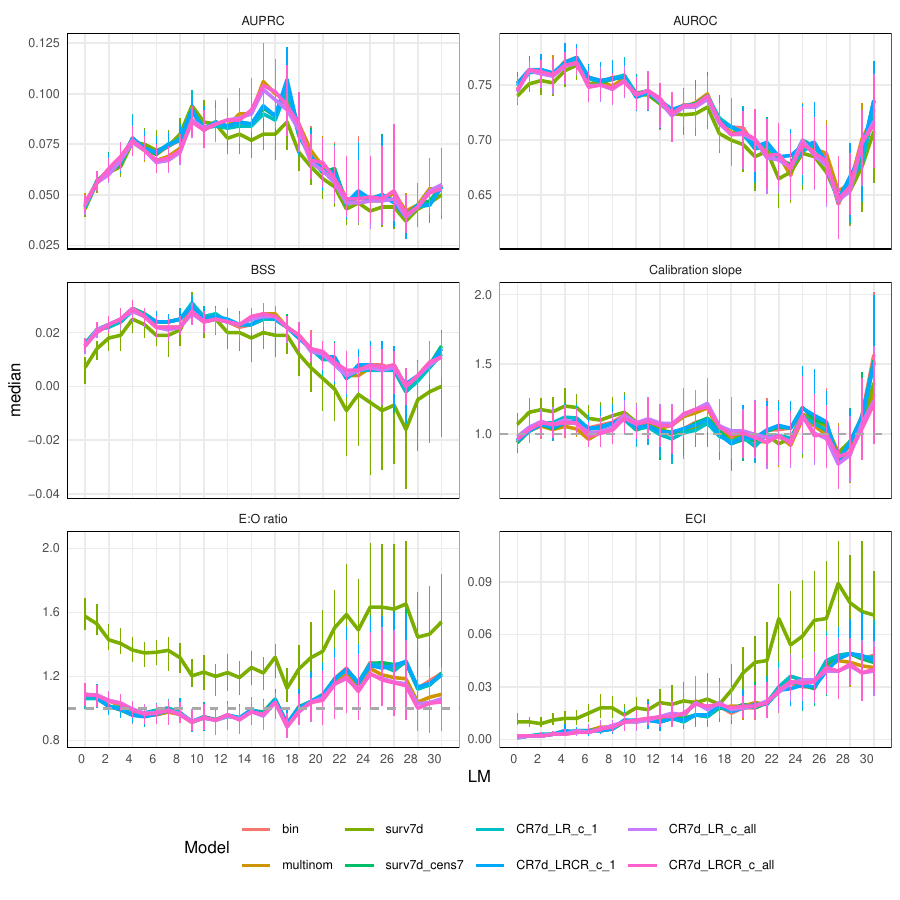}
\caption{\label{fig:t-dynamic-cutoff-independent}Prediction performance for dynamic models - time dependent metrics. The median value of each metric is plotted over time (landmark) and the vertical bars indicate the IQR}
\end{figure}

Figure \ref{fig:t-dynamic-cutoff-independent} shows the dynamic models' performance. All models reach their maximum AUROC at LM5. Similar to baseline models, the survival model with competing events censored at their occurrence time (surv7d) shows overestimated predicted risks (median E:O ratio of 1.58 at LM0 and 1.35 at LM5, compared to values between 1.06 and 1.09 at LM0 and between 0.95 and 0.97 at LM5 for the other models) and slightly poorer discrimination (median AUROC 0.740 at LM0 and 0.768 at LM5, compared to AUROC between 0.744 and 0.752 at LM 0 and 0.770 and 0.775 at LM 5 for the other models). Censoring the competing event at the time of interest for predictions instead of at its event time (surv7d\_cens7 model) corrects this loss of calibration and discrimination. The models using multiple outcome levels choose at the first splits in the trees antibiotics, other infection than BSI and ICU (Figure \ref{fig:split-depth-boxplots}), while the other models favor TPN at early splits. They also differ in tuned hyperparameter values but do not exhibit an obvious difference in predicted risk distribution (Supplementary Material 7) or in performance metrics. Dynamic models perform better than baseline baseline models at LM0 in terms of discrimination (AUROC) and BSS, but worse in terms of calibration (Supplementary Material 7).

\subsection{Computational speed (baseline and dynamic)}\label{computational-speed-baseline-and-dynamic}

\begin{figure}
\centering
\includegraphics{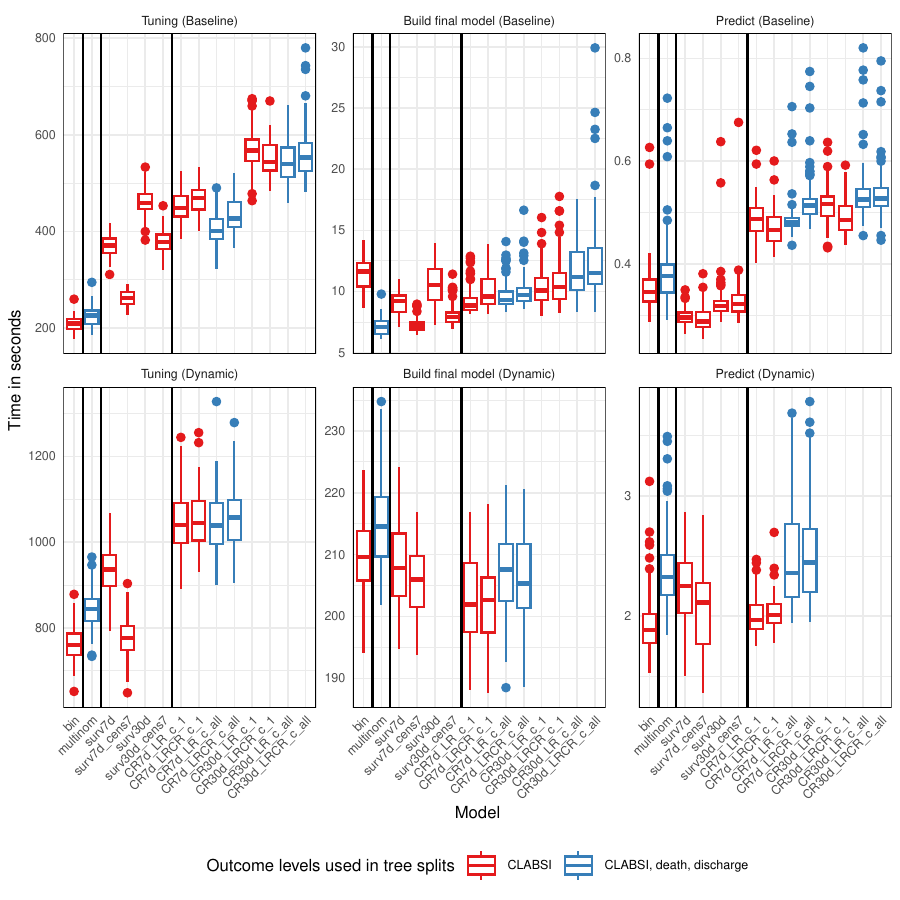}
\caption{\label{fig:timings-all}Runtimes for all models}
\end{figure}

The runtimes are presented in Figure \ref{fig:timings-all}. The tuning time increases with model complexity, with binary, multinomial and the survival model with administrative censoring at day 7 and event censoring at day 7 being the fastest for both baseline and dynamic models; censoring at a later time (30 days) produces higher runtimes for baseline models and competing risks models take generally longest to tune. The runtimes vary less for the final model building, probably depending more on the final hyperparameter values. The prediction times also increase with model complexity but are fast for all models, with a median of less than one second for an entire baseline test set (on average slightly more than 10000 observations) and less than two seconds for a dynamic test set (on average slightly more than 80000 observations).

\section{Discussion}\label{discussion}

Survival and competing risks models are not widely used in machine learning prediction studies, and even less so on EHR data (Goldstein et al. 2017). To our knowledge no other study compared RF models under competing risks settings. We compared different modelling options (binary, multinomial, survival and competing risks outcome) for building baseline and dynamic random forest models for predicting the 7 days CLABSI risk using the randomforestSRC R package.

In our settings, complex models did not display a competitive advantage in terms of model performance. Moreover, complex models were slower to tune. Binary classification models remain the simplest choice for model building, implemented in a wide range of software libraries. Survival RF models that censor the competing event at its time of occurrence yield inferior calibration and discrimination compared to all the other modelling choices. Keeping the observations with competing events in the risk set until the prediction horizon, as done in Fine-Gray models (Fine and Gray 1999), provides reliable estimates of the cumulative incidence function and superior prediction performance. This finding is potentially valuable for researchers with interest in predictions on multiple time horizons but unable to build a competing risks model due to software limitations. For instance, the ranger R package (Wright et al. 2019) supports survival models but not competing risks models; this might be the situation for other model choices or software libraries.

A companion study using regression models on the same dataset also found that binary, multinomial and competing risk models had similar performance (Gao et al. 2024). The RF models in this analysis had higher discrimination: median AUROC up to 0.742 for baseline RF models and 0.775 for dynamic RF models at landmark 5, compared to 0.721 and 0.747 respectively for regression models. The calibration was similar for RF and regression models: median E:O ratio at LM5 between 0.95 and 0.97 for RF models and between 0.940 and 0.977 for regression models. Lower performance for survival models that censor the competing risks at their time of occurrence was also observed in the companion study, and this finding is in line with the extensive research done on statistical survival models (Austin, Lee, and Fine 2016).

For baseline models, different modeling approaches lead to similar model performance in terms of AUROC but different risk predictions in individuals. This is in line with a growing body of research studying the ``between model'' variability for different model types (Ledger et al. 2023) or feature selection choices (Pate et al. 2019). We observed that models using all levels of the outcome (multinomial and competing risks) predicted higher risks than the other models, displayed higher AUPRC and slight miscalibration (higher ECI), but did not display considerable differences in AUROC. Despite the miscalibration, the BSS does not penalize these models, as there is gain of predictive performance in the high predicted risks. Zhou et al demonstrate that for low event rate outcomes, AUPRC puts larger weight on the model's discriminating ability in high risk predictions compared AUROC and conduct a numerical study in which they observe that AUPRC is often correlated with the Brier Score (Zhou et al. 2021). Li et al (Li et al. 2020) conducted a study to predict cardiovascular disease in the presence of censoring using survival models (accounting for censoring) and ML methods that did not account for censoring. They observed that misspecified models (models that ignored censoring) produced lower, more conservative predicted risks and correctly specified models (not ignoring censoring) produced higher predicted risks while the differences in AUROC were minimal.

Despite extensive functionality in randomforestSRC (unsupervised and supervised learning with various outcome types, multiple split rules, missing data imputation, variable importance metrics, variable selection strategies) which makes it stand out in the software packages space, limitations exist with regards to tuning strategies, out-of-bag sampling and its feasibility for large datasets. The tuning objectives (e.g.: logloss, Brier Score) cannot be changed in the default \emph{tune} function of the package and the subsample size is not a tunable parameter; we have in change opted to tune the hyperparameters using mlrMBO (Bischl et al. 2017) which offers great flexibility. The randomforestSRC package does not offer the possibility of user-defined in-bags and out-of-bags of different sample sizes, which are useful for dynamic and clustered data. Users can opt in turn for cross-validation, a slower procedure compared to tuning based on OOB predictions; we have opted to adjust the in-bags to a minimum size and gain computational efficiency. Even after applying all documented suggestions for improving computation time for survival and competing risks models (Ishwaran, Lu, and Kogalur 2021), running these models on our large dynamic dataset would not have been feasible. Discretizing event times and applying administrative censoring, as explained in the Methods section, resulted in split statistics to be calculated over a limited number of horizons of interest and came with significant computational gain. This solution has broader applications; other studies resort to using only binary classification ML models for computational efficiency, even in the presence of censoring (Li et al. 2020), which can lead to decreased sample sizes or introduce bias.

The current study has limitations. As a methodological comparison study, we neither recommend a specific model for clinical use nor evaluate the clinical usefulness of the models. Two important topics remain to be studied further. First, we assume the only outcome of interest is CLABSI event in any of the next 7 days; the undeniable advantage of survival and competing risks models is that they can present the user with predictions for other time horizons in situations where one prediction horizon is of main interest but other horizons might be of secondary interest. Binary models are the fastest to tune, but if we were to build binary models for horizons 1 to 7 (either independent models or built with a monotonicity constraint on predictions), their computational advantage would most probably not persist. Moreover, we are not exploring multiple prediction horizons of interest (7, 14, 30, \ldots) due to the high computational time on our large dataset.

Second, the models that use all levels of the outcome (multinomial and competing risks) display some particularities in tuned hyperparameters, variables used in early tree splits, distribution of predicted risks and slight advantages in some of the performance metrics. The difference in minimal depth for these models is most probably a consequence of optimizing the split statistic over multiple outcome classes. We selected the feature set with CLABSI prediction in mind and we did not explicitly include predictive features for death and discharge. Extensive research would be needed, either using simulated data or real datasets including strong predictors for all outcome levels to understand to which extent the feature selection impacts the model performance when multiple outcome levels are used.

\section{Conclusion}\label{conclusion}

In our studied settings, complex models did not considerably improve the predictive performance compared to a binary model, which can be considered the easiest choice both in terms of model development and in computational time. Importantly, censoring the competing events at their time of occurrence should be avoided in survival models. More research is needed to study the impact of feature selection in models with multiple levels of outcome (multinomial and competing risks).

\section{Acknowledgements}\label{acknowledgements}

This work was supported by the Internal Funds KU Leuven {[}grant C24M/20/064{]}. The funding sources had no role in the conception, design, data collection, analysis, or reporting of this study.

The resources and services used in this work were provided by the VSC (Flemish Supercomputer Center), funded by the Research Foundation - Flanders (FWO) and the Flemish Government.

\section{Author contribution}\label{author-contribution}

ALBU Elena: Conceptualization, Data curation, Formal analysis, Investigation, Methodology,
Software, Visualization, Writing - original draft. GAO Shan: Data curation, Investigation, Formal analysis, Writing - review \& editing. STIJNEN Pieter: Conceptualization, Data curation, Funding acquisition, Writing - review \& editing. RADEMAKERS Frank: Supervision, Data curation, Writing - review \& editing. COSSEY Veerle: Supervision, Data curation, Writing - review \& editing. DEBAVEYE Yves: Supervision, Data curation, Writing - review \& editing. JANSSENS Christel: Supervision, Data curation, Writing - review \& editing. VAN CALSTER Ben: Conceptualization, Funding acquisition, Methodology, Project administration, Supervision, Writing - review \& editing. WYNANTS Laure: Conceptualization, Funding acquisition, Methodology, Project administration, Supervision, Writing - review \& editing.

\section{Conflicts of interest}\label{conflicts-of-interest}

The authors declare that they have no conflicts of interests to disclose.

\section{Data availability}\label{data-availability}

The data used in this study cannot be shared publicly due to for the privacy of individuals that participated in the study.

\section{Ethics}\label{ethics}

The study adhered to the principles of the Declaration of Helsinki (current version), the principles of Good Clinical Practice (GCP), and all relevant regulatory requirements. Ethical review was sought from the Research Ethics Committee UZ/KU Leuven, Belgium, and local ethics committees at participating hospitals. The collection, processing and disclosure of personal data, such as patient health and medical information were in compliance with applicable personal data protection and the processing of personal data (Directive 95/46/EC and Belgian law of December 8, 1992 on the Protection of the Privacy in relation to the Processing of Personal Data). Patient stay identifiers were coded using the pseudo-identifier available in the data warehouse of the participating hospital.

\section{Abbreviations}\label{abbreviations}

AUPRC: Area Under the Precision Recall Curve

AUROC: Area Under the Receiver Operating Characteristic curve

BS: Brier Score

BSI: Bloodstream Infection

BSS: Brier Skill Score

CICC: Centrally Inserted Central Catheters (CICC)

CIF: Cumulative Incidence Functions

CLABSI: Central Line-Associated Bloodstream Infections

CR: Competing Risks

CRP: C-reactive protein

CVC: Central Venous Catheter

ECI: Estimated Calibration Index

EHR: Electronic Health Records

FWO: Research Foundation - Flanders

ICD: International Classification of Diseases

ICU: Intensive Care Unit

IQR: Interquartile Range

KWS: Klinisch Werkstation (EHR System)

LC-BSI: Laboratory-Confirmed Bloodstream Infection

LM: Landmark

LR: logrank

LRCR: logrankCR (logrank competing risks)

LWS: Laboratorium Werkstation (Laboratory System)

ML: Machine Learning

NMSE: Normalized Mean Square Error

OOB: Out of bag

PICC: peripherally inserted central catheter

PDMS: Patient Data Management System (ICU Management System)

RAM: Random Access Memory

RF: Random Forest

TIVAD: Totally Implanted Vascular Access Devices

TPN: Total Parenteral nutrition

VSC: Flemish Supercomputer Center

WBC: White Blood Cells count

\section*{References}\label{references}
\addcontentsline{toc}{section}{References}

\phantomsection\label{refs}

\section{Supplementary material}\label{supplementary-material}

\subsection{Supplementary material 1 - Example of stacked dataset for dynamic model building}\label{suppl-data-representation}

An example of stacked dataset for dynamic model building is presented in Table \ref{tab:example-dyn-dataset}. The admission id is kept in the dataset for train/test split but not retained for model building.

\begin{table}

\caption{\label{tab:example-dyn-dataset}Example of stacked data structure including some selected features; missing values are represented with NA; admission IDs are fictive.}
\centering
\resizebox{\linewidth}{!}{
\begin{tabular}[t]{rrrrrrrrlr}
\toprule
Admission ID & Catheter episode & LM & CVC & Subclavian & TPN & Temperature & WBC & Event type & Event time\\
\midrule
1 & 1 & 0 & 1 & 1 & 0 & 38.6 & 4.0 & CLABSI & 9.7\\
1 & 1 & 1 & 1 & 1 & 0 & 38.3 & 3.8 & CLABSI & 9.7\\
1 & 1 & 2 & 1 & 1 & 0 & 38.0 & 5.7 & CLABSI & 9.7\\
1 & 1 & 3 & 1 & 1 & 0 & 37.2 & 3.7 & CLABSI & 9.7\\
1 & 1 & 4 & 1 & 1 & 0 & 37.8 & 3.2 & CLABSI & 9.7\\
\addlinespace
1 & 1 & 5 & 1 & 1 & 0 & 38.8 & 2.7 & CLABSI & 9.7\\
1 & 1 & 6 & 1 & 1 & 0 & 39.5 & 2.4 & CLABSI & 9.7\\
1 & 1 & 7 & 1 & 1 & 0 & 38.3 & 2.6 & CLABSI & 9.7\\
1 & 1 & 8 & 1 & 1 & 0 & 38.1 & 4.4 & CLABSI & 9.7\\
1 & 1 & 9 & 1 & 1 & 0 & 37.0 & 2.7 & CLABSI & 9.7\\
\addlinespace
1 & 2 & 0 & 1 & 1 & 0 & 39.5 & 0.1 & Death & 1.2\\
1 & 2 & 1 & 1 & 1 & 0 & 39.8 & 0.2 & Death & 1.2\\
2 & 1 & 0 & 1 & 1 & 0 & 38.3 & 8.7 & Discharge & 3.5\\
2 & 1 & 1 & 1 & 1 & 0 & 38.1 & NA & Discharge & 3.5\\
2 & 1 & 2 & 1 & 1 & 0 & 37.1 & 9.1 & Discharge & 3.5\\
\addlinespace
2 & 1 & 3 & 1 & 1 & 0 & 37.0 & NA & Discharge & 3.5\\
3 & 1 & 0 & 1 & 1 & 0 & 37.2 & 5.6 & Discharge & 4.2\\
3 & 1 & 1 & 1 & 1 & 0 & 36.6 & 7.2 & Discharge & 4.2\\
3 & 1 & 2 & 1 & 1 & 0 & 36.5 & 8.3 & Discharge & 4.2\\
3 & 1 & 3 & 1 & 1 & 0 & 37.1 & NA & Discharge & 4.2\\
\addlinespace
3 & 1 & 4 & 1 & 1 & 0 & 37.0 & 6.2 & Discharge & 4.2\\
3 & 2 & 0 & 1 & 1 & 0 & 37.7 & NA & Discharge & 2.5\\
3 & 2 & 1 & 1 & 1 & 0 & 37.3 & 6.2 & Discharge & 2.5\\
3 & 2 & 2 & 1 & 1 & 0 & 37.4 & NA & Discharge & 2.5\\
\bottomrule
\end{tabular}}
\end{table}

\subsection{Supplementary material 2 - Features descriptions}\label{suppl-feat-desc}

Baseline variables are invariant for a catheter episode and are known at the start of the catheter episode. Time-varying features represent features that can vary from one landmark to another.

\textbf{Continuous features}: Whenever multiple measurements are available during a time window (typically 24 hours) for continuous features these are aggregated into a single landmark value. The aggregation rule (e.g.: maximum, minimum) with the most clinical significance is chosen (e.g.: maximum temperature in last 24 hours). Whenever no measurements are taken in the time window, the feature value is represented as a missing value.

\textbf{Binary features} are coded for presence or absence of specific clinical events that are recorded in the EHR only when present, e.g.: total parenteral nutrition (TPN).

\textbf{Categorical features} (e.g.: catheter type) that might occur simultaneously in the aggregation window are coded as binary features (0/1) for all values recorded (e.g.: if a patient has two catheters: CICC or TIVAD, two binary features are kept for CICC or TIVAD). Whenever categorical features (with two or more categories) are expected to be recorded regardless of the category (e.g.: admission source), and no value is present, the feature value is represented as a missing value.

Values outside the possible range have been deleted before feature aggregation, for the following variables and ranges (min -- max): temperature, {[}30, 45{]}; systolic and diastolic blood pressure, {[}30, 370{]}; respiratory rate, {[}0, 900{]}; heart rate, {[}0, 500{]}; oxygen saturation, {[}0, 100{]}; CVP (central venous pressure), {[}-5, 20{]}; weight, {[}0.05, 250{]}; length, {[}0.05, 250{]}; glycemia, {[}0, 2000{]}. These deletions might result in missing values to be imputed later.

The descriptions of variables selected for the model building are presented in Table \ref{tab:table-vars-selected}.

\begin{table}

\caption{\label{tab:table-vars-selected}Features included in the model (binary\_all means that all feature values encountered in the aggregation window are kept as binary values (0/1) for categorical features}
\centering
\resizebox{\linewidth}{!}{
\begin{tabular}[t]{ll>{\raggedright\arraybackslash}p{26em}ll}
\toprule
Short name & Feature & Description & Type & Baseline / time-varying\\
\midrule
CVC & CAT\_catheter\_type\_binary\_all\_CVC & Was there a catheter of type CICC connected since previous LM? & binary & TV\\
Port catheter & CAT\_catheter\_type\_binary\_all\_Port\_a\_cath & Was there a catheter of type TIVAD connected since previous LM? & binary & TV\\
Tunneled CVC & CAT\_catheter\_type\_binary\_all\_Tunneled\_CVC & Was there a catheter of type t-CICC or tc-CICCconnected since previous LM? & binary & TV\\
PICC & CAT\_catheter\_type\_binary\_all\_PICC & Was there a catheter of type PICC connected since previous LM? & binary & TV\\
Jugular & CAT\_catheter\_location\_binary\_all\_Collarbone & Was there a catheter connected at location Collarbone since previous LM? & binary & TV\\
\addlinespace
Subclavian & CAT\_catheter\_location\_binary\_all\_Neck & Was there a catheter connected at location Neck since previous LM? & binary & TV\\
CLABSI history & CLABSI\_history & Did the patient experience a CLABSI event in the past 3 months since LM time? & binary & TV\\
Admission source Home & ADM\_admission\_source\_binary\_all\_Home & Admission source (home or other places) & binary & BASE\\
TPN & MED\_7d\_TPN & Has TPN (total parenteral nutrition) been ordered for the patient in the previous 7 days from LM time & binary & TV\\
AB & MED\_L2\_7d\_J01\_ANTIBACTERIALS\_FOR\_SYSTEMIC\_USE & Have any drugs in ATC group (level 2) J01 (ANTIBACTERIALS FOR SYSTEMIC USE) been ordered for the patient in the previous 7 days from LM time & binary & TV\\
\addlinespace
Chemotherapy & MED\_L2\_7d\_L01\_ANTINEOPLASTIC\_AGENTS & Have any drugs in ATC group (level 2) L01 (ANTINEOPLASTIC AGENTS) been ordered for the patient in the previous 7 days from LM time & binary & TV\\
Systolic BP & CARE\_VS\_systolic\_BP\_last & Last value of systolic blood pressure since previous landmark. For baseline (LM 0) the last value from the previous 24 hours is used. & cont & TV\\
Temperature & CARE\_VS\_temperature\_max & Maximum value of temperature since previous landmark. For baseline (LM 0) the last value from the previous 24 hours is used. Only temperatures in the range (30 °C, 45 °C) are kept, the others are deleted. Maximum value is used to correct for very low temperatures measured by devices in ICU, when the temperature falls closer to the room temperature. & cont & TV\\
MV & CARE\_VS\_MV & Is the patient on mechanical ventilation (MV) since previous landmark? A patient is considered on MV if at least one value of PEEP or FiO2 are recorded between 2 landmarks. Only valid for ICU patients & binary & TV\\
ICU & MS\_is\_ICU\_unit & Is the patient now (at the exact second of the current LM) in ICU? & binary & TV\\
\addlinespace
Lymphoma history & COM\_lymphoma\_before\_LM & Has lymphoma been registered as a comorbidity before current LM time? & binary & TV\\
Tumor history & COM\_PATH\_tumor\_before\_LM & Has a tumour pathology been registered before current LM time? & binary & TV\\
Transplant history & COM\_PATH\_transplant\_before\_LM & Has a transplant pathology been registered before current LM time? & binary & TV\\
CRP & LAB\_CRP\_last & CRP, last value since previous LM. Unit: mg/L & cont & TV\\
WBC & LAB\_WBC\_count\_last & WBC count, last value since previous LM. Unit: 10**9/L & cont & TV\\
\addlinespace
Other infection than BSI & MB\_other\_infection\_than\_BSI\_during\_window & Has there been a positive culture, of any other type than blood, in the last 17 days (time window used for secondary BSIs). The validation time of the sample is used (as opposed to the CLABSI calculation, where the date foreseen for the sample collection is used) & binary & TV\\
\bottomrule
\end{tabular}}
\end{table}

\subsection{Supplementary material 3 - Cumulative incidence function curves}\label{suppl-CIF}

\begin{figure}
\centering
\includegraphics{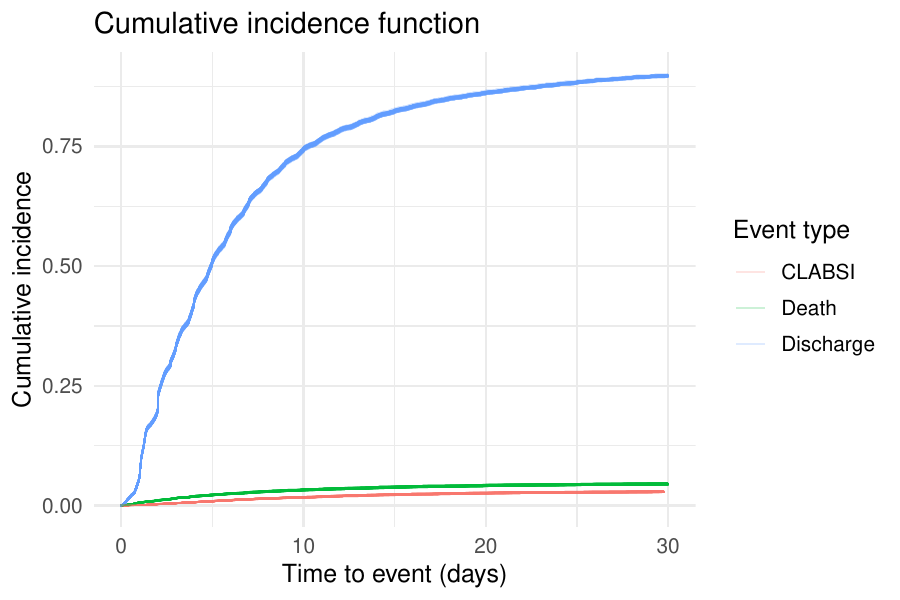}
\caption{\label{fig:CIF-all}Cumulative incidence function curves for all events (all train sets)}
\end{figure}

\begin{figure}
\centering
\includegraphics{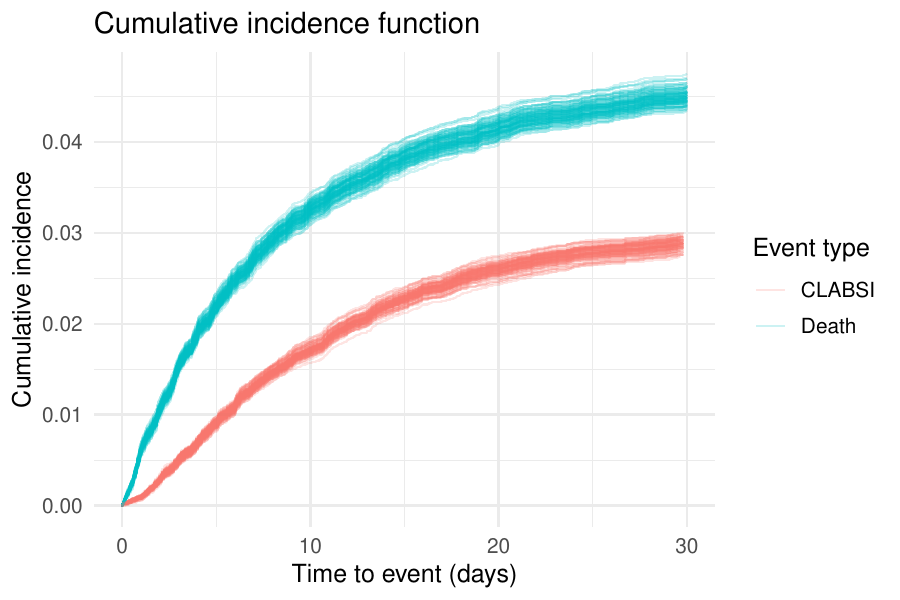}
\caption{\label{fig:CIF-death-CLABSI}Cumulative incidence function curves for CLABSI and Death events (all train sets)}
\end{figure}

Cumulative incidence function curves for all events are presented in Figure \ref{fig:CIF-all}. Cumulative incidence function curves for death and discharge are presented in Figure \ref{fig:CIF-death-CLABSI}.

\subsection{Supplementary material 4 - Missing data imputation}\label{suppl-missing-data}

Missing data have been imputed in the larger context (all 302 features) using a combination of mean/mode imputation, normal value imputation and the missForestPredict algorithm (Albu 2023). The test sets are imputed using the mean/mode or the missForestPredict models learned on the train sets. Table \ref{tab:table-miss-model} describes the missingness of the features included in the baseline and dynamic prediction models.

\begin{table}

\caption{\label{tab:table-miss-model}Features included in the model with missing values; for baseline datasets, catheter episodes are the unit of observation; for dynamic datasets landmarks are the unit of observation}
\centering
\resizebox{\linewidth}{!}{
\begin{tabular}[t]{lrrllll}
\toprule
Feature & Number present & Number missing & Percentage missing & Imputation strategy & Rationale & Baseline/dynamic\\
\midrule
Admission source Home & 30365 & 497 & 1.6\% & mode & less than 3\% missing & Baseline\\
Temperature & 22919 & 7943 & 25.7\% & missForestPredict &  & Baseline\\
Systolic BP & 22616 & 8246 & 26.7\% & missForestPredict &  & Baseline\\
WBC & 17990 & 12872 & 41.7\% & missForestPredict &  & Baseline\\
CRP & 16480 & 14382 & 46.6\% & missForestPredict &  & Baseline\\
\addlinespace
Admission source Home & 226244 & 1684 & 0.74\% & mode & less than 3\% missing & Dynamic\\
Temperature & 215107 & 12821 & 5.63\% & missForestPredict &  & Dynamic\\
Systolic BP & 205661 & 22267 & 9.77\% & missForestPredict &  & Dynamic\\
WBC & 137581 & 90347 & 39.64\% & missForestPredict &  & Dynamic\\
CRP & 135015 & 92913 & 40.76\% & missForestPredict &  & Dynamic\\
\bottomrule
\end{tabular}}
\end{table}

Missing data imputation has been performed separately for baseline data (using only the data available at LM0) and for the dynamic data (``pooling'' together all landmarks from all catheter episodes as independent observations, as in the dataset example in Table \ref{tab:example-dyn-dataset}. We have separated the baseline imputation, as we consider that studies presenting static prediction models utilize only data available at baseline. Three steps have been applied in the imputation process:

\begin{itemize}
\item
  Feature exclusion based on missingness rate or sparsity
\item
  Simple imputation
\item
  missForestPredict imputation
\end{itemize}

\subsubsection{Feature exclusion based on missingness rate or sparsity}\label{feature-exclusion-based-on-missingness-rate-or-sparsity}

Before imputation, some preliminary cleaning has been performed using the following rules for both baseline and dynamic data:

\begin{itemize}
\item
  Based on an exploration of missingness over time, features completely missing for a part of the study timeframe have been excluded (e.g.: RASS, Richmond Agitation Sedation Scale started being recorded in the system as from February 2013 and was completely missing before).
\item
  Features with less than 500 non-missing values are excluded; we consider that building an imputation model for a feature that has less than 500 non-missing values might prove unstable. Applying this rule will result in a different set of excluded features at baseline than in the dynamic data. As dynamic data includes all landmarks, some features that have less than 500 recorded values at baseline will have more than 500 values across all landmarks.
\item
  Some sparse levels in the categorical feature ``medical specialty'' (less than 200 observations in a category) have been collapsed into category ``Other''.
\end{itemize}

\subsubsection{Simple imputation}\label{simple-imputation}

Further, for the three situations below, features have been imputed at baseline using mean/mode, normal value or based on clinical knowledge and using last-observation-carried-forward (LOCF) in the dynamic data. In case of mean/mode imputation, the mean/mode of the training set is used to impute the test set.

\begin{itemize}
\item
  Missing rate less than 3\%
\item
  Features that we consider ``difficult to impute''. To assess if a feature is ``difficult to impute'' we have run the missForestPredict algorithm using one iteration on the full baseline and dynamic datasets before train/test split and inspected the OOB (out-of-bag) normalized mean square error (NMSE) (Albu 2023). We have used only one iteration because at the first iteration the OOB error is less subject to bias than at later iterations (when imputed values of one feature are reinforced by imputed values of another feature that in turn was imputed based on the first feature). A value of NMSE close to 1 means that the algorithm does not provide imputations much superior to the mean imputation. Features with OOB NMSE at first iteration greater than 0.9 have been considered ``difficult to impute.''
\item
  The number of catheter lumens is a special situation because their imputation makes sense only within the categories of catheter types. They have been imputed with typical (normal) values or based on clinical knowledge.
\end{itemize}

The baseline and dynamic features included and the imputation strategy are listed in the tables below.

After imputation of lumens per catheter type, the total number of lumens has been calculated and kept as a feature and the number of lumens per catheter type features has been removed.

\subsubsection{missForestPredict imputation}\label{missforestpredict-imputation}

For the remaining features with missing values, missing data imputation has been performed on each training set using the missForestPredict algorithm (Albu 2023). Complete features have been included in imputation as ``predictors'' for the features with missing values. The missing values are first imputed with mean/mode and then iteratively imputed using random forest models for 5 iterations. The default hyperparameter values of the ranger function are used for random forest imputation models except the maximum tree depth which has been set to 10. The outcome is not included in the imputation. Patient length and weight missing data indicators have been preserved before imputation; after imputation, the BMI has been calculated and the patient length and weight have been deleted.

Test sets have been imputed using the missForestPredict imputation models learned on training data.

\begin{figure}
\centering
\includegraphics{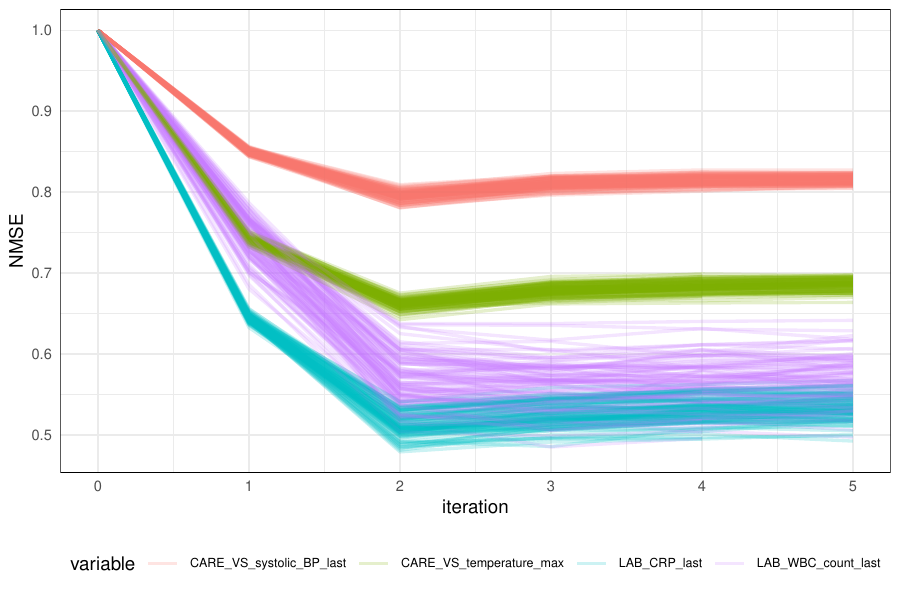}
\caption{\label{fig:MF-errors-base}OOB NMSE for baseline imputation (features included in the model) for all train sets}
\end{figure}

\begin{figure}
\centering
\includegraphics{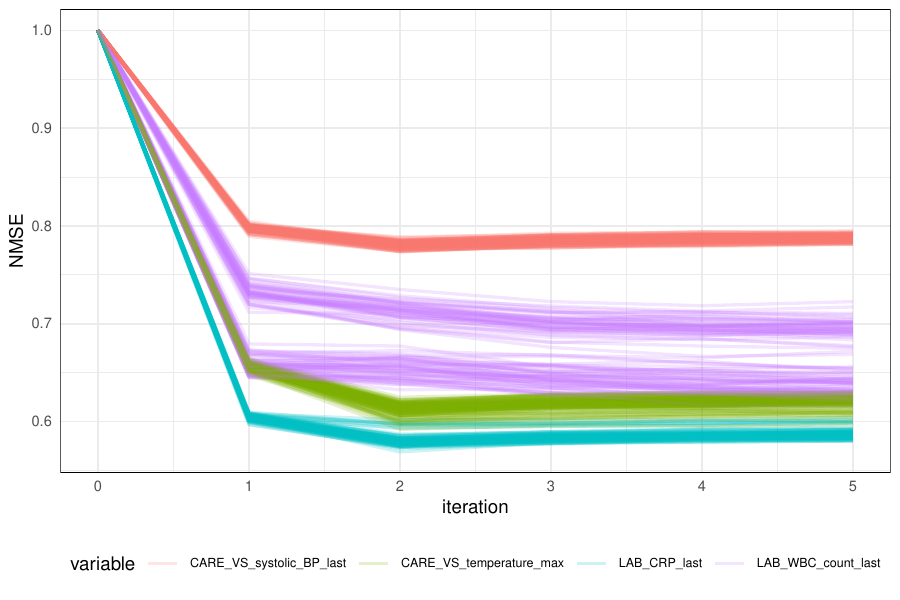}
\caption{\label{fig:MF-errors-dyn}OOB NMSE for dynamic imputation (features included in the model) for all train sets}
\end{figure}

All features included in the imputation for the baseline dataset are presented in Table \ref{tab:table-miss-base-excluded}. All features included in the imputation for the dynamic dataset are presented in Table \ref{tab:table-miss-dyn-excluded}. The OOB errors (NMSE) for each iteration are presented in Figures \ref{fig:MF-errors-base} and \ref{fig:MF-errors-dyn} Each line represents the NMSE error over iterations for one of the 100 training sets.

\begin{table}

\caption{\label{tab:table-miss-base-excluded}Features and imputation method (baseline)}
\centering
\resizebox{\linewidth}{!}{
\begin{tabular}[t]{llrrl>{\raggedright\arraybackslash}p{8em}>{\raggedright\arraybackslash}p{8em}}
\toprule
Feature short name & Feature & Number present & Number missing & Percentage missing & Imputation strategy & Rationale\\
\midrule
 & CAT\_lumens\_PICC & 30800 & 62 & 0.2\% & 1 (fixed value) & Lumens special situation\\
 & CAT\_lumens\_Tunneled\_CVC & 30775 & 87 & 0.3\% & 3 (fixed value) & Lumens special situation\\
 & MS\_medical\_specialty & 30701 & 161 & 0.5\% & mode & less than 3\% missing\\
 & MS\_alternative\_flag & 30701 & 161 & 0.5\% & mode & less than 3\% missing\\
 & ADM\_admission\_referral\_binary\_all\_GP & 30379 & 483 & 1.6\% & mode & less than 3\% missing\\
\addlinespace
Admission source Home & ADM\_admission\_source\_binary\_all\_Home & 30365 & 497 & 1.6\% & mode & less than 3\% missing\\
 & ADM\_admission\_reason\_binary\_all\_Accident & 30301 & 561 & 1.8\% & mode & less than 3\% missing\\
 & ADM\_admission\_type\_binary\_all\_Emergency & 30245 & 617 & 2\% & mode & less than 3\% missing\\
 & CAT\_lumens\_CVC & 25841 & 5021 & 16.3\% & 2 (fixed value) & Lumens special situation\\
Temperature & CARE\_VS\_temperature\_max & 22919 & 7943 & 25.7\% & missForestPredict & \\
\addlinespace
 & CARE\_VS\_heart\_rate\_max & 22851 & 8011 & 26\% & missForestPredict & \\
Systolic BP & CARE\_VS\_systolic\_BP\_last & 22616 & 8246 & 26.7\% & missForestPredict & \\
 & CARE\_SAF\_mobility\_assistance\_binary\_all\_partial\_help & 21782 & 9080 & 29.4\% & missForestPredict & \\
 & CARE\_SAF\_mobility\_assistance\_binary\_all\_no\_help & 21782 & 9080 & 29.4\% & missForestPredict & \\
 & CARE\_SAF\_mobility\_assistance\_binary\_all\_full\_help & 21782 & 9080 & 29.4\% & missForestPredict & \\
\addlinespace
 & LAB\_is\_neutropenia & 18032 & 12830 & 41.6\% & missForestPredict & \\
 & LAB\_Hemoglobine\_last & 18010 & 12852 & 41.6\% & missForestPredict & \\
WBC & LAB\_WBC\_count\_last & 17990 & 12872 & 41.7\% & missForestPredict & \\
 & LAB\_Platelet\_count\_last & 17948 & 12914 & 41.8\% & missForestPredict & \\
 & LAB\_RBC\_count\_last & 17865 & 12997 & 42.1\% & missForestPredict & \\
\addlinespace
 & LAB\_haematocrit\_last & 17851 & 13011 & 42.2\% & missForestPredict & \\
 & LAB\_potassium\_last & 17409 & 13453 & 43.6\% & missForestPredict & \\
 & LAB\_natrium\_last & 17401 & 13461 & 43.6\% & missForestPredict & \\
 & LAB\_creatinine\_last & 17319 & 13543 & 43.9\% & missForestPredict & \\
 & LAB\_urea\_last & 17291 & 13571 & 44\% & missForestPredict & \\
\addlinespace
CRP & LAB\_CRP\_last & 16480 & 14382 & 46.6\% & missForestPredict & \\
 & CAT\_bandage\_type\_binary\_last\_polyurethane & 16469 & 14393 & 46.6\% & missForestPredict & \\
 & CAT\_bandage\_type\_binary\_last\_gauze & 16469 & 14393 & 46.6\% & missForestPredict & \\
 & CAT\_needle\_length\_max & 13747 & 17115 & 55.5\% & mode & Difficult to impute\\
 & LAB\_PT\_sec\_last & 13668 & 17194 & 55.7\% & missForestPredict & \\
\addlinespace
 & LAB\_PT\_INR\_last & 13664 & 17198 & 55.7\% & missForestPredict & \\
 & LAB\_PT\_percent\_last & 13663 & 17199 & 55.7\% & missForestPredict & \\
 & LAB\_APTT\_last & 11038 & 19824 & 64.2\% & missForestPredict & \\
 & LAB\_O2\_saturation\_last & 10709 & 20153 & 65.3\% & missForestPredict & \\
 & LAB\_pO2\_last & 10703 & 20159 & 65.3\% & missForestPredict & \\
\addlinespace
 & LAB\_pH\_last & 10695 & 20167 & 65.3\% & missForestPredict & \\
 & LAB\_glucose\_arterial\_last & 10621 & 20241 & 65.6\% & missForestPredict & \\
 & CARE\_PHY\_weight\_mean & 10498 & 20364 & 66\% & missForestPredict & \\
 & LAB\_AST\_last & 10412 & 20450 & 66.3\% & missForestPredict & \\
 & LAB\_ALT\_last & 10372 & 20490 & 66.4\% & missForestPredict & \\
\addlinespace
 & LAB\_WBC\_Neutrophils\_last & 10282 & 20580 & 66.7\% & missForestPredict & \\
 & LAB\_WBC\_Monocytes\_last & 10254 & 20608 & 66.8\% & missForestPredict & \\
 & LAB\_bilirubin\_last & 10000 & 20862 & 67.6\% & missForestPredict & \\
 & LAB\_glucose\_last & 9972 & 20890 & 67.7\% & missForestPredict & \\
 & CAT\_bandage\_observation\_binary\_all\_Abnormal & 9185 & 21677 & 70.2\% & missForestPredict & \\
\addlinespace
 & CARE\_VS\_oxygen\_saturation\_last & 8954 & 21908 & 71\% & missForestPredict & \\
 & LAB\_LDH\_last & 8729 & 22133 & 71.7\% & missForestPredict & \\
 & CARE\_WND\_wound\_type\_binary\_all\_open\_wound & 5557 & 25305 & 82\% & missForestPredict & \\
 & CARE\_WND\_wound\_type\_binary\_all\_closed\_wound & 5557 & 25305 & 82\% & missForestPredict & \\
 & CARE\_WND\_wound\_type\_binary\_all\_suture\_and\_post & 5557 & 25305 & 82\% & missForestPredict & \\
\addlinespace
 & CARE\_PHY\_length\_mean & 3923 & 26939 & 87.3\% & missForestPredict & \\
 & CARE\_SAF\_patient\_position\_binary\_all\_Fowler & 3809 & 27053 & 87.7\% & missForestPredict & \\
 & CARE\_SAF\_patient\_position\_binary\_all\_lateral & 3809 & 27053 & 87.7\% & missForestPredict & \\
 & CARE\_SAF\_patient\_position\_binary\_all\_supine & 3809 & 27053 & 87.7\% & missForestPredict & \\
 & CARE\_SAF\_patient\_position\_binary\_all\_sitting & 3809 & 27053 & 87.7\% & missForestPredict & \\
\addlinespace
 & LAB\_CK\_last & 3661 & 27201 & 88.1\% & 112 for males, 88 for females (“normal” values) & Difficult to impute\\
 & CARE\_VS\_respiratory\_rate\_last & 3447 & 27415 & 88.8\% & missForestPredict & \\
 & LAB\_fibrinogen\_last & 2057 & 28805 & 93.3\% & missForestPredict & \\
 & CARE\_NEU\_GCS\_score\_last & 2028 & 28834 & 93.4\% & missForestPredict & \\
 & LAB\_TSH\_last & 1159 & 29703 & 96.2\% & 2.5 (“normal” value) & Difficult to impute\\
\addlinespace
 & LAB\_ferritin\_last & 1069 & 29793 & 96.5\% & missForestPredict & \\
 & LAB\_D\_dimer\_last & 791 & 30071 & 97.4\% & missForestPredict & \\
 & CARE\_VS\_CVP\_last & 710 & 30152 & 97.7\% & 8 (“normal” value) & Difficult to impute\\
\bottomrule
\end{tabular}}
\end{table}

\begin{table}

\caption{\label{tab:table-miss-dyn-excluded}Features and imputation method (dynamic)}
\centering
\resizebox{\linewidth}{!}{
\begin{tabular}[t]{llrrlll}
\toprule
Feature short name & Feature & Number present & Number missing & Percentage missing & Imputation strategy & Rationale\\
\midrule
 & MS\_medical\_specialty & 227736 & 192 & 0.08\% & mode & less than 3\% missing\\
 & MS\_alternative\_flag & 227736 & 192 & 0.08\% & mode & less than 3\% missing\\
 & CAT\_lumens\_PICC & 226439 & 1489 & 0.65\% & 1 (fixed value) & Lumens special situation\\
 & CAT\_lumens\_Tunneled\_CVC & 226419 & 1509 & 0.66\% & 3 (fixed value) & Lumens special situation\\
 & ADM\_admission\_referral\_binary\_all\_GP & 226371 & 1557 & 0.68\% & mode & less than 3\% missing\\
\addlinespace
Admission source Home & ADM\_admission\_source\_binary\_all\_Home & 226244 & 1684 & 0.74\% & mode & less than 3\% missing\\
 & ADM\_admission\_reason\_binary\_all\_Accident & 225772 & 2156 & 0.95\% & mode & less than 3\% missing\\
 & ADM\_admission\_type\_binary\_all\_Emergency & 225046 & 2882 & 1.26\% & mode & less than 3\% missing\\
Temperature & CARE\_VS\_temperature\_max & 215107 & 12821 & 5.63\% & missForestPredict & \\
 & CARE\_VS\_heart\_rate\_max & 207451 & 20477 & 8.98\% & missForestPredict & \\
\addlinespace
Systolic BP & CARE\_VS\_systolic\_BP\_last & 205661 & 22267 & 9.77\% & missForestPredict & \\
 & CAT\_lumens\_CVC & 181021 & 46907 & 20.58\% & 2 (fixed value) & Lumens special situation\\
 & CARE\_SAF\_mobility\_assistance\_binary\_all\_partial\_help & 173326 & 54602 & 23.96\% & missForestPredict & \\
 & CARE\_SAF\_mobility\_assistance\_binary\_all\_no\_help & 173326 & 54602 & 23.96\% & missForestPredict & \\
 & CARE\_SAF\_mobility\_assistance\_binary\_all\_full\_help & 173326 & 54602 & 23.96\% & missForestPredict & \\
\addlinespace
 & CAT\_bandage\_type\_binary\_last\_polyurethane & 170157 & 57771 & 25.35\% & missForestPredict & \\
 & CAT\_bandage\_type\_binary\_last\_gauze & 170157 & 57771 & 25.35\% & missForestPredict & \\
 & CAT\_bandage\_observation\_binary\_all\_Normal & 166577 & 61351 & 26.92\% & missForestPredict & \\
 & CAT\_bandage\_observation\_binary\_all\_Bloody\_or\_Moist & 166577 & 61351 & 26.92\% & missForestPredict & \\
 & CAT\_bandage\_observation\_binary\_all\_Red & 166577 & 61351 & 26.92\% & missForestPredict & \\
\addlinespace
 & CAT\_bandage\_observation\_binary\_all\_Other\_Hema\_Pus\_Loose\_Necro & 166577 & 61351 & 26.92\% & missForestPredict & \\
 & LAB\_is\_neutropenia & 138216 & 89712 & 39.36\% & missForestPredict & \\
 & LAB\_Hemoglobine\_last & 137959 & 89969 & 39.47\% & missForestPredict & \\
WBC & LAB\_WBC\_count\_last & 137581 & 90347 & 39.64\% & missForestPredict & \\
 & LAB\_potassium\_last & 137361 & 90567 & 39.73\% & missForestPredict & \\
\addlinespace
 & LAB\_natrium\_last & 137200 & 90728 & 39.81\% & missForestPredict & \\
 & LAB\_Platelet\_count\_last & 136846 & 91082 & 39.96\% & missForestPredict & \\
 & LAB\_creatinine\_last & 136062 & 91866 & 40.3\% & missForestPredict & \\
 & LAB\_urea\_last & 135748 & 92180 & 40.44\% & missForestPredict & \\
 & LAB\_RBC\_count\_last & 135566 & 92362 & 40.52\% & missForestPredict & \\
\addlinespace
 & LAB\_haematocrit\_last & 135330 & 92598 & 40.63\% & missForestPredict & \\
CRP & LAB\_CRP\_last & 135015 & 92913 & 40.76\% & missForestPredict & \\
 & CARE\_WND\_wound\_type\_binary\_all\_suture & 107641 & 120287 & 52.77\% & missForestPredict & \\
 & CARE\_WND\_wound\_type\_binary\_all\_open\_wound & 107641 & 120287 & 52.77\% & missForestPredict & \\
 & CARE\_WND\_wound\_type\_binary\_all\_post\_suture & 107641 & 120287 & 52.77\% & missForestPredict & \\
\addlinespace
 & CARE\_WND\_wound\_type\_binary\_all\_closed\_wound & 107641 & 120287 & 52.77\% & missForestPredict & \\
 & CARE\_VS\_oxygen\_saturation\_last & 103740 & 124188 & 54.49\% & missForestPredict & \\
 & LAB\_bilirubin\_last & 97977 & 129951 & 57.01\% & missForestPredict & \\
 & LAB\_AST\_last & 81969 & 145959 & 64.04\% & missForestPredict & \\
 & LAB\_ALT\_last & 81759 & 146169 & 64.13\% & missForestPredict & \\
\addlinespace
 & LAB\_PT\_sec\_last & 77964 & 149964 & 65.79\% & missForestPredict & \\
 & LAB\_PT\_INR\_last & 77948 & 149980 & 65.8\% & missForestPredict & \\
 & LAB\_PT\_percent\_last & 77946 & 149982 & 65.8\% & missForestPredict & \\
 & LAB\_LDH\_last & 73269 & 154659 & 67.85\% & missForestPredict & \\
 & CAT\_needle\_length\_max & 68161 & 159767 & 70.1\% & missForestPredict & \\
\addlinespace
 & CARE\_SAF\_patient\_position\_binary\_all\_Fowler & 67789 & 160139 & 70.26\% & missForestPredict & \\
 & CARE\_SAF\_patient\_position\_binary\_all\_lateral & 67789 & 160139 & 70.26\% & missForestPredict & \\
 & CARE\_SAF\_patient\_position\_binary\_all\_supine & 67789 & 160139 & 70.26\% & missForestPredict & \\
 & CARE\_SAF\_patient\_position\_binary\_all\_sitting & 67789 & 160139 & 70.26\% & missForestPredict & \\
 & LAB\_WBC\_Neutrophils\_last & 64666 & 163262 & 71.63\% & missForestPredict & \\
\addlinespace
 & LAB\_WBC\_Monocytes\_last & 64165 & 163763 & 71.85\% & missForestPredict & \\
 & LAB\_APTT\_last & 63028 & 164900 & 72.35\% & missForestPredict & \\
 & CARE\_PHY\_weight\_mean & 56798 & 171130 & 75.08\% & missForestPredict & \\
 & CARE\_VS\_respiratory\_rate\_last & 46708 & 181220 & 79.51\% & missForestPredict & \\
 & LAB\_pO2\_last & 46426 & 181502 & 79.63\% & missForestPredict & \\
\addlinespace
 & LAB\_O2\_saturation\_last & 46396 & 181532 & 79.64\% & missForestPredict & \\
 & LAB\_pH\_last & 46385 & 181543 & 79.65\% & missForestPredict & \\
 & LAB\_glucose\_last & 45920 & 182008 & 79.85\% & missForestPredict & \\
 & LAB\_glucose\_arterial\_last & 45875 & 182053 & 79.87\% & missForestPredict & \\
 & CAT\_lumens\_flushed & 41384 & 186544 & 81.84\% & missForestPredict & \\
\addlinespace
 & CARE\_NEU\_GCS\_score\_last & 39412 & 188516 & 82.71\% & missForestPredict & \\
 & CARE\_VS\_CVP\_last & 31985 & 195943 & 85.97\% & missForestPredict & \\
 & LAB\_CK\_last & 22477 & 205451 & 90.14\% & missForestPredict & \\
 & LAB\_vancomycine\_last & 8912 & 219016 & 96.09\% & missForestPredict & \\
 & CAT\_result\_infusion\_binary\_all\_normal & 7565 & 220363 & 96.68\% & missForestPredict & \\
\addlinespace
 & CAT\_result\_infusion\_binary\_all\_difficult & 7565 & 220363 & 96.68\% & missForestPredict & \\
 & CAT\_result\_infusion\_binary\_all\_impossible & 7565 & 220363 & 96.68\% & missForestPredict & \\
 & CAT\_result\_aspiration\_binary\_all\_normal & 7564 & 220364 & 96.68\% & missForestPredict & \\
 & CAT\_result\_aspiration\_binary\_all\_difficult & 7564 & 220364 & 96.68\% & missForestPredict & \\
 & CAT\_result\_aspiration\_binary\_all\_impossible & 7564 & 220364 & 96.68\% & missForestPredict & \\
\addlinespace
 & CARE\_PHY\_length\_mean & 6846 & 221082 & 97\% & missForestPredict & \\
 & LAB\_fibrinogen\_last & 6837 & 221091 & 97\% & missForestPredict & \\
 & LAB\_aspergillus\_ag\_last & 5450 & 222478 & 97.61\% & missForestPredict & \\
 & LAB\_ferritin\_last & 3969 & 223959 & 98.26\% & missForestPredict & \\
 & LAB\_TSH\_last & 3689 & 224239 & 98.38\% & missForestPredict & \\
\addlinespace
 & LAB\_creatinine\_clearance\_last & 2917 & 225011 & 98.72\% & missForestPredict & \\
 & LAB\_SPE\_albumin\_last & 2665 & 225263 & 98.83\% & missForestPredict & \\
 & LAB\_SPE\_albumin\_alpha\_1\_globulin\_last & 2665 & 225263 & 98.83\% & missForestPredict & \\
 & LAB\_SPE\_albumin\_alpha\_2\_globulin\_last & 2665 & 225263 & 98.83\% & missForestPredict & \\
 & LAB\_SPE\_albumin\_beta\_globulin\_last & 2665 & 225263 & 98.83\% & missForestPredict & \\
\addlinespace
 & LAB\_SPE\_albumin\_gamma\_globulin\_last & 2665 & 225263 & 98.83\% & missForestPredict & \\
 & LAB\_ciclosporin\_last & 2409 & 225519 & 98.94\% & missForestPredict & \\
 & LAB\_D\_dimer\_last & 2009 & 225919 & 99.12\% & missForestPredict & \\
\bottomrule
\end{tabular}}
\end{table}

\subsection{Supplementary material 5 - Comparison to ranger model}\label{suppl-compare-ranger}

\begin{figure}
\centering
\includegraphics{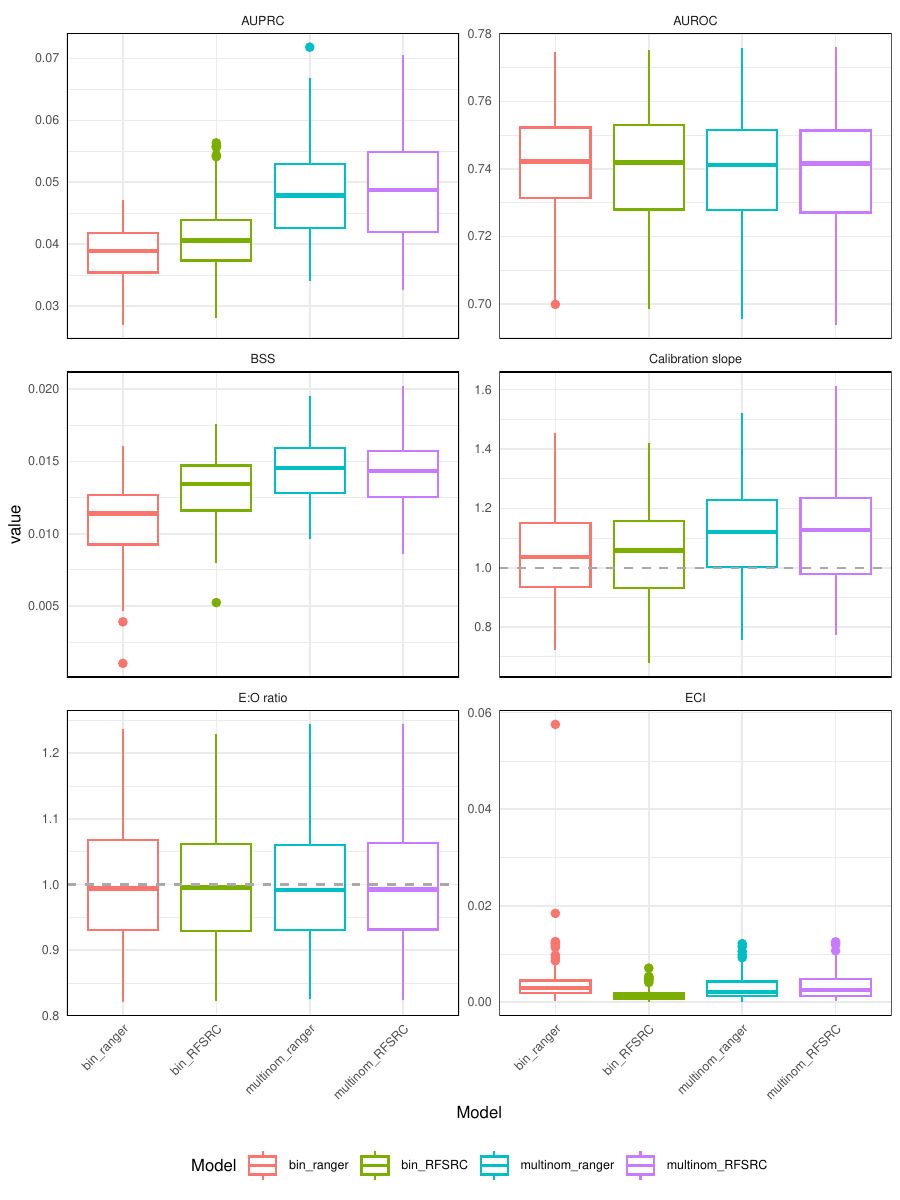}
\caption{\label{fig:baseline-cutoff-independent-suppl}Prediction performance for baseline models (ranger vs.~RFSRC)}
\end{figure}

\begin{figure}
\centering
\includegraphics{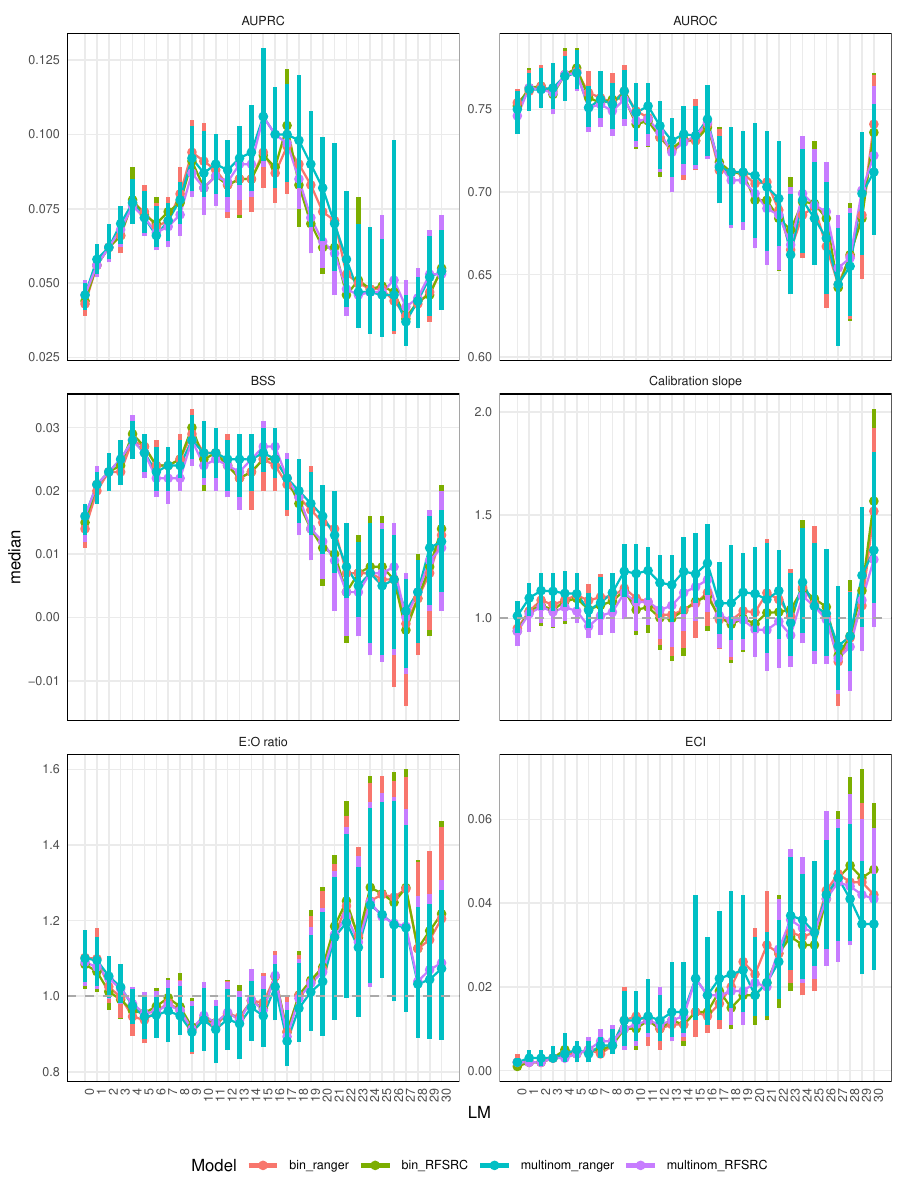}
\caption{\label{fig:t-dynamic-cutoff-independent-suppl}Prediction performance for dynamic models - time dependent metrics (ranger vs.~RFSRC)}
\end{figure}

We have investigated the potential bias introduced by allowing a small number of observations belonging to the same admission to fall both in-bag and out-of-bag, incentivizing the tuning strategy to lean towards models that would memorize such admissions. This procedure affects on average (over all dynamic train sets and all inbags) between 1.1\% and 1.7\% of the inbag observations depending on the subsample size. We have performed a comparison for the static and dynamic binary and multinomial models against models built using the ranger package (Wright et al. 2019), which does not impose the limitation of equal size inbags, using the same inbags before adjusting for minimum size and we see no noticeable difference in results: Figure \ref{fig:baseline-cutoff-independent-suppl} for baseline models and Figure \ref{fig:t-dynamic-cutoff-independent-suppl} for dynamic models. The models are evaluated on the test sets following the same procedure as in the main results. Given the small extent to which bias might potentially leak in and considering that all models (binary, multinomial, survival, CR) would suffer from the same bias, we will consider our comparisons valid under this limitation.

\subsection{Supplementary material 6 - Additional documentation for methods}\label{suppl-methods}

\subsubsection{Model tuning}\label{model-tuning}

The tuned hyperparameters are the number of variables selected at each split (mtry) and the minimum size of a terminal node (nodesize) for both baseline and dynamic models. Additionally for dynamic models, the subsample size is tuned. The candidate tuned values of the hyperaparameters are limited to a min-max range: 2 to 15 for mtry, 50 to 4000 for minimum node size and 30\% to 80\% for the subsample size. The tuning is based on the out-of-bag binary logloss for next 7 days. Model based optimization tuning is performed using the mlrMBO R package (Bischl et al. 2017) with 20 design steps (randomly selected hyperparameter value combinations) and 30 optimization steps (aimed to find better hyperparameter values that minimze the logloss). An initial ``design'' is generated by randomly picking 20 points in the hyperparameter space (within the min-max range for each hyperparameter). Models are fit using the ``design'' hyperparameter values and evaluated using the evaluation metric (logloss). We have used the out-of-bag logloss for the binary outcome (CLABSI within 7 days) as evaluation metric for all models, as we prefer to tune the models against the outcome of interest (instead of using survival metrics, for example). A ``surrogate function'' is fit through these design points modelling the evaluation metric (logloss) in function of the hyperparameters (mtry, node size, subsample size). A Kriging model is used for the ``surrogate function'' which performs well for continuous hyperparameters (Roustant, Ginsbourger, and Deville 2012). Further, 30 ``optimization'' steps are performed. At each step, new points in the hyperparameter space are proposed with either expected good model performance or with potential to improve the ``surrogate function'' (``trade-off exploitation and exploration'', Bischl et al. (2017)). Finally, the best hyperparameters from the 50 steps are chosen and a random forest model is built using these hyperparameters.

\subsubsection{Linear analogues of RF models}\label{linear-analogues-of-rf-models}

A correspondence between the built RF models and linear models have been attempted in Table \ref{tab:table-models-2}. The correspondence is established based on the outcome type and split rule of the RF and it is possibly not a perfect equivalence.

\begin{table}

\caption{\label{tab:table-models-2}Models}
\centering
\resizebox{\linewidth}{!}{
\begin{tabular}[t]{lll>{\raggedright\arraybackslash}p{16em}>{\raggedright\arraybackslash}p{16em}ll}
\toprule
Model name & Outcome type & Splitrule & Other hyperparameters / options & Linear model analogue & Baseline model & Dynamic model\\
\midrule
bin & Binary & gini &  & Logistic regression with binary outcome & YES & YES\\
multinom & Multinomial & gini &  & Multinomial logistic regression & YES & YES\\
surv7d & Survival & logrank & Administrative censoring at day 7; Censoring death and discharge at event time & Cox proportional hazards with censoring for competing events at their event time and administrative censoring at day 7 & YES & YES\\
surv7d\_cens7 & Survival & logrank & Administrative censoring at day 7; Censoring death and discharge at time 7 & Cox proportional hazards with censoring for competing events at day 7 and administrative censoring at day 7; also analogue to Fine-Gray subdistribution hazard model with administrative censoring at day 7 (in the presence of no other censoring) & YES & YES\\
surv30d & Survival & logrank & Administrative censoring at day 30; Censoring death and discharge at event time & Cox proportional hazards with censoring for competing events at their event time and administrative censoring at day 30 & YES & NO\\
\addlinespace
surv30d\_cens7 & Survival & logrank & Administrative censoring at day 30; Censoring death and discharge at time 7 & Cox proportional hazards with censoring for competing events at day 7 and administrative censoring at day 30; also analogue to Fine-Gray subdistribution hazard model with administrative censoring at day 30 (in the presence of no other censoring) & YES & NO\\
CR7d\_LRCR\_c\_1 & Competing risks & logrankCR & cause = 1 (CLABSI); Administrative censoring at day 7 & Fine-Gray subdistribution hazard model with administrative censoring at day 7 & YES & YES\\
CR7d\_LR\_c\_1 & Competing risks & logrank & cause = 1 (CLABSI); Administrative censoring at day 7 & No analogue & YES & YES\\
CR7d\_LRCR\_c\_all & Competing risks & logrankCR & cause = default (all events have equal weights); Administrative censoring at day 7 & No analogue & YES & YES\\
CR7d\_LR\_c\_all & Competing risks & logrank & cause = default (all events have equal weights); Administrative censoring at day 7 & Cause-specific hazard regression model with administrative censoring at day 7 & YES & YES\\
\addlinespace
CR30d\_LRCR\_c\_1 & Competing risks & logrankCR & cause = 1 (CLABSI); Administrative censoring at day 30 & Fine-Gray subdistribution hazard model with administrative censoring at day 30 & YES & NO\\
CR30d\_LR\_c\_1 & Competing risks & logrank & cause = 1 (CLABSI); Administrative censoring at day 30 & No analogue & YES & NO\\
CR30d\_LRCR\_c\_all & Competing risks & logrankCR & cause = default (all events have equal weights); Administrative censoring at day 30 & No analogue & YES & NO\\
CR30d\_LR\_c\_all & Competing risks & logrank & cause = default (all events have equal weights); Administrative censoring at day 30 & Cause-specific hazard regression model with administrative censoring at day 30 & YES & NO\\
\bottomrule
\end{tabular}}
\end{table}

\subsection{Supplementary material 7 - Additional performance evaluation - baseline models}\label{suppl-baseline-perf}

\subsubsection{Baseline metrics table}\label{baseline-metrics-table}

The performance of baseline models is presented in Table \ref{tab:baseline-metrics-table}.

\begin{table}

\caption{\label{tab:baseline-metrics-table}Baseline metrics table}
\centering
\resizebox{\linewidth}{!}{
\begin{tabular}[t]{lllllll}
\toprule
Model & BSS & AUPRC & AUROC & Calibration slope & ECI & E:O ratio\\
\midrule
bin & 0.013 (0.012 - 0.015) & 0.041 (0.037 - 0.044) & 0.742 (0.728 - 0.753) & 1.058 (0.933 - 1.157) & 0.001 (0.001 - 0.002) & 0.995 (0.929 - 1.062)\\
multinom & 0.014 (0.013 - 0.016) & 0.049 (0.042 - 0.055) & 0.742 (0.727 - 0.751) & 1.128 (0.979 - 1.234) & 0.003 (0.001 - 0.005) & 0.993 (0.931 - 1.063)\\
surv7d & 0.01 (0.007 - 0.012) & 0.038 (0.036 - 0.043) & 0.729 (0.716 - 0.745) & 1.211 (1.115 - 1.35) & 0.005 (0.004 - 0.007) & 1.444 (1.353 - 1.541)\\
surv7d\_cens7 & 0.013 (0.012 - 0.015) & 0.041 (0.038 - 0.044) & 0.742 (0.729 - 0.753) & 1.04 (0.944 - 1.162) & 0.001 (0.001 - 0.002) & 0.996 (0.929 - 1.063)\\
surv30d & 0.009 (0.007 - 0.012) & 0.039 (0.037 - 0.043) & 0.724 (0.713 - 0.74) & 1.259 (1.14 - 1.387) & 0.005 (0.004 - 0.007) & 1.469 (1.375 - 1.565)\\
\addlinespace
surv30d\_cens7 & 0.014 (0.012 - 0.015) & 0.042 (0.039 - 0.047) & 0.739 (0.728 - 0.752) & 1.092 (0.99 - 1.169) & 0.001 (0.001 - 0.002) & 0.995 (0.93 - 1.06)\\
CR7d\_LR\_c\_1 & 0.013 (0.012 - 0.014) & 0.041 (0.037 - 0.044) & 0.739 (0.725 - 0.749) & 1.116 (0.994 - 1.234) & 0.001 (0.001 - 0.003) & 0.996 (0.934 - 1.061)\\
CR7d\_LRCR\_c\_1 & 0.013 (0.012 - 0.015) & 0.041 (0.037 - 0.045) & 0.742 (0.729 - 0.752) & 1.062 (0.947 - 1.157) & 0.001 (0.001 - 0.002) & 0.995 (0.929 - 1.062)\\
CR7d\_LR\_c\_all & 0.014 (0.012 - 0.015) & 0.047 (0.042 - 0.052) & 0.736 (0.724 - 0.75) & 1.133 (1.052 - 1.255) & 0.003 (0.001 - 0.005) & 0.991 (0.931 - 1.063)\\
CR7d\_LRCR\_c\_all & 0.014 (0.012 - 0.015) & 0.047 (0.042 - 0.054) & 0.738 (0.727 - 0.75) & 1.151 (1.04 - 1.242) & 0.002 (0.001 - 0.004) & 0.99 (0.93 - 1.062)\\
\addlinespace
CR30d\_LR\_c\_1 & 0.013 (0.011 - 0.015) & 0.042 (0.039 - 0.047) & 0.737 (0.725 - 0.751) & 1.111 (1.035 - 1.213) & 0.001 (0.001 - 0.003) & 0.995 (0.931 - 1.06)\\
CR30d\_LRCR\_c\_1 & 0.014 (0.012 - 0.015) & 0.043 (0.04 - 0.047) & 0.741 (0.729 - 0.752) & 1.076 (0.964 - 1.186) & 0.002 (0.001 - 0.003) & 0.995 (0.931 - 1.063)\\
CR30d\_LR\_c\_all & 0.014 (0.012 - 0.015) & 0.049 (0.043 - 0.055) & 0.735 (0.721 - 0.746) & 1.188 (1.062 - 1.289) & 0.003 (0.002 - 0.005) & 0.989 (0.932 - 1.062)\\
CR30d\_LRCR\_c\_all & 0.014 (0.013 - 0.016) & 0.047 (0.043 - 0.052) & 0.741 (0.728 - 0.75) & 1.097 (0.994 - 1.209) & 0.002 (0.001 - 0.004) & 0.991 (0.933 - 1.063)\\
\bottomrule
\end{tabular}}
\end{table}

\subsubsection{Comparison of baseline and dynamic models performance at baseline}\label{comparison-of-baseline-and-dynamic-models-performance-at-baseline}

\begin{figure}
\centering
\includegraphics{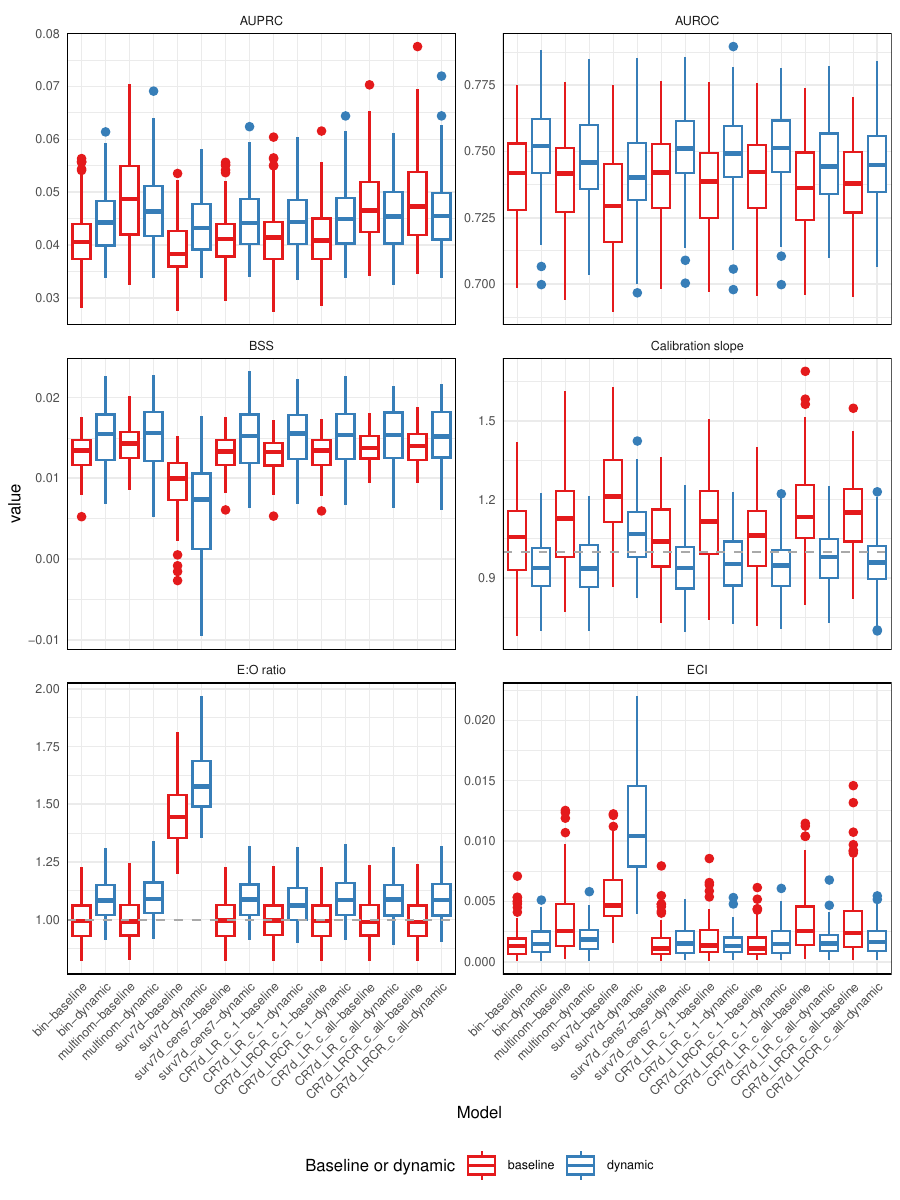}
\caption{\label{fig:baseline-dyn-compare}Prediction performance for baseline and dynamic models at baseline (LM 0)}
\end{figure}

Dynamic models make predictions at all landmarks, including at baseline (LM0). A comparison between baseline models and dynamic models evaluated at baseline is presented in Figure \ref{fig:baseline-dyn-compare}. Baseline models with administrative censoring at day 30 have been excluded as they do not have their dynamic counterpart. Dynamic models perform better at baseline in terms of discrimination (AUROC) and BSS, but worse in terms of calibration (E:O ratio, calibration slope and intercept and ECI).

\subsubsection{ROC curves}\label{roc-curves}

\begin{figure}
\centering
\includegraphics{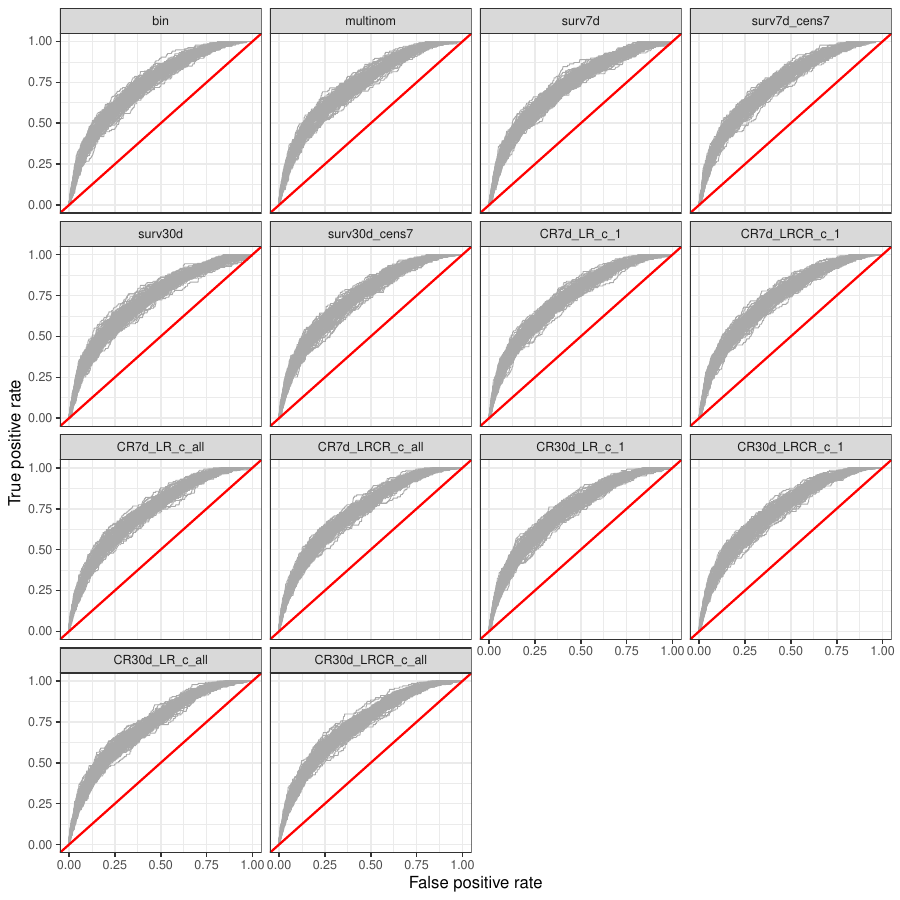}
\caption{\label{fig:ROC-curves-base}ROC curves for baseline models}
\end{figure}

ROC curves for each test set and each model are presented in Figure \ref{fig:ROC-curves-base}.

\subsubsection{Precision-recall curves}\label{precision-recall-curves}

\begin{figure}
\centering
\includegraphics{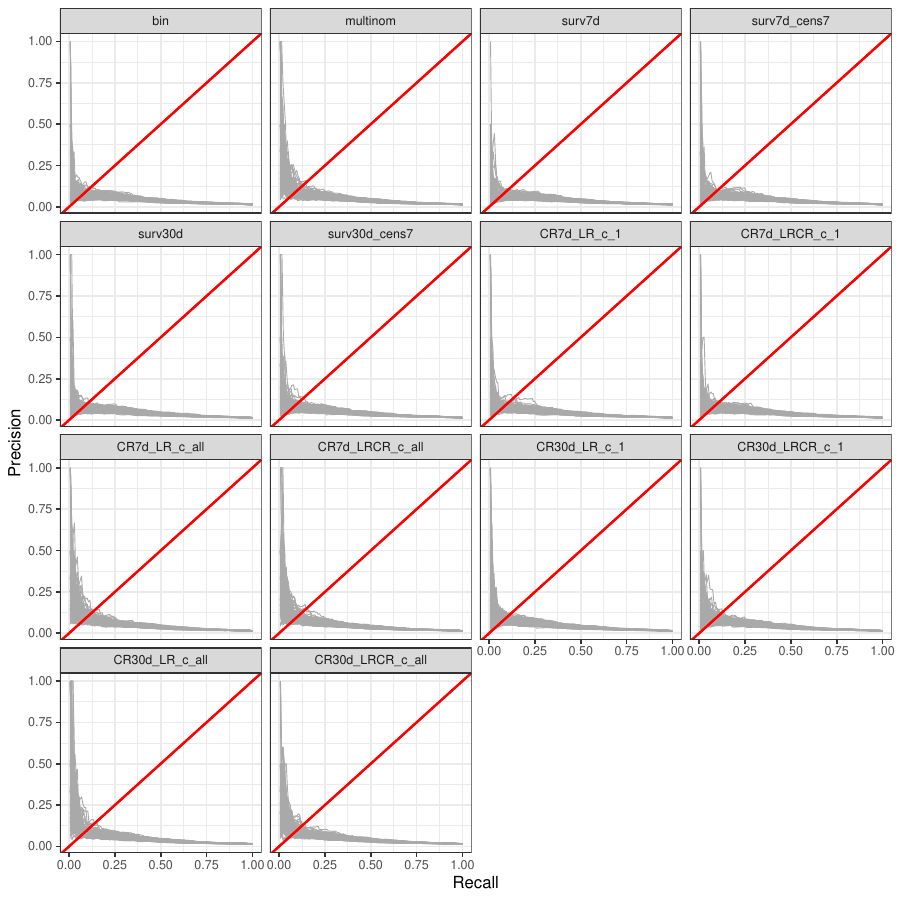}
\caption{\label{fig:PR-curves-base}Precision-recall curves for baseline models}
\end{figure}

Precision-recall curves for each test set and each model are presented in Figure \ref{fig:PR-curves-base}.

\subsubsection{Calibration curves - deciles}\label{calibration-curves---deciles}

\begin{figure}
\centering
\includegraphics{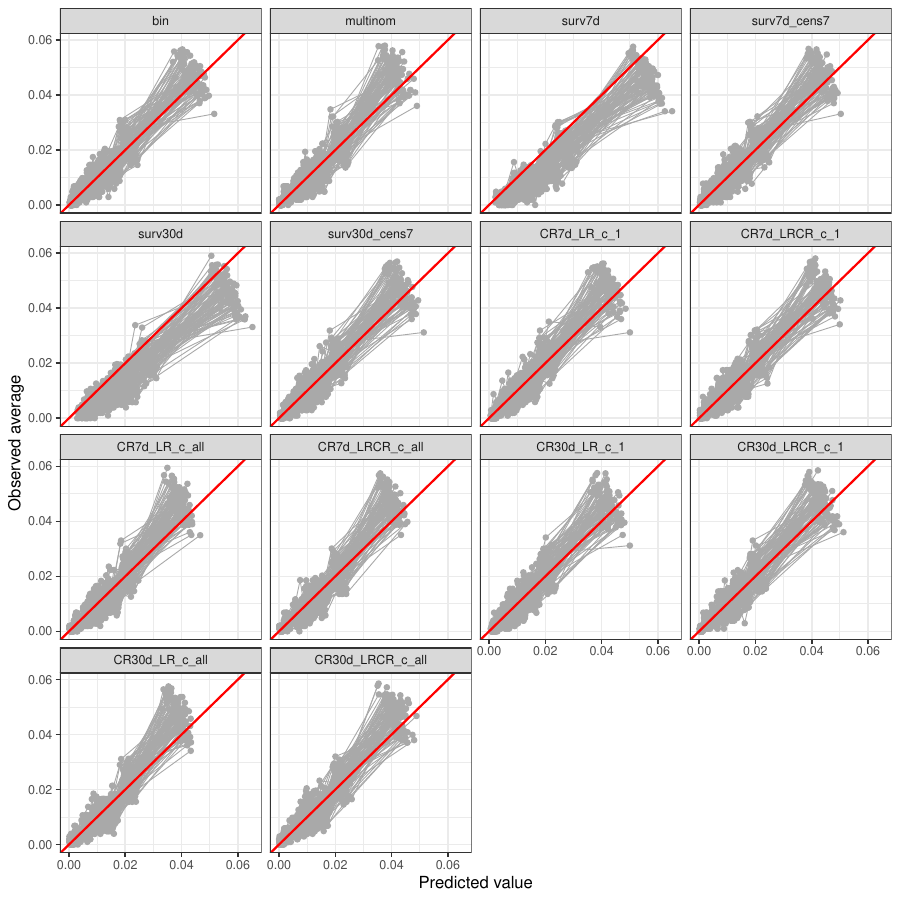}
\caption{\label{fig:calib-deciles-curves-base}Calibration curves (deciles) for baseline models}
\end{figure}

Using the deciles of the predicted 7 days risks on the test set, the test observations are grouped in ten groups. The average of the binary 7 day outcome within the groups is plotted against the average of predictions within the groups. Survival models with competing risks censored at the time of event (surv7d and surv30d) show overestimated predictions. The deciles calibration curves are presented in \ref{fig:calib-deciles-curves-base}.

\subsubsection{Calibration curves - splines}\label{calibration-curves---splines}

\begin{figure}
\centering
\includegraphics{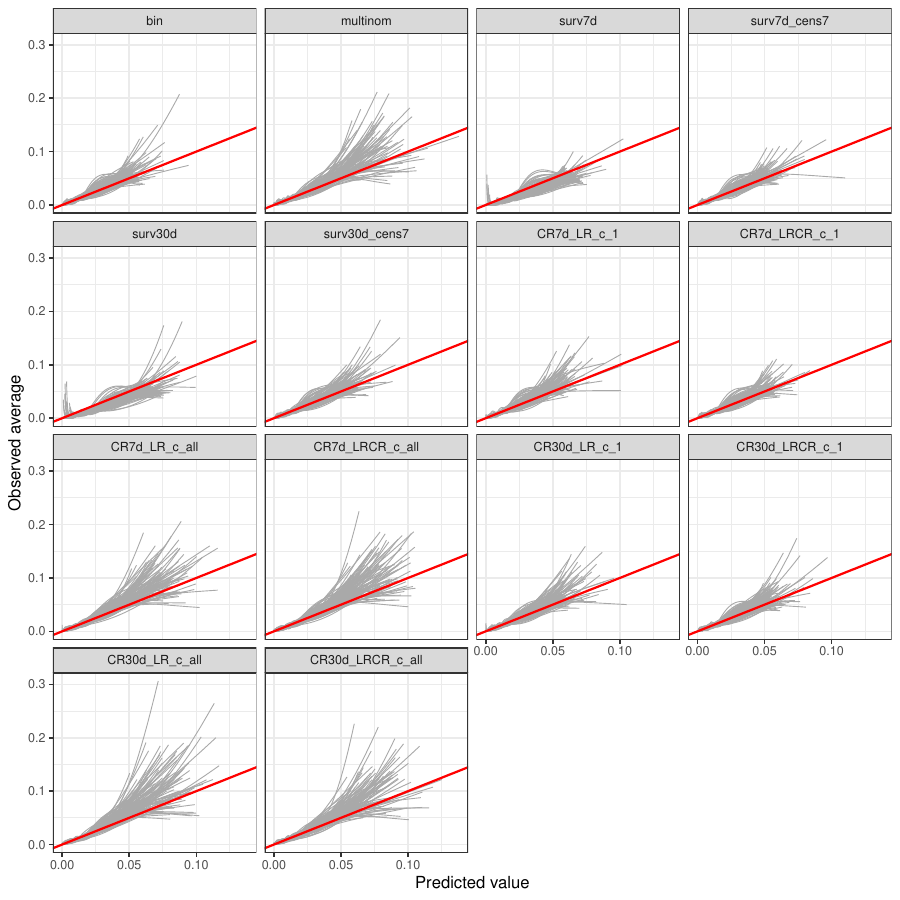}
\caption{\label{fig:calib-splines-curves-base}Calibration curves (splines) for baseline models}
\end{figure}

The binary outcome is regressed against natural cubic splines with 6 degrees of freedom of \(logit(model \ predictions)\) and the resulting predictions and plotted against \(expit(model \ predictions)\) to obtain calibration curves for each test set. Survival models with competing risks censored at the time of event (surv7d and surv30d) show overestimated predictions. The calibration curves based on cubic splines are presented in \ref{fig:calib-splines-curves-base}.

\subsubsection{Decision curves}\label{decision-curves}

\begin{figure}
\centering
\includegraphics{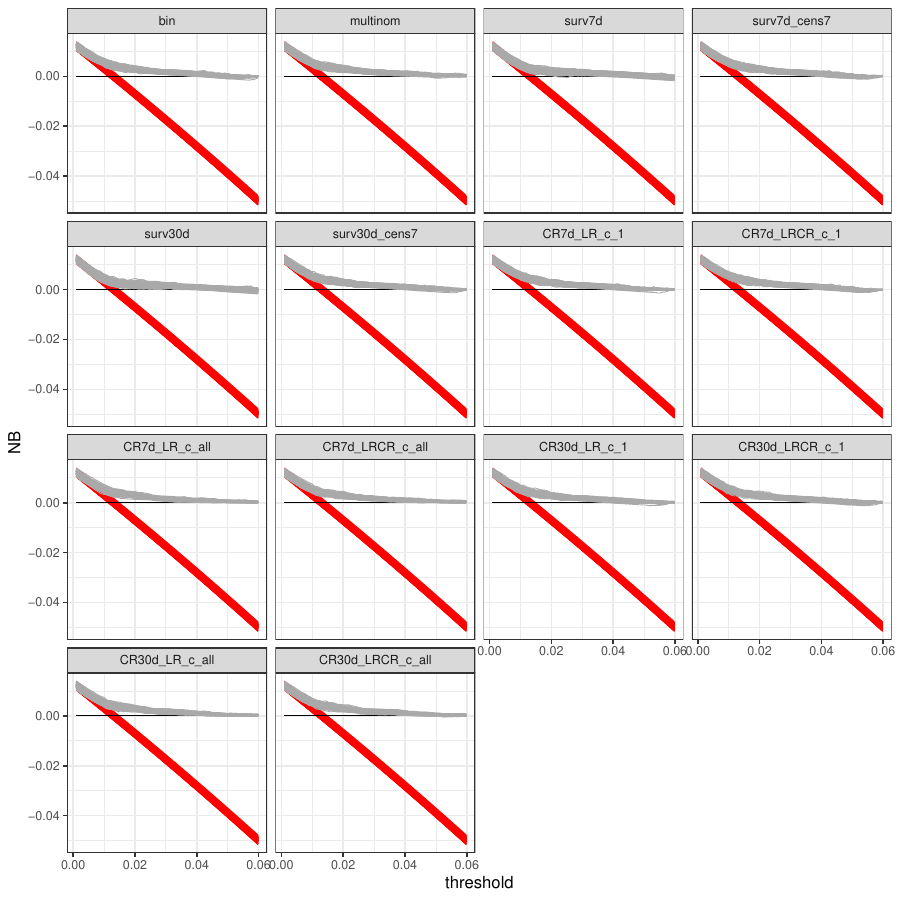}
\caption{\label{fig:decision-curves-base}Decision curves for baseline models}
\end{figure}

Decision analysis curves ((A. J. Vickers and Elkin 2006), (A. Vickers, Van Calster, and Steyerberg, n.d.)) are shown for prediction thresholds between 0 and 6\% for all test sets and all models. The net benefit of the model is plotted in grey, the net benefit for the decision to treat all in red and the net benefit for treat none in black. The decision curves are presented in Figure \ref{fig:decision-curves-base}.

\subsubsection{Predictions density curves}\label{predictions-density-curves}

\begin{figure}
\centering
\includegraphics{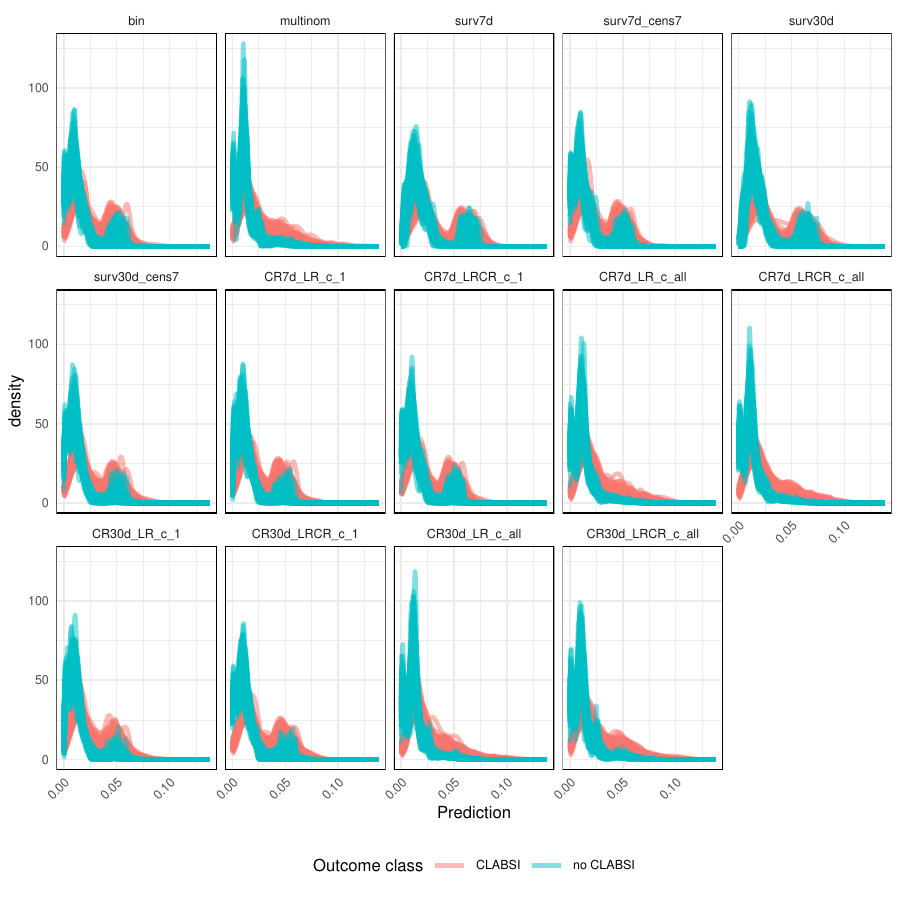}
\caption{\label{fig:pred-density-curves-base}Decision curves for baseline models}
\end{figure}

Prediction density curves for the positive class (CLABSI) and the negative class (no CLABSI) are shown in Figure \ref{fig:pred-density-curves-base} for the predictions on the test sets. Models that do not include death and discharge in the outcome definition (binary, survival and competing risks models with zero weights for death and discharge in the splitrule) display a bimodal distribution of the predicted risks.

\subsubsection{Tuned hyperparameters}\label{tuned-hyperparameters}

\begin{figure}
\centering
\includegraphics{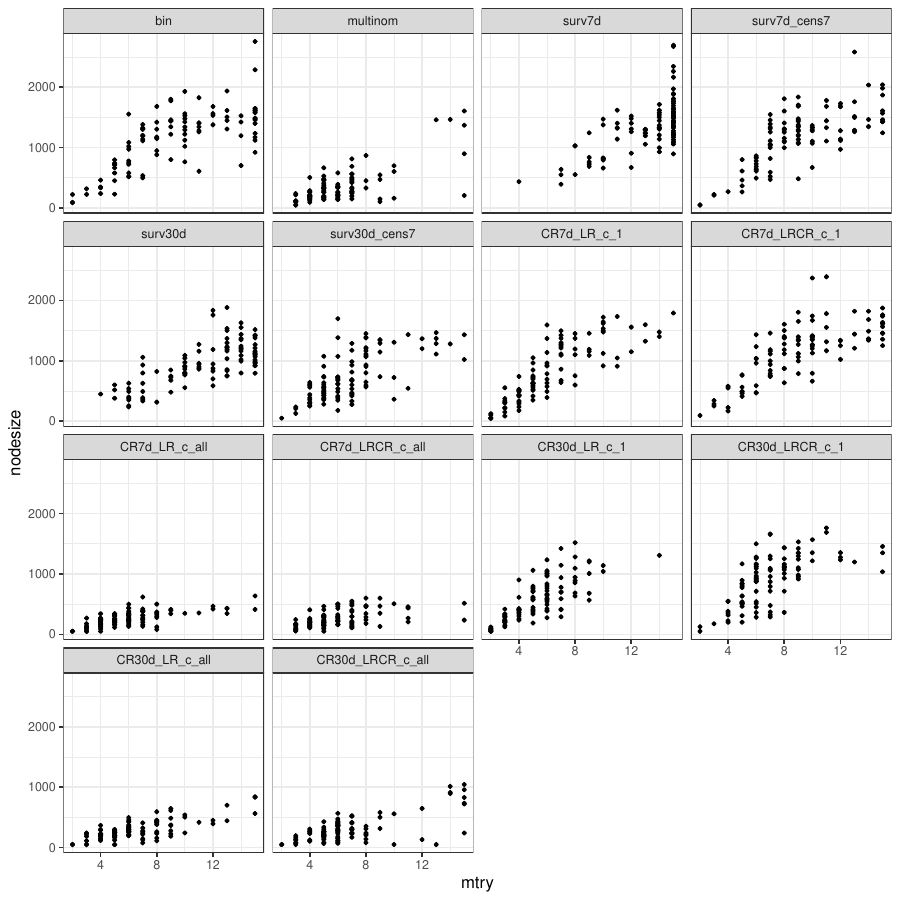}
\caption{\label{fig:hyperparams-base}Tuned hyperparameters}
\end{figure}

The tuned hyperparameters for the baseline models (mtry and nodesize) are shown in Figure \ref{fig:hyperparams-base}. The models using all events in the outcome definition (multinomial and CR models weighting all causes) have tuned nodesizes lower than the other models, indicating a tendency to use more splits in building the trees to achieve the best results for minimizing the binary logloss (all models are tuned for the same metric, based on the binary outcome). As each split is optimizing a loss function over multiple levels of the outcome, it might become less efficient to optimize for the binary outcome, therefore requiring more splits. These models also display a slightly larger ECI; the slight miscalibration might be due to the fact that less observations are left in the final nodes.

\subsubsection{Variable importance}\label{variable-importance}

\begin{figure}
\centering
\includegraphics{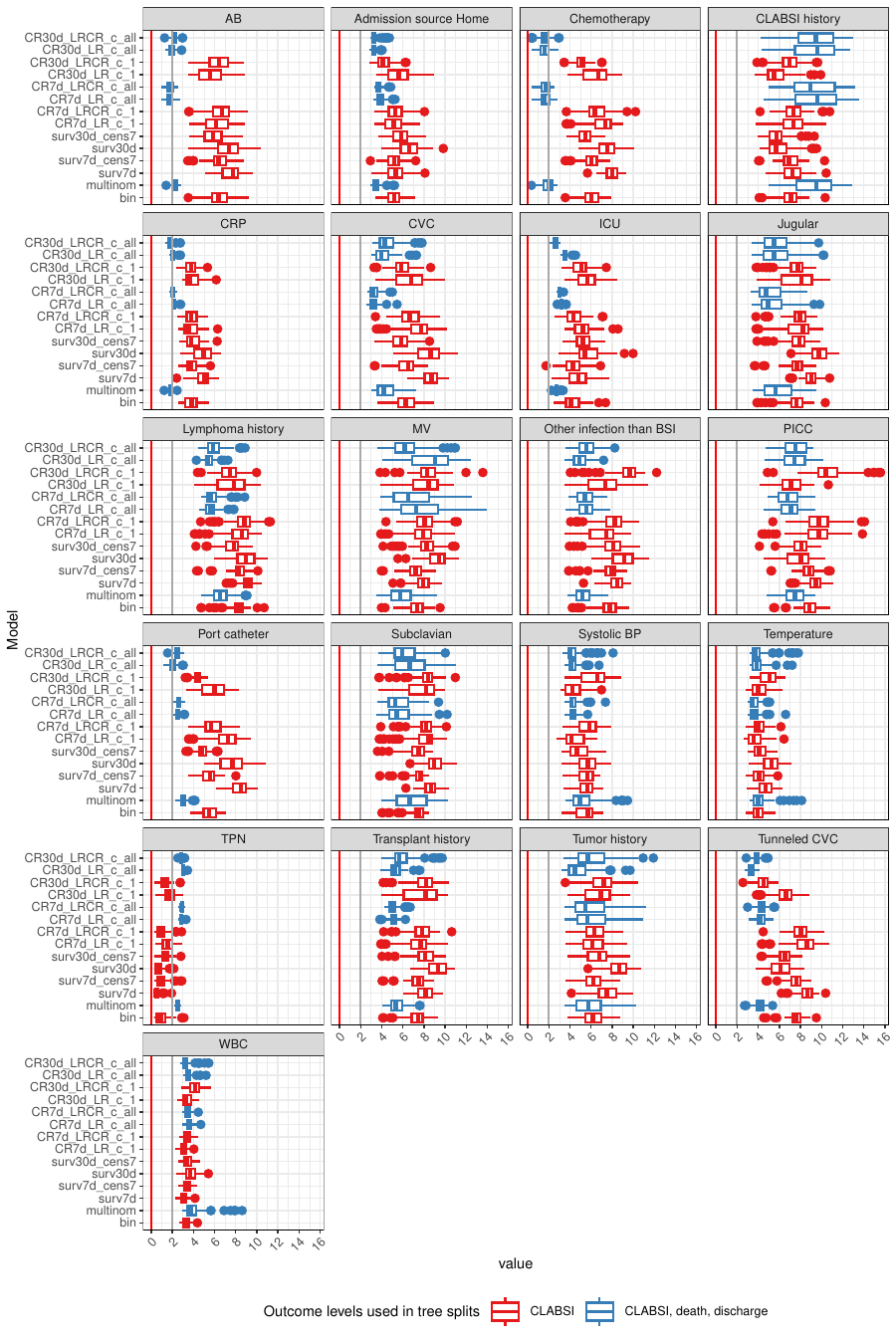}
\caption{\label{fig:var-imp-base}Variable importance}
\end{figure}

The minimal depth of the maximal subtree is used as a variable importance metric (Ishwaran et al. 2021), which is the depth in a tree on which the first split is made on a variable \(v\), averaged over all trees in the forest. The lowest possible value is 0 (root node split). The minimal depth of the maximal subtree is presented in Figure \ref{fig:var-imp-base}. A guiding line has been added to the plot on value 2, an aleatory choice to guide the focus on most important variables.

Models using multiple levels of the outcome put less weight on the TPN variable (splits tend to made lower down in the trees) and more weight (splits closer to the root) on variables like: ICU, antibacterials, antineoplastic agents (chemotherapy), CRP and port-catheter. Presumably, these variables could play a role in predicting discharge or death.

\subsection{Supplementary material 8 - Additional performance evaluation - dynamic models}\label{suppl-dynamic-perf}

\subsubsection{Number of CLABSI events and prevalence at each landmark}\label{number-of-clabsi-events-and-prevalence-at-each-landmark}

\begin{figure}
\centering
\includegraphics{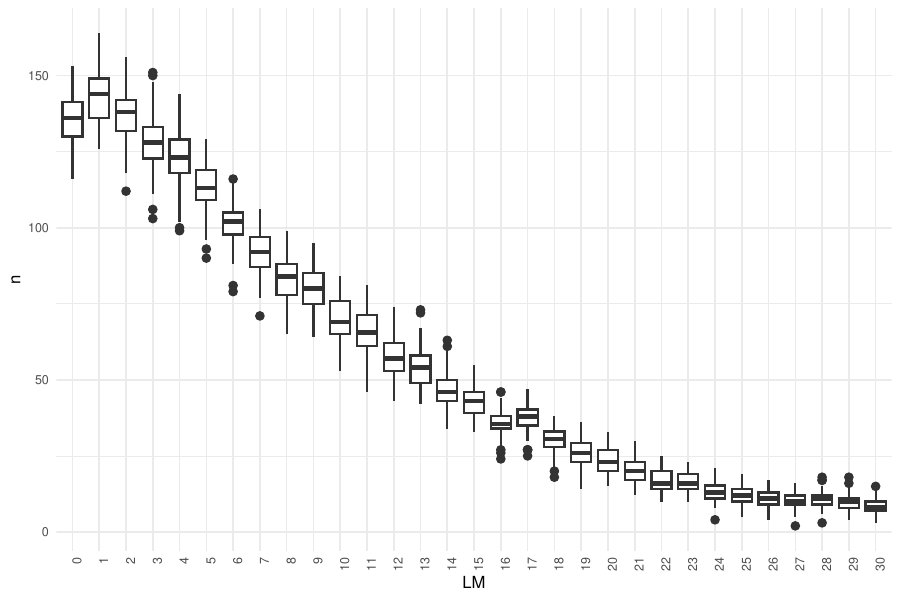}
\caption{\label{fig:dyn-n-LM}Number of CLABSI events at each landmark over all test sets}
\end{figure}

\begin{figure}
\centering
\includegraphics{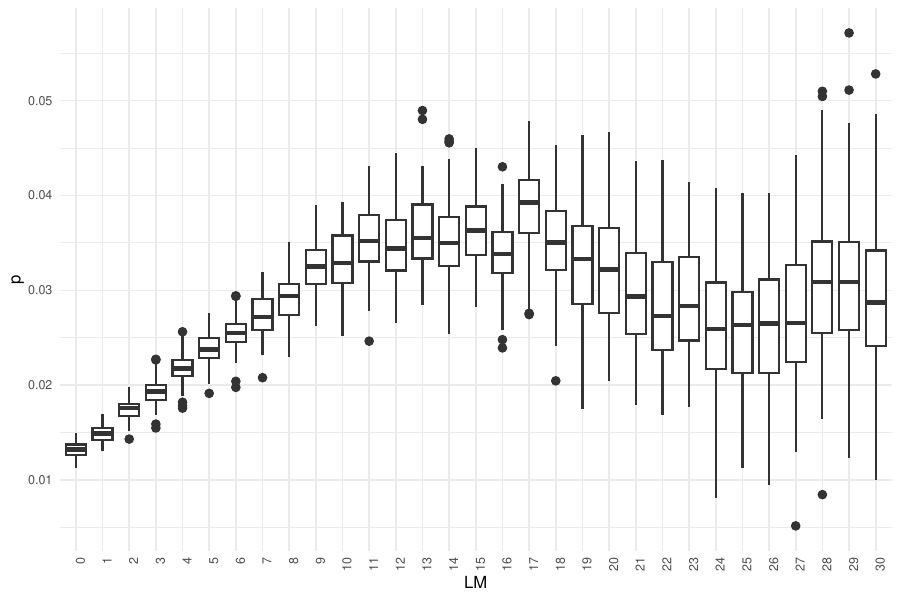}
\caption{\label{fig:dyn-prev-LM}CLABSI event prevalence at each landmark over all test sets}
\end{figure}

The number of CLABSI events and prevalence at each landmark over all test sets are presented in the Figures \ref{fig:dyn-n-LM} and \ref{fig:dyn-prev-LM}. The number of CLABSI events and prevalence at each landmark in train sets will be proportional, as the train/test splits have been performed by random sampling with a train:test ratio of 2:1.

\subsubsection{Time-dependent metrics table}\label{time-dependent-metrics-table}

The dynamic model performance metrics per LM are presented in Table \ref{tab:time-dependent-metrics-table}

\begin{longtable}[t]{lr>{\raggedright\arraybackslash}p{5em}>{\raggedright\arraybackslash}p{5em}>{\raggedright\arraybackslash}p{5em}>{\raggedright\arraybackslash}p{5em}>{\raggedright\arraybackslash}p{5em}>{\raggedright\arraybackslash}p{5em}}
\caption{\label{tab:time-dependent-metrics-table}Time-dependent metrics table}\\
\toprule
Model & LM & BSS & AUPRC & AUROC & Calibration slope & ECI & E:O ratio\\
\midrule
\endfirsthead
\caption[]{\label{tab:time-dependent-metrics-table}Time-dependent metrics table \textit{(continued)}}\\
\toprule
Model & LM & BSS & AUPRC & AUROC & Calibration slope & ECI & E:O ratio\\
\midrule
\endhead

\endfoot
\bottomrule
\endlastfoot
bin & 0 & 0.015 (0.012 - 0.018) & 0.044 (0.04 - 0.048) & 0.752 (0.742 - 0.762) & 0.938 (0.871 - 1.015) & 0.001 (0.001 - 0.002) & 1.084 (1.019 - 1.15)\\
bin & 1 & 0.021 (0.019 - 0.024) & 0.056 (0.052 - 0.062) & 0.763 (0.753 - 0.775) & 1.033 (0.941 - 1.111) & 0.002 (0.001 - 0.003) & 1.065 (1.012 - 1.146)\\
bin & 2 & 0.023 (0.02 - 0.026) & 0.062 (0.057 - 0.07) & 0.764 (0.753 - 0.777) & 1.063 (0.965 - 1.169) & 0.002 (0.001 - 0.004) & 1.012 (0.963 - 1.067)\\
bin & 3 & 0.024 (0.021 - 0.027) & 0.066 (0.061 - 0.074) & 0.759 (0.747 - 0.775) & 1.041 (0.953 - 1.137) & 0.003 (0.002 - 0.004) & 0.991 (0.94 - 1.045)\\
bin & 4 & 0.029 (0.026 - 0.032) & 0.078 (0.071 - 0.089) & 0.771 (0.759 - 0.787) & 1.082 (0.974 - 1.214) & 0.005 (0.002 - 0.009) & 0.961 (0.904 - 1.005)\\
\addlinespace
bin & 5 & 0.027 (0.024 - 0.029) & 0.073 (0.068 - 0.081) & 0.775 (0.762 - 0.787) & 1.096 (1.006 - 1.196) & 0.004 (0.003 - 0.006) & 0.955 (0.891 - 1.009)\\
bin & 6 & 0.024 (0.021 - 0.028) & 0.07 (0.065 - 0.079) & 0.757 (0.743 - 0.769) & 1.043 (0.947 - 1.139) & 0.005 (0.002 - 0.007) & 0.973 (0.908 - 1.021)\\
bin & 7 & 0.024 (0.02 - 0.027) & 0.074 (0.066 - 0.079) & 0.754 (0.743 - 0.771) & 1.063 (0.954 - 1.198) & 0.005 (0.003 - 0.008) & 0.996 (0.907 - 1.049)\\
bin & 8 & 0.025 (0.022 - 0.028) & 0.077 (0.07 - 0.085) & 0.756 (0.742 - 0.771) & 1.083 (0.981 - 1.213) & 0.006 (0.004 - 0.009) & 0.972 (0.909 - 1.061)\\
bin & 9 & 0.03 (0.026 - 0.033) & 0.091 (0.083 - 0.1) & 0.757 (0.744 - 0.773) & 1.135 (1.004 - 1.261) & 0.01 (0.006 - 0.018) & 0.918 (0.855 - 0.994)\\
\addlinespace
bin & 10 & 0.025 (0.02 - 0.029) & 0.082 (0.075 - 0.091) & 0.741 (0.726 - 0.759) & 1.041 (0.919 - 1.19) & 0.01 (0.005 - 0.016) & 0.941 (0.869 - 1.038)\\
bin & 11 & 0.026 (0.021 - 0.03) & 0.086 (0.077 - 0.097) & 0.743 (0.727 - 0.758) & 1.054 (0.93 - 1.2) & 0.012 (0.006 - 0.021) & 0.93 (0.834 - 0.989)\\
bin & 12 & 0.024 (0.019 - 0.028) & 0.083 (0.073 - 0.093) & 0.733 (0.709 - 0.75) & 1.002 (0.846 - 1.144) & 0.01 (0.005 - 0.018) & 0.954 (0.867 - 1.043)\\
bin & 13 & 0.022 (0.017 - 0.027) & 0.085 (0.072 - 0.098) & 0.726 (0.7 - 0.74) & 1 (0.793 - 1.127) & 0.012 (0.007 - 0.021) & 0.941 (0.85 - 1.029)\\
bin & 14 & 0.023 (0.018 - 0.027) & 0.085 (0.076 - 0.1) & 0.732 (0.708 - 0.745) & 1.038 (0.817 - 1.183) & 0.011 (0.006 - 0.019) & 0.988 (0.902 - 1.068)\\
\addlinespace
bin & 15 & 0.025 (0.022 - 0.028) & 0.093 (0.083 - 0.104) & 0.731 (0.718 - 0.755) & 1.082 (0.949 - 1.271) & 0.014 (0.01 - 0.022) & 0.967 (0.884 - 1.058)\\
bin & 16 & 0.025 (0.021 - 0.028) & 0.089 (0.078 - 0.099) & 0.739 (0.72 - 0.759) & 1.106 (0.938 - 1.328) & 0.014 (0.009 - 0.02) & 1.052 (0.953 - 1.119)\\
bin & 17 & 0.022 (0.018 - 0.026) & 0.103 (0.085 - 0.122) & 0.718 (0.698 - 0.739) & 1.007 (0.884 - 1.215) & 0.019 (0.01 - 0.03) & 0.905 (0.831 - 1.004)\\
bin & 18 & 0.018 (0.013 - 0.023) & 0.083 (0.069 - 0.096) & 0.712 (0.681 - 0.738) & 0.971 (0.785 - 1.218) & 0.015 (0.01 - 0.025) & 1.003 (0.908 - 1.12)\\
bin & 19 & 0.014 (0.009 - 0.021) & 0.07 (0.06 - 0.084) & 0.711 (0.683 - 0.729) & 1 (0.839 - 1.192) & 0.018 (0.013 - 0.028) & 1.042 (0.937 - 1.228)\\
\addlinespace
bin & 20 & 0.011 (0.005 - 0.016) & 0.062 (0.053 - 0.075) & 0.695 (0.666 - 0.723) & 0.973 (0.834 - 1.184) & 0.018 (0.011 - 0.028) & 1.077 (0.93 - 1.29)\\
bin & 21 & 0.01 (0.002 - 0.016) & 0.062 (0.048 - 0.076) & 0.695 (0.661 - 0.721) & 1.027 (0.801 - 1.257) & 0.021 (0.012 - 0.032) & 1.185 (0.97 - 1.373)\\
bin & 22 & 0.004 (-0.004 - 0.012) & 0.046 (0.039 - 0.059) & 0.684 (0.652 - 0.707) & 1.03 (0.801 - 1.243) & 0.028 (0.015 - 0.042) & 1.252 (1.016 - 1.517)\\
bin & 23 & 0.007 (-0.001 - 0.013) & 0.051 (0.039 - 0.079) & 0.677 (0.648 - 0.709) & 1.043 (0.835 - 1.216) & 0.032 (0.019 - 0.048) & 1.155 (0.977 - 1.393)\\
bin & 24 & 0.008 (-0.004 - 0.016) & 0.048 (0.036 - 0.067) & 0.694 (0.661 - 0.729) & 1.142 (0.933 - 1.474) & 0.03 (0.018 - 0.044) & 1.288 (1.057 - 1.583)\\
\addlinespace
bin & 25 & 0.008 (-0.005 - 0.016) & 0.049 (0.038 - 0.073) & 0.693 (0.665 - 0.731) & 1.095 (0.869 - 1.412) & 0.03 (0.021 - 0.042) & 1.271 (1.077 - 1.566)\\
bin & 26 & 0.006 (-0.008 - 0.013) & 0.047 (0.037 - 0.07) & 0.684 (0.644 - 0.707) & 1.055 (0.842 - 1.329) & 0.041 (0.027 - 0.061) & 1.248 (1.042 - 1.592)\\
bin & 27 & -0.002 (-0.014 - 0.006) & 0.038 (0.029 - 0.048) & 0.642 (0.611 - 0.673) & 0.819 (0.64 - 1.035) & 0.046 (0.03 - 0.062) & 1.287 (1.028 - 1.601)\\
bin & 28 & 0.003 (-0.004 - 0.01) & 0.044 (0.035 - 0.053) & 0.662 (0.622 - 0.693) & 0.914 (0.701 - 1.185) & 0.049 (0.035 - 0.07) & 1.126 (0.977 - 1.36)\\
bin & 29 & 0.008 (-0.003 - 0.013) & 0.046 (0.038 - 0.058) & 0.684 (0.65 - 0.713) & 1.134 (0.864 - 1.365) & 0.046 (0.033 - 0.072) & 1.173 (0.976 - 1.377)\\
\addlinespace
bin & 30 & 0.014 (0.002 - 0.021) & 0.055 (0.042 - 0.069) & 0.736 (0.7 - 0.772) & 1.568 (1.183 - 2.016) & 0.048 (0.029 - 0.064) & 1.218 (0.99 - 1.463)\\
multinom & 0 & 0.016 (0.012 - 0.018) & 0.046 (0.042 - 0.051) & 0.746 (0.736 - 0.76) & 0.936 (0.866 - 1.028) & 0.002 (0.001 - 0.003) & 1.09 (1.028 - 1.161)\\
multinom & 1 & 0.021 (0.018 - 0.024) & 0.056 (0.052 - 0.061) & 0.763 (0.752 - 0.772) & 1.022 (0.935 - 1.088) & 0.002 (0.001 - 0.003) & 1.077 (1.019 - 1.15)\\
multinom & 2 & 0.023 (0.02 - 0.026) & 0.062 (0.057 - 0.068) & 0.762 (0.751 - 0.773) & 1.056 (0.977 - 1.135) & 0.002 (0.001 - 0.003) & 1.047 (0.992 - 1.102)\\
multinom & 3 & 0.025 (0.021 - 0.028) & 0.068 (0.062 - 0.075) & 0.76 (0.747 - 0.777) & 1.031 (0.956 - 1.145) & 0.003 (0.002 - 0.005) & 1.019 (0.98 - 1.083)\\
\addlinespace
multinom & 4 & 0.028 (0.025 - 0.032) & 0.076 (0.07 - 0.084) & 0.771 (0.757 - 0.782) & 1.054 (0.986 - 1.156) & 0.003 (0.002 - 0.007) & 0.98 (0.929 - 1.025)\\
multinom & 5 & 0.026 (0.022 - 0.029) & 0.072 (0.066 - 0.08) & 0.772 (0.761 - 0.786) & 1.035 (0.976 - 1.158) & 0.004 (0.003 - 0.007) & 0.954 (0.896 - 1.01)\\
multinom & 6 & 0.022 (0.019 - 0.026) & 0.067 (0.061 - 0.075) & 0.751 (0.736 - 0.763) & 0.961 (0.903 - 1.064) & 0.005 (0.003 - 0.008) & 0.96 (0.9 - 1.014)\\
multinom & 7 & 0.022 (0.018 - 0.025) & 0.069 (0.062 - 0.075) & 0.753 (0.739 - 0.769) & 1.012 (0.919 - 1.118) & 0.007 (0.004 - 0.01) & 0.977 (0.9 - 1.042)\\
multinom & 8 & 0.022 (0.02 - 0.026) & 0.073 (0.066 - 0.081) & 0.749 (0.734 - 0.763) & 1.03 (0.928 - 1.15) & 0.007 (0.004 - 0.011) & 0.962 (0.906 - 1.039)\\
\addlinespace
multinom & 9 & 0.028 (0.024 - 0.031) & 0.087 (0.079 - 0.097) & 0.756 (0.74 - 0.768) & 1.123 (0.996 - 1.226) & 0.01 (0.005 - 0.015) & 0.913 (0.865 - 0.988)\\
multinom & 10 & 0.024 (0.021 - 0.029) & 0.082 (0.073 - 0.093) & 0.743 (0.727 - 0.758) & 1.078 (0.94 - 1.183) & 0.011 (0.006 - 0.017) & 0.949 (0.873 - 1.038)\\
multinom & 11 & 0.025 (0.02 - 0.03) & 0.086 (0.076 - 0.095) & 0.744 (0.728 - 0.763) & 1.08 (0.969 - 1.226) & 0.012 (0.008 - 0.02) & 0.929 (0.843 - 0.989)\\
multinom & 12 & 0.024 (0.019 - 0.028) & 0.084 (0.074 - 0.098) & 0.736 (0.712 - 0.75) & 1.04 (0.902 - 1.176) & 0.011 (0.007 - 0.02) & 0.957 (0.88 - 1.038)\\
multinom & 13 & 0.023 (0.017 - 0.028) & 0.09 (0.078 - 0.103) & 0.724 (0.7 - 0.74) & 1.062 (0.862 - 1.177) & 0.012 (0.007 - 0.021) & 0.937 (0.843 - 1.014)\\
\addlinespace
multinom & 14 & 0.025 (0.02 - 0.029) & 0.09 (0.079 - 0.108) & 0.73 (0.71 - 0.745) & 1.124 (0.939 - 1.248) & 0.013 (0.008 - 0.023) & 0.99 (0.897 - 1.072)\\
multinom & 15 & 0.027 (0.023 - 0.031) & 0.106 (0.089 - 0.125) & 0.734 (0.714 - 0.747) & 1.152 (0.989 - 1.303) & 0.022 (0.012 - 0.031) & 0.963 (0.878 - 1.054)\\
multinom & 16 & 0.027 (0.023 - 0.029) & 0.1 (0.086 - 0.115) & 0.742 (0.72 - 0.757) & 1.187 (1.031 - 1.357) & 0.018 (0.011 - 0.027) & 1.055 (0.948 - 1.11)\\
multinom & 17 & 0.022 (0.018 - 0.026) & 0.097 (0.084 - 0.113) & 0.716 (0.696 - 0.734) & 1.014 (0.858 - 1.193) & 0.018 (0.012 - 0.031) & 0.893 (0.83 - 0.984)\\
multinom & 18 & 0.019 (0.013 - 0.024) & 0.085 (0.075 - 0.099) & 0.707 (0.678 - 0.738) & 0.991 (0.813 - 1.243) & 0.019 (0.011 - 0.028) & 0.994 (0.913 - 1.098)\\
\addlinespace
multinom & 19 & 0.014 (0.009 - 0.02) & 0.072 (0.06 - 0.082) & 0.707 (0.677 - 0.729) & 0.992 (0.847 - 1.207) & 0.019 (0.011 - 0.028) & 1.034 (0.938 - 1.182)\\
multinom & 20 & 0.012 (0.006 - 0.017) & 0.064 (0.055 - 0.079) & 0.699 (0.666 - 0.727) & 0.946 (0.814 - 1.18) & 0.021 (0.013 - 0.03) & 1.064 (0.914 - 1.234)\\
multinom & 21 & 0.009 (0.001 - 0.015) & 0.06 (0.046 - 0.073) & 0.69 (0.656 - 0.72) & 0.943 (0.745 - 1.152) & 0.02 (0.015 - 0.03) & 1.163 (0.955 - 1.332)\\
multinom & 22 & 0.004 (-0.003 - 0.013) & 0.048 (0.039 - 0.065) & 0.686 (0.653 - 0.711) & 0.985 (0.76 - 1.151) & 0.029 (0.018 - 0.041) & 1.212 (1.011 - 1.447)\\
multinom & 23 & 0.004 (-0.003 - 0.01) & 0.046 (0.036 - 0.063) & 0.668 (0.643 - 0.703) & 0.918 (0.764 - 1.11) & 0.036 (0.025 - 0.053) & 1.143 (0.95 - 1.372)\\
\addlinespace
multinom & 24 & 0.007 (-0.006 - 0.015) & 0.047 (0.035 - 0.065) & 0.699 (0.669 - 0.734) & 1.108 (0.879 - 1.37) & 0.034 (0.021 - 0.051) & 1.249 (1.025 - 1.513)\\
multinom & 25 & 0.007 (-0.007 - 0.015) & 0.047 (0.034 - 0.073) & 0.692 (0.664 - 0.726) & 1.058 (0.779 - 1.312) & 0.033 (0.022 - 0.05) & 1.21 (1.049 - 1.538)\\
multinom & 26 & 0.008 (-0.006 - 0.015) & 0.051 (0.034 - 0.07) & 0.688 (0.643 - 0.718) & 0.994 (0.78 - 1.273) & 0.041 (0.025 - 0.062) & 1.193 (0.996 - 1.527)\\
multinom & 27 & 0 (-0.009 - 0.007) & 0.042 (0.031 - 0.051) & 0.654 (0.618 - 0.686) & 0.806 (0.634 - 1.078) & 0.045 (0.026 - 0.06) & 1.185 (0.962 - 1.496)\\
multinom & 28 & 0.004 (-0.004 - 0.009) & 0.045 (0.036 - 0.055) & 0.66 (0.631 - 0.69) & 0.862 (0.649 - 1.087) & 0.044 (0.03 - 0.066) & 1.039 (0.912 - 1.265)\\
\addlinespace
multinom & 29 & 0.009 (0.001 - 0.017) & 0.053 (0.039 - 0.068) & 0.701 (0.669 - 0.734) & 1.088 (0.84 - 1.438) & 0.042 (0.027 - 0.06) & 1.07 (0.905 - 1.271)\\
multinom & 30 & 0.011 (0.001 - 0.02) & 0.053 (0.041 - 0.073) & 0.722 (0.686 - 0.764) & 1.285 (0.957 - 1.658) & 0.041 (0.029 - 0.058) & 1.088 (0.922 - 1.307)\\
surv7d & 0 & 0.007 (0.001 - 0.011) & 0.043 (0.039 - 0.048) & 0.74 (0.732 - 0.753) & 1.069 (0.982 - 1.152) & 0.01 (0.008 - 0.015) & 1.576 (1.49 - 1.688)\\
surv7d & 1 & 0.014 (0.01 - 0.017) & 0.056 (0.051 - 0.062) & 0.751 (0.744 - 0.766) & 1.155 (1.068 - 1.255) & 0.01 (0.007 - 0.013) & 1.529 (1.457 - 1.653)\\
surv7d & 2 & 0.018 (0.013 - 0.022) & 0.061 (0.056 - 0.068) & 0.754 (0.741 - 0.764) & 1.174 (1.083 - 1.259) & 0.009 (0.007 - 0.013) & 1.429 (1.363 - 1.526)\\
\addlinespace
surv7d & 3 & 0.019 (0.013 - 0.024) & 0.064 (0.059 - 0.071) & 0.752 (0.74 - 0.765) & 1.159 (1.063 - 1.267) & 0.011 (0.008 - 0.015) & 1.406 (1.338 - 1.504)\\
surv7d & 4 & 0.025 (0.02 - 0.03) & 0.076 (0.069 - 0.087) & 0.763 (0.748 - 0.78) & 1.2 (1.102 - 1.331) & 0.012 (0.008 - 0.016) & 1.365 (1.291 - 1.444)\\
surv7d & 5 & 0.023 (0.018 - 0.028) & 0.075 (0.068 - 0.083) & 0.768 (0.755 - 0.78) & 1.189 (1.092 - 1.285) & 0.012 (0.008 - 0.017) & 1.347 (1.261 - 1.413)\\
surv7d & 6 & 0.019 (0.015 - 0.026) & 0.072 (0.066 - 0.082) & 0.751 (0.738 - 0.763) & 1.111 (1.005 - 1.205) & 0.015 (0.01 - 0.02) & 1.351 (1.271 - 1.43)\\
surv7d & 7 & 0.019 (0.011 - 0.025) & 0.074 (0.065 - 0.082) & 0.75 (0.736 - 0.764) & 1.099 (0.989 - 1.232) & 0.018 (0.012 - 0.026) & 1.363 (1.233 - 1.445)\\
\addlinespace
surv7d & 8 & 0.021 (0.015 - 0.028) & 0.08 (0.07 - 0.088) & 0.757 (0.741 - 0.77) & 1.13 (1.037 - 1.262) & 0.018 (0.012 - 0.025) & 1.317 (1.224 - 1.412)\\
surv7d & 9 & 0.03 (0.023 - 0.035) & 0.094 (0.085 - 0.102) & 0.757 (0.74 - 0.77) & 1.155 (1.06 - 1.3) & 0.014 (0.008 - 0.02) & 1.204 (1.131 - 1.307)\\
surv7d & 10 & 0.025 (0.018 - 0.03) & 0.086 (0.076 - 0.097) & 0.741 (0.726 - 0.759) & 1.085 (0.976 - 1.219) & 0.018 (0.011 - 0.024) & 1.227 (1.114 - 1.336)\\
surv7d & 11 & 0.025 (0.019 - 0.03) & 0.085 (0.078 - 0.092) & 0.741 (0.726 - 0.756) & 1.073 (0.964 - 1.239) & 0.017 (0.009 - 0.023) & 1.2 (1.079 - 1.276)\\
surv7d & 12 & 0.02 (0.014 - 0.026) & 0.078 (0.07 - 0.088) & 0.733 (0.713 - 0.747) & 1.028 (0.887 - 1.153) & 0.021 (0.013 - 0.029) & 1.224 (1.11 - 1.338)\\
\addlinespace
surv7d & 13 & 0.02 (0.013 - 0.027) & 0.08 (0.07 - 0.093) & 0.724 (0.703 - 0.741) & 1.012 (0.839 - 1.154) & 0.02 (0.014 - 0.031) & 1.193 (1.081 - 1.298)\\
surv7d & 14 & 0.018 (0.009 - 0.025) & 0.077 (0.068 - 0.09) & 0.723 (0.705 - 0.738) & 1.022 (0.846 - 1.163) & 0.022 (0.015 - 0.033) & 1.256 (1.145 - 1.362)\\
surv7d & 15 & 0.02 (0.014 - 0.025) & 0.08 (0.072 - 0.09) & 0.724 (0.704 - 0.741) & 1.042 (0.941 - 1.246) & 0.021 (0.015 - 0.035) & 1.222 (1.113 - 1.324)\\
surv7d & 16 & 0.019 (0.011 - 0.024) & 0.08 (0.067 - 0.09) & 0.73 (0.71 - 0.748) & 1.088 (0.911 - 1.297) & 0.023 (0.017 - 0.038) & 1.32 (1.191 - 1.4)\\
surv7d & 17 & 0.019 (0.012 - 0.025) & 0.086 (0.072 - 0.104) & 0.706 (0.688 - 0.727) & 1.018 (0.827 - 1.172) & 0.02 (0.011 - 0.035) & 1.126 (1.04 - 1.255)\\
\addlinespace
surv7d & 18 & 0.012 (0.004 - 0.017) & 0.071 (0.062 - 0.083) & 0.7 (0.679 - 0.723) & 0.945 (0.788 - 1.165) & 0.028 (0.017 - 0.043) & 1.247 (1.123 - 1.395)\\
surv7d & 19 & 0.007 (-0.002 - 0.016) & 0.064 (0.055 - 0.076) & 0.696 (0.671 - 0.718) & 1 (0.806 - 1.147) & 0.038 (0.02 - 0.053) & 1.317 (1.17 - 1.541)\\
surv7d & 20 & 0.003 (-0.009 - 0.014) & 0.058 (0.048 - 0.071) & 0.685 (0.654 - 0.713) & 0.95 (0.79 - 1.18) & 0.044 (0.024 - 0.059) & 1.357 (1.156 - 1.614)\\
surv7d & 21 & -0.001 (-0.013 - 0.011) & 0.054 (0.044 - 0.07) & 0.69 (0.655 - 0.711) & 0.994 (0.768 - 1.227) & 0.045 (0.029 - 0.067) & 1.504 (1.224 - 1.738)\\
surv7d & 22 & -0.009 (-0.026 - 0.003) & 0.043 (0.035 - 0.056) & 0.665 (0.638 - 0.696) & 0.929 (0.808 - 1.155) & 0.069 (0.04 - 0.089) & 1.587 (1.285 - 1.901)\\
\addlinespace
surv7d & 23 & -0.003 (-0.022 - 0.006) & 0.046 (0.035 - 0.06) & 0.671 (0.644 - 0.698) & 0.97 (0.845 - 1.204) & 0.054 (0.035 - 0.085) & 1.491 (1.228 - 1.809)\\
surv7d & 24 & -0.006 (-0.033 - 0.007) & 0.042 (0.033 - 0.056) & 0.688 (0.655 - 0.716) & 1.134 (0.829 - 1.325) & 0.059 (0.036 - 0.091) & 1.634 (1.339 - 2.033)\\
surv7d & 25 & -0.009 (-0.031 - 0.007) & 0.044 (0.034 - 0.061) & 0.685 (0.66 - 0.711) & 1.099 (0.847 - 1.361) & 0.068 (0.036 - 0.096) & 1.633 (1.383 - 2.03)\\
surv7d & 26 & -0.007 (-0.029 - 0.005) & 0.044 (0.033 - 0.057) & 0.671 (0.639 - 0.697) & 1.06 (0.821 - 1.277) & 0.069 (0.035 - 0.102) & 1.621 (1.333 - 2.028)\\
surv7d & 27 & -0.016 (-0.038 - -0.001) & 0.037 (0.029 - 0.044) & 0.647 (0.616 - 0.678) & 0.865 (0.693 - 1.047) & 0.089 (0.06 - 0.113) & 1.653 (1.302 - 2.044)\\
\addlinespace
surv7d & 28 & -0.005 (-0.024 - 0.003) & 0.043 (0.035 - 0.05) & 0.654 (0.623 - 0.686) & 0.945 (0.701 - 1.148) & 0.078 (0.056 - 0.105) & 1.446 (1.229 - 1.731)\\
surv7d & 29 & -0.002 (-0.021 - 0.007) & 0.046 (0.036 - 0.056) & 0.676 (0.635 - 0.7) & 1.061 (0.866 - 1.366) & 0.073 (0.049 - 0.113) & 1.465 (1.224 - 1.764)\\
surv7d & 30 & 0 (-0.019 - 0.015) & 0.05 (0.038 - 0.064) & 0.708 (0.661 - 0.733) & 1.366 (0.995 - 1.772) & 0.071 (0.047 - 0.096) & 1.541 (1.272 - 1.841)\\
surv7d\_cens7 & 0 & 0.015 (0.012 - 0.018) & 0.044 (0.04 - 0.049) & 0.751 (0.742 - 0.761) & 0.938 (0.86 - 1.019) & 0.002 (0.001 - 0.003) & 1.088 (1.021 - 1.153)\\
surv7d\_cens7 & 1 & 0.021 (0.019 - 0.023) & 0.057 (0.052 - 0.061) & 0.763 (0.752 - 0.776) & 1.02 (0.945 - 1.104) & 0.002 (0.001 - 0.003) & 1.069 (1.02 - 1.148)\\
\addlinespace
surv7d\_cens7 & 2 & 0.023 (0.02 - 0.026) & 0.062 (0.058 - 0.071) & 0.763 (0.753 - 0.778) & 1.069 (0.969 - 1.165) & 0.002 (0.001 - 0.004) & 1.015 (0.963 - 1.078)\\
surv7d\_cens7 & 3 & 0.024 (0.021 - 0.027) & 0.065 (0.061 - 0.073) & 0.76 (0.748 - 0.774) & 1.037 (0.955 - 1.135) & 0.003 (0.002 - 0.005) & 0.991 (0.945 - 1.053)\\
surv7d\_cens7 & 4 & 0.029 (0.026 - 0.032) & 0.078 (0.07 - 0.087) & 0.77 (0.759 - 0.785) & 1.079 (0.983 - 1.182) & 0.004 (0.002 - 0.009) & 0.958 (0.906 - 1.004)\\
surv7d\_cens7 & 5 & 0.027 (0.024 - 0.029) & 0.073 (0.068 - 0.082) & 0.775 (0.762 - 0.787) & 1.08 (1.007 - 1.192) & 0.005 (0.002 - 0.007) & 0.951 (0.887 - 1.005)\\
surv7d\_cens7 & 6 & 0.024 (0.021 - 0.028) & 0.07 (0.065 - 0.079) & 0.756 (0.744 - 0.768) & 1.034 (0.927 - 1.126) & 0.005 (0.003 - 0.007) & 0.968 (0.909 - 1.019)\\
\addlinespace
surv7d\_cens7 & 7 & 0.024 (0.02 - 0.027) & 0.075 (0.066 - 0.08) & 0.753 (0.742 - 0.771) & 1.051 (0.956 - 1.203) & 0.005 (0.003 - 0.008) & 0.989 (0.903 - 1.05)\\
surv7d\_cens7 & 8 & 0.025 (0.022 - 0.029) & 0.077 (0.07 - 0.084) & 0.755 (0.743 - 0.77) & 1.074 (0.982 - 1.212) & 0.006 (0.004 - 0.01) & 0.968 (0.91 - 1.057)\\
surv7d\_cens7 & 9 & 0.03 (0.026 - 0.033) & 0.092 (0.084 - 0.099) & 0.758 (0.744 - 0.774) & 1.104 (0.995 - 1.276) & 0.01 (0.006 - 0.017) & 0.918 (0.858 - 0.99)\\
surv7d\_cens7 & 10 & 0.025 (0.02 - 0.03) & 0.083 (0.074 - 0.093) & 0.739 (0.726 - 0.76) & 1.028 (0.923 - 1.174) & 0.011 (0.005 - 0.016) & 0.941 (0.868 - 1.038)\\
surv7d\_cens7 & 11 & 0.026 (0.021 - 0.03) & 0.086 (0.078 - 0.096) & 0.743 (0.728 - 0.759) & 1.057 (0.927 - 1.199) & 0.011 (0.006 - 0.022) & 0.928 (0.84 - 0.994)\\
\addlinespace
surv7d\_cens7 & 12 & 0.024 (0.019 - 0.028) & 0.083 (0.075 - 0.093) & 0.734 (0.71 - 0.75) & 1.02 (0.814 - 1.169) & 0.01 (0.006 - 0.02) & 0.958 (0.87 - 1.047)\\
surv7d\_cens7 & 13 & 0.023 (0.017 - 0.028) & 0.086 (0.074 - 0.096) & 0.727 (0.701 - 0.742) & 1.002 (0.798 - 1.117) & 0.012 (0.008 - 0.02) & 0.941 (0.844 - 1.016)\\
surv7d\_cens7 & 14 & 0.023 (0.018 - 0.028) & 0.085 (0.076 - 0.097) & 0.731 (0.71 - 0.744) & 1.029 (0.854 - 1.168) & 0.012 (0.006 - 0.017) & 0.987 (0.9 - 1.068)\\
surv7d\_cens7 & 15 & 0.025 (0.022 - 0.029) & 0.093 (0.083 - 0.105) & 0.733 (0.717 - 0.753) & 1.076 (0.933 - 1.277) & 0.014 (0.009 - 0.022) & 0.978 (0.887 - 1.062)\\
surv7d\_cens7 & 16 & 0.025 (0.021 - 0.029) & 0.089 (0.078 - 0.099) & 0.74 (0.717 - 0.762) & 1.113 (0.947 - 1.32) & 0.013 (0.009 - 0.021) & 1.058 (0.957 - 1.121)\\
\addlinespace
surv7d\_cens7 & 17 & 0.022 (0.018 - 0.027) & 0.104 (0.085 - 0.12) & 0.718 (0.697 - 0.741) & 1.021 (0.828 - 1.183) & 0.018 (0.012 - 0.035) & 0.903 (0.834 - 1.005)\\
surv7d\_cens7 & 18 & 0.018 (0.013 - 0.023) & 0.08 (0.069 - 0.096) & 0.712 (0.679 - 0.734) & 0.93 (0.761 - 1.175) & 0.017 (0.009 - 0.026) & 1.011 (0.907 - 1.115)\\
surv7d\_cens7 & 19 & 0.014 (0.009 - 0.02) & 0.068 (0.059 - 0.083) & 0.705 (0.676 - 0.73) & 0.959 (0.807 - 1.19) & 0.02 (0.014 - 0.026) & 1.042 (0.95 - 1.234)\\
surv7d\_cens7 & 20 & 0.011 (0.005 - 0.016) & 0.061 (0.052 - 0.075) & 0.693 (0.664 - 0.72) & 0.953 (0.817 - 1.175) & 0.018 (0.012 - 0.033) & 1.08 (0.925 - 1.289)\\
surv7d\_cens7 & 21 & 0.01 (0.001 - 0.017) & 0.063 (0.048 - 0.075) & 0.697 (0.66 - 0.721) & 1.02 (0.797 - 1.233) & 0.021 (0.013 - 0.036) & 1.185 (0.966 - 1.378)\\
\addlinespace
surv7d\_cens7 & 22 & 0.003 (-0.003 - 0.011) & 0.045 (0.038 - 0.062) & 0.682 (0.655 - 0.704) & 1.05 (0.794 - 1.235) & 0.027 (0.017 - 0.041) & 1.243 (1.018 - 1.513)\\
surv7d\_cens7 & 23 & 0.007 (-0.002 - 0.013) & 0.051 (0.039 - 0.074) & 0.678 (0.652 - 0.707) & 1.042 (0.85 - 1.225) & 0.033 (0.02 - 0.048) & 1.157 (0.958 - 1.396)\\
surv7d\_cens7 & 24 & 0.008 (-0.005 - 0.016) & 0.048 (0.037 - 0.063) & 0.7 (0.662 - 0.728) & 1.19 (0.916 - 1.442) & 0.031 (0.016 - 0.041) & 1.28 (1.06 - 1.582)\\
surv7d\_cens7 & 25 & 0.006 (-0.005 - 0.017) & 0.05 (0.036 - 0.08) & 0.694 (0.664 - 0.73) & 1.102 (0.861 - 1.435) & 0.029 (0.019 - 0.048) & 1.286 (1.085 - 1.579)\\
surv7d\_cens7 & 26 & 0.007 (-0.01 - 0.014) & 0.047 (0.036 - 0.071) & 0.68 (0.639 - 0.708) & 1.081 (0.782 - 1.275) & 0.043 (0.025 - 0.057) & 1.274 (1.034 - 1.611)\\
\addlinespace
surv7d\_cens7 & 27 & 0 (-0.014 - 0.006) & 0.04 (0.029 - 0.046) & 0.647 (0.612 - 0.676) & 0.812 (0.629 - 1.033) & 0.048 (0.03 - 0.063) & 1.285 (1.018 - 1.62)\\
surv7d\_cens7 & 28 & 0.004 (-0.005 - 0.01) & 0.045 (0.035 - 0.052) & 0.665 (0.626 - 0.695) & 0.948 (0.699 - 1.189) & 0.049 (0.036 - 0.067) & 1.126 (0.969 - 1.379)\\
surv7d\_cens7 & 29 & 0.008 (-0.001 - 0.013) & 0.047 (0.038 - 0.058) & 0.686 (0.644 - 0.724) & 1.121 (0.888 - 1.446) & 0.046 (0.035 - 0.073) & 1.146 (0.97 - 1.384)\\
surv7d\_cens7 & 30 & 0.015 (0.003 - 0.021) & 0.054 (0.041 - 0.069) & 0.736 (0.692 - 0.771) & 1.528 (1.183 - 1.968) & 0.044 (0.028 - 0.066) & 1.221 (1.01 - 1.445)\\
CR7d\_LR\_c\_1 & 0 & 0.016 (0.012 - 0.018) & 0.044 (0.04 - 0.048) & 0.749 (0.74 - 0.76) & 0.955 (0.872 - 1.04) & 0.001 (0.001 - 0.002) & 1.061 (0.998 - 1.136)\\
\addlinespace
CR7d\_LR\_c\_1 & 1 & 0.021 (0.018 - 0.023) & 0.056 (0.052 - 0.062) & 0.761 (0.75 - 0.774) & 1.037 (0.971 - 1.123) & 0.002 (0.001 - 0.003) & 1.061 (1.004 - 1.141)\\
CR7d\_LR\_c\_1 & 2 & 0.022 (0.02 - 0.025) & 0.062 (0.057 - 0.07) & 0.763 (0.751 - 0.775) & 1.077 (0.99 - 1.175) & 0.003 (0.001 - 0.004) & 1.013 (0.962 - 1.071)\\
CR7d\_LR\_c\_1 & 3 & 0.024 (0.021 - 0.027) & 0.066 (0.061 - 0.072) & 0.759 (0.748 - 0.773) & 1.081 (0.975 - 1.158) & 0.003 (0.002 - 0.005) & 1.005 (0.953 - 1.064)\\
CR7d\_LR\_c\_1 & 4 & 0.028 (0.025 - 0.032) & 0.078 (0.07 - 0.09) & 0.771 (0.758 - 0.785) & 1.12 (1.021 - 1.225) & 0.004 (0.002 - 0.009) & 0.98 (0.926 - 1.03)\\
CR7d\_LR\_c\_1 & 5 & 0.027 (0.024 - 0.029) & 0.073 (0.069 - 0.082) & 0.774 (0.761 - 0.786) & 1.114 (1.025 - 1.213) & 0.005 (0.003 - 0.007) & 0.973 (0.911 - 1.026)\\
\addlinespace
CR7d\_LR\_c\_1 & 6 & 0.024 (0.021 - 0.028) & 0.071 (0.064 - 0.079) & 0.756 (0.744 - 0.769) & 1.037 (0.947 - 1.149) & 0.005 (0.002 - 0.008) & 0.989 (0.923 - 1.037)\\
CR7d\_LR\_c\_1 & 7 & 0.024 (0.02 - 0.027) & 0.074 (0.065 - 0.08) & 0.753 (0.741 - 0.771) & 1.042 (0.935 - 1.181) & 0.005 (0.003 - 0.009) & 1.003 (0.917 - 1.06)\\
CR7d\_LR\_c\_1 & 8 & 0.025 (0.022 - 0.029) & 0.078 (0.069 - 0.085) & 0.756 (0.741 - 0.773) & 1.074 (0.982 - 1.203) & 0.006 (0.004 - 0.01) & 0.974 (0.912 - 1.063)\\
CR7d\_LR\_c\_1 & 9 & 0.031 (0.026 - 0.034) & 0.092 (0.083 - 0.102) & 0.759 (0.744 - 0.775) & 1.1 (0.99 - 1.261) & 0.011 (0.006 - 0.018) & 0.914 (0.853 - 0.987)\\
CR7d\_LR\_c\_1 & 10 & 0.026 (0.021 - 0.03) & 0.083 (0.076 - 0.094) & 0.742 (0.728 - 0.759) & 1.036 (0.905 - 1.145) & 0.01 (0.006 - 0.016) & 0.941 (0.858 - 1.025)\\
\addlinespace
CR7d\_LR\_c\_1 & 11 & 0.027 (0.021 - 0.03) & 0.084 (0.077 - 0.094) & 0.745 (0.73 - 0.759) & 1.048 (0.922 - 1.152) & 0.011 (0.007 - 0.019) & 0.927 (0.832 - 0.991)\\
CR7d\_LR\_c\_1 & 12 & 0.024 (0.018 - 0.028) & 0.083 (0.073 - 0.094) & 0.735 (0.712 - 0.75) & 0.995 (0.832 - 1.13) & 0.01 (0.005 - 0.02) & 0.949 (0.867 - 1.029)\\
CR7d\_LR\_c\_1 & 13 & 0.023 (0.017 - 0.028) & 0.084 (0.073 - 0.096) & 0.727 (0.701 - 0.741) & 0.964 (0.782 - 1.092) & 0.012 (0.007 - 0.019) & 0.942 (0.841 - 1.025)\\
CR7d\_LR\_c\_1 & 14 & 0.023 (0.018 - 0.028) & 0.084 (0.073 - 0.096) & 0.731 (0.708 - 0.746) & 1.009 (0.839 - 1.158) & 0.01 (0.006 - 0.02) & 0.987 (0.897 - 1.062)\\
CR7d\_LR\_c\_1 & 15 & 0.025 (0.021 - 0.029) & 0.09 (0.082 - 0.101) & 0.732 (0.712 - 0.751) & 1.026 (0.909 - 1.233) & 0.014 (0.007 - 0.02) & 0.962 (0.88 - 1.047)\\
\addlinespace
CR7d\_LR\_c\_1 & 16 & 0.025 (0.02 - 0.028) & 0.087 (0.077 - 0.097) & 0.74 (0.717 - 0.758) & 1.076 (0.905 - 1.27) & 0.013 (0.007 - 0.019) & 1.046 (0.95 - 1.112)\\
CR7d\_LR\_c\_1 & 17 & 0.022 (0.018 - 0.026) & 0.102 (0.085 - 0.115) & 0.72 (0.698 - 0.738) & 0.983 (0.811 - 1.174) & 0.019 (0.01 - 0.032) & 0.9 (0.825 - 0.997)\\
CR7d\_LR\_c\_1 & 18 & 0.018 (0.013 - 0.023) & 0.079 (0.069 - 0.094) & 0.712 (0.685 - 0.733) & 0.946 (0.734 - 1.173) & 0.016 (0.009 - 0.026) & 1.001 (0.896 - 1.112)\\
CR7d\_LR\_c\_1 & 19 & 0.014 (0.009 - 0.02) & 0.069 (0.059 - 0.082) & 0.706 (0.68 - 0.731) & 0.975 (0.818 - 1.145) & 0.018 (0.012 - 0.029) & 1.045 (0.927 - 1.226)\\
CR7d\_LR\_c\_1 & 20 & 0.01 (0.004 - 0.015) & 0.061 (0.052 - 0.075) & 0.692 (0.661 - 0.72) & 0.91 (0.792 - 1.108) & 0.019 (0.012 - 0.03) & 1.087 (0.914 - 1.281)\\
\addlinespace
CR7d\_LR\_c\_1 & 21 & 0.01 (0.001 - 0.016) & 0.06 (0.047 - 0.073) & 0.695 (0.659 - 0.72) & 0.995 (0.763 - 1.188) & 0.021 (0.013 - 0.036) & 1.183 (0.958 - 1.368)\\
CR7d\_LR\_c\_1 & 22 & 0.003 (-0.004 - 0.011) & 0.045 (0.038 - 0.061) & 0.683 (0.653 - 0.7) & 0.988 (0.808 - 1.18) & 0.03 (0.019 - 0.044) & 1.242 (1.008 - 1.509)\\
CR7d\_LR\_c\_1 & 23 & 0.006 (-0.001 - 0.012) & 0.05 (0.037 - 0.076) & 0.675 (0.648 - 0.698) & 0.97 (0.836 - 1.164) & 0.036 (0.021 - 0.049) & 1.165 (0.963 - 1.398)\\
CR7d\_LR\_c\_1 & 24 & 0.006 (-0.005 - 0.016) & 0.047 (0.036 - 0.064) & 0.69 (0.659 - 0.727) & 1.125 (0.855 - 1.414) & 0.034 (0.019 - 0.043) & 1.28 (1.063 - 1.572)\\
CR7d\_LR\_c\_1 & 25 & 0.006 (-0.006 - 0.016) & 0.05 (0.035 - 0.075) & 0.696 (0.667 - 0.729) & 1.075 (0.867 - 1.399) & 0.033 (0.02 - 0.045) & 1.267 (1.079 - 1.569)\\
\addlinespace
CR7d\_LR\_c\_1 & 26 & 0.006 (-0.008 - 0.014) & 0.047 (0.035 - 0.072) & 0.677 (0.642 - 0.71) & 1.024 (0.818 - 1.306) & 0.045 (0.026 - 0.062) & 1.235 (1.023 - 1.585)\\
CR7d\_LR\_c\_1 & 27 & -0.002 (-0.014 - 0.006) & 0.039 (0.03 - 0.045) & 0.644 (0.614 - 0.671) & 0.806 (0.62 - 1.014) & 0.048 (0.034 - 0.064) & 1.296 (1.027 - 1.593)\\
CR7d\_LR\_c\_1 & 28 & 0.002 (-0.005 - 0.01) & 0.044 (0.035 - 0.053) & 0.665 (0.625 - 0.697) & 0.945 (0.696 - 1.177) & 0.049 (0.036 - 0.069) & 1.121 (0.98 - 1.371)\\
CR7d\_LR\_c\_1 & 29 & 0.007 (-0.003 - 0.012) & 0.045 (0.036 - 0.057) & 0.678 (0.644 - 0.715) & 1.086 (0.824 - 1.341) & 0.047 (0.034 - 0.07) & 1.149 (0.983 - 1.392)\\
CR7d\_LR\_c\_1 & 30 & 0.012 (0 - 0.02) & 0.054 (0.04 - 0.065) & 0.728 (0.686 - 0.767) & 1.513 (1.078 - 1.957) & 0.047 (0.033 - 0.068) & 1.209 (0.996 - 1.466)\\
\addlinespace
CR7d\_LRCR\_c\_1 & 0 & 0.015 (0.012 - 0.018) & 0.045 (0.04 - 0.049) & 0.751 (0.742 - 0.762) & 0.949 (0.869 - 1.008) & 0.001 (0.001 - 0.003) & 1.084 (1.019 - 1.158)\\
CR7d\_LRCR\_c\_1 & 1 & 0.021 (0.018 - 0.023) & 0.056 (0.052 - 0.061) & 0.764 (0.752 - 0.775) & 1.019 (0.944 - 1.121) & 0.002 (0.001 - 0.003) & 1.07 (1.013 - 1.154)\\
CR7d\_LRCR\_c\_1 & 2 & 0.023 (0.02 - 0.026) & 0.062 (0.057 - 0.07) & 0.764 (0.753 - 0.778) & 1.067 (0.978 - 1.162) & 0.002 (0.001 - 0.003) & 1.012 (0.966 - 1.073)\\
CR7d\_LRCR\_c\_1 & 3 & 0.025 (0.021 - 0.027) & 0.067 (0.061 - 0.074) & 0.761 (0.748 - 0.774) & 1.048 (0.946 - 1.138) & 0.003 (0.002 - 0.005) & 0.995 (0.941 - 1.051)\\
CR7d\_LRCR\_c\_1 & 4 & 0.029 (0.025 - 0.032) & 0.078 (0.07 - 0.088) & 0.77 (0.76 - 0.788) & 1.081 (0.998 - 1.204) & 0.005 (0.002 - 0.009) & 0.958 (0.906 - 1.007)\\
\addlinespace
CR7d\_LRCR\_c\_1 & 5 & 0.026 (0.024 - 0.029) & 0.073 (0.069 - 0.081) & 0.775 (0.763 - 0.787) & 1.097 (1.006 - 1.162) & 0.004 (0.003 - 0.007) & 0.948 (0.888 - 1.003)\\
CR7d\_LRCR\_c\_1 & 6 & 0.024 (0.021 - 0.028) & 0.071 (0.064 - 0.078) & 0.757 (0.744 - 0.768) & 1.039 (0.952 - 1.126) & 0.005 (0.002 - 0.008) & 0.963 (0.906 - 1.02)\\
CR7d\_LRCR\_c\_1 & 7 & 0.024 (0.02 - 0.027) & 0.075 (0.066 - 0.081) & 0.754 (0.743 - 0.771) & 1.048 (0.959 - 1.195) & 0.005 (0.003 - 0.009) & 0.988 (0.906 - 1.05)\\
CR7d\_LRCR\_c\_1 & 8 & 0.025 (0.022 - 0.028) & 0.077 (0.07 - 0.084) & 0.756 (0.741 - 0.771) & 1.077 (0.988 - 1.205) & 0.006 (0.003 - 0.01) & 0.967 (0.91 - 1.058)\\
CR7d\_LRCR\_c\_1 & 9 & 0.03 (0.025 - 0.033) & 0.092 (0.082 - 0.098) & 0.757 (0.743 - 0.772) & 1.117 (1.003 - 1.274) & 0.011 (0.006 - 0.016) & 0.92 (0.854 - 0.989)\\
\addlinespace
CR7d\_LRCR\_c\_1 & 10 & 0.025 (0.021 - 0.029) & 0.082 (0.074 - 0.093) & 0.739 (0.726 - 0.757) & 1.047 (0.925 - 1.186) & 0.011 (0.005 - 0.015) & 0.947 (0.867 - 1.036)\\
CR7d\_LRCR\_c\_1 & 11 & 0.026 (0.021 - 0.03) & 0.086 (0.078 - 0.096) & 0.743 (0.726 - 0.76) & 1.058 (0.932 - 1.179) & 0.011 (0.006 - 0.022) & 0.928 (0.835 - 0.997)\\
CR7d\_LRCR\_c\_1 & 12 & 0.025 (0.019 - 0.028) & 0.084 (0.074 - 0.093) & 0.736 (0.711 - 0.751) & 1.016 (0.872 - 1.149) & 0.01 (0.005 - 0.018) & 0.959 (0.865 - 1.046)\\
CR7d\_LRCR\_c\_1 & 13 & 0.023 (0.017 - 0.028) & 0.086 (0.074 - 0.099) & 0.728 (0.7 - 0.741) & 1.012 (0.81 - 1.121) & 0.012 (0.007 - 0.018) & 0.941 (0.848 - 1.026)\\
CR7d\_LRCR\_c\_1 & 14 & 0.023 (0.018 - 0.028) & 0.085 (0.076 - 0.096) & 0.731 (0.71 - 0.747) & 1.038 (0.848 - 1.182) & 0.011 (0.006 - 0.019) & 0.994 (0.899 - 1.067)\\
\addlinespace
CR7d\_LRCR\_c\_1 & 15 & 0.026 (0.022 - 0.029) & 0.094 (0.083 - 0.105) & 0.732 (0.714 - 0.755) & 1.064 (0.941 - 1.278) & 0.014 (0.009 - 0.021) & 0.964 (0.882 - 1.055)\\
CR7d\_LRCR\_c\_1 & 16 & 0.025 (0.021 - 0.028) & 0.089 (0.079 - 0.101) & 0.739 (0.718 - 0.759) & 1.104 (0.939 - 1.295) & 0.014 (0.008 - 0.02) & 1.054 (0.954 - 1.113)\\
CR7d\_LRCR\_c\_1 & 17 & 0.022 (0.018 - 0.026) & 0.107 (0.084 - 0.123) & 0.717 (0.697 - 0.739) & 0.995 (0.836 - 1.171) & 0.02 (0.01 - 0.034) & 0.907 (0.83 - 1.006)\\
CR7d\_LRCR\_c\_1 & 18 & 0.018 (0.014 - 0.023) & 0.081 (0.07 - 0.095) & 0.712 (0.68 - 0.737) & 0.93 (0.763 - 1.222) & 0.016 (0.009 - 0.026) & 1.005 (0.912 - 1.12)\\
CR7d\_LRCR\_c\_1 & 19 & 0.014 (0.009 - 0.02) & 0.069 (0.059 - 0.082) & 0.707 (0.678 - 0.729) & 0.976 (0.823 - 1.204) & 0.018 (0.012 - 0.028) & 1.041 (0.943 - 1.225)\\
\addlinespace
CR7d\_LRCR\_c\_1 & 20 & 0.011 (0.005 - 0.016) & 0.062 (0.053 - 0.076) & 0.694 (0.663 - 0.722) & 0.967 (0.809 - 1.178) & 0.02 (0.012 - 0.03) & 1.077 (0.926 - 1.276)\\
CR7d\_LRCR\_c\_1 & 21 & 0.011 (0.001 - 0.016) & 0.061 (0.046 - 0.074) & 0.698 (0.662 - 0.718) & 1.028 (0.799 - 1.239) & 0.02 (0.012 - 0.031) & 1.18 (0.975 - 1.373)\\
CR7d\_LRCR\_c\_1 & 22 & 0.003 (-0.004 - 0.012) & 0.046 (0.038 - 0.061) & 0.685 (0.655 - 0.707) & 1.061 (0.785 - 1.23) & 0.028 (0.016 - 0.04) & 1.246 (1.014 - 1.512)\\
CR7d\_LRCR\_c\_1 & 23 & 0.008 (-0.001 - 0.014) & 0.052 (0.038 - 0.076) & 0.686 (0.65 - 0.71) & 1.04 (0.879 - 1.233) & 0.029 (0.019 - 0.049) & 1.157 (0.959 - 1.406)\\
CR7d\_LRCR\_c\_1 & 24 & 0.008 (-0.005 - 0.017) & 0.048 (0.036 - 0.065) & 0.695 (0.666 - 0.733) & 1.183 (0.898 - 1.485) & 0.031 (0.017 - 0.043) & 1.269 (1.077 - 1.568)\\
\addlinespace
CR7d\_LRCR\_c\_1 & 25 & 0.007 (-0.005 - 0.017) & 0.05 (0.037 - 0.079) & 0.698 (0.665 - 0.735) & 1.133 (0.861 - 1.466) & 0.03 (0.021 - 0.045) & 1.252 (1.071 - 1.575)\\
CR7d\_LRCR\_c\_1 & 26 & 0.006 (-0.008 - 0.015) & 0.046 (0.034 - 0.073) & 0.679 (0.641 - 0.71) & 1.087 (0.818 - 1.31) & 0.039 (0.027 - 0.059) & 1.269 (1.027 - 1.598)\\
CR7d\_LRCR\_c\_1 & 27 & 0.001 (-0.013 - 0.007) & 0.039 (0.028 - 0.047) & 0.646 (0.612 - 0.673) & 0.805 (0.614 - 1.046) & 0.046 (0.028 - 0.064) & 1.294 (1.012 - 1.585)\\
CR7d\_LRCR\_c\_1 & 28 & 0.004 (-0.005 - 0.01) & 0.045 (0.035 - 0.052) & 0.667 (0.629 - 0.697) & 0.938 (0.703 - 1.169) & 0.049 (0.035 - 0.068) & 1.134 (0.989 - 1.34)\\
CR7d\_LRCR\_c\_1 & 29 & 0.008 (-0.003 - 0.013) & 0.045 (0.037 - 0.057) & 0.689 (0.646 - 0.719) & 1.121 (0.85 - 1.385) & 0.048 (0.033 - 0.072) & 1.155 (0.964 - 1.359)\\
\addlinespace
CR7d\_LRCR\_c\_1 & 30 & 0.014 (0.003 - 0.021) & 0.054 (0.041 - 0.068) & 0.736 (0.691 - 0.772) & 1.507 (1.122 - 1.999) & 0.045 (0.028 - 0.065) & 1.212 (0.991 - 1.419)\\
CR7d\_LR\_c\_all & 0 & 0.015 (0.012 - 0.018) & 0.045 (0.04 - 0.05) & 0.744 (0.734 - 0.757) & 0.981 (0.901 - 1.05) & 0.001 (0.001 - 0.002) & 1.086 (1.018 - 1.149)\\
CR7d\_LR\_c\_all & 1 & 0.021 (0.018 - 0.024) & 0.056 (0.052 - 0.061) & 0.763 (0.75 - 0.772) & 1.048 (0.942 - 1.114) & 0.002 (0.001 - 0.003) & 1.085 (1.013 - 1.149)\\
CR7d\_LR\_c\_all & 2 & 0.023 (0.02 - 0.025) & 0.06 (0.056 - 0.067) & 0.76 (0.747 - 0.771) & 1.089 (0.995 - 1.176) & 0.002 (0.001 - 0.004) & 1.051 (0.992 - 1.106)\\
CR7d\_LR\_c\_all & 3 & 0.025 (0.021 - 0.028) & 0.067 (0.062 - 0.073) & 0.758 (0.742 - 0.773) & 1.065 (0.974 - 1.194) & 0.003 (0.002 - 0.005) & 1.034 (0.993 - 1.09)\\
\addlinespace
CR7d\_LR\_c\_all & 4 & 0.028 (0.024 - 0.032) & 0.076 (0.07 - 0.084) & 0.767 (0.751 - 0.78) & 1.102 (1.008 - 1.206) & 0.003 (0.002 - 0.006) & 0.997 (0.95 - 1.046)\\
CR7d\_LR\_c\_all & 5 & 0.026 (0.022 - 0.029) & 0.072 (0.067 - 0.079) & 0.77 (0.759 - 0.783) & 1.07 (1.008 - 1.186) & 0.005 (0.002 - 0.006) & 0.97 (0.909 - 1.029)\\
CR7d\_LR\_c\_all & 6 & 0.022 (0.018 - 0.026) & 0.066 (0.061 - 0.074) & 0.749 (0.735 - 0.763) & 1.01 (0.918 - 1.085) & 0.004 (0.003 - 0.009) & 0.975 (0.91 - 1.026)\\
CR7d\_LR\_c\_all & 7 & 0.021 (0.018 - 0.025) & 0.067 (0.061 - 0.074) & 0.75 (0.735 - 0.766) & 1.016 (0.923 - 1.128) & 0.006 (0.004 - 0.011) & 0.989 (0.902 - 1.052)\\
CR7d\_LR\_c\_all & 8 & 0.022 (0.019 - 0.026) & 0.071 (0.065 - 0.077) & 0.746 (0.732 - 0.76) & 1.045 (0.936 - 1.157) & 0.008 (0.004 - 0.011) & 0.966 (0.908 - 1.04)\\
\addlinespace
CR7d\_LR\_c\_all & 9 & 0.027 (0.024 - 0.031) & 0.087 (0.078 - 0.094) & 0.754 (0.738 - 0.767) & 1.137 (0.994 - 1.241) & 0.01 (0.005 - 0.015) & 0.919 (0.86 - 0.991)\\
CR7d\_LR\_c\_all & 10 & 0.024 (0.02 - 0.028) & 0.083 (0.073 - 0.091) & 0.741 (0.728 - 0.757) & 1.078 (0.971 - 1.218) & 0.01 (0.005 - 0.017) & 0.94 (0.867 - 1.029)\\
CR7d\_LR\_c\_all & 11 & 0.025 (0.02 - 0.029) & 0.085 (0.076 - 0.091) & 0.745 (0.727 - 0.763) & 1.108 (0.986 - 1.267) & 0.011 (0.008 - 0.02) & 0.925 (0.836 - 0.986)\\
CR7d\_LR\_c\_all & 12 & 0.024 (0.019 - 0.029) & 0.087 (0.077 - 0.098) & 0.735 (0.711 - 0.749) & 1.077 (0.907 - 1.213) & 0.013 (0.007 - 0.021) & 0.952 (0.869 - 1.028)\\
CR7d\_LR\_c\_all & 13 & 0.023 (0.017 - 0.027) & 0.088 (0.076 - 0.099) & 0.722 (0.699 - 0.74) & 1.074 (0.887 - 1.216) & 0.015 (0.008 - 0.022) & 0.935 (0.848 - 1.014)\\
\addlinespace
CR7d\_LR\_c\_all & 14 & 0.025 (0.021 - 0.029) & 0.09 (0.081 - 0.108) & 0.73 (0.712 - 0.746) & 1.136 (0.987 - 1.332) & 0.014 (0.008 - 0.029) & 0.98 (0.895 - 1.056)\\
CR7d\_LR\_c\_all & 15 & 0.026 (0.022 - 0.029) & 0.102 (0.088 - 0.12) & 0.73 (0.711 - 0.746) & 1.177 (0.992 - 1.317) & 0.021 (0.013 - 0.034) & 0.952 (0.868 - 1.041)\\
CR7d\_LR\_c\_all & 16 & 0.026 (0.022 - 0.029) & 0.097 (0.083 - 0.11) & 0.737 (0.717 - 0.758) & 1.219 (1.014 - 1.399) & 0.017 (0.011 - 0.032) & 1.037 (0.935 - 1.091)\\
CR7d\_LR\_c\_all & 17 & 0.022 (0.016 - 0.025) & 0.092 (0.079 - 0.112) & 0.715 (0.691 - 0.733) & 1.058 (0.873 - 1.254) & 0.021 (0.012 - 0.033) & 0.884 (0.817 - 0.97)\\
CR7d\_LR\_c\_all & 18 & 0.019 (0.013 - 0.023) & 0.082 (0.071 - 0.096) & 0.705 (0.677 - 0.735) & 1.021 (0.817 - 1.265) & 0.017 (0.011 - 0.03) & 0.977 (0.9 - 1.089)\\
\addlinespace
CR7d\_LR\_c\_all & 19 & 0.013 (0.009 - 0.019) & 0.066 (0.058 - 0.077) & 0.705 (0.674 - 0.725) & 1.021 (0.847 - 1.179) & 0.018 (0.012 - 0.024) & 1.04 (0.917 - 1.19)\\
CR7d\_LR\_c\_all & 20 & 0.012 (0.006 - 0.017) & 0.062 (0.054 - 0.076) & 0.7 (0.671 - 0.727) & 0.996 (0.825 - 1.244) & 0.019 (0.013 - 0.029) & 1.065 (0.895 - 1.23)\\
CR7d\_LR\_c\_all & 21 & 0.008 (0.002 - 0.014) & 0.056 (0.046 - 0.074) & 0.684 (0.651 - 0.715) & 0.974 (0.751 - 1.154) & 0.022 (0.014 - 0.03) & 1.153 (0.949 - 1.311)\\
CR7d\_LR\_c\_all & 22 & 0.004 (-0.003 - 0.012) & 0.046 (0.039 - 0.063) & 0.682 (0.652 - 0.713) & 0.98 (0.778 - 1.146) & 0.027 (0.016 - 0.041) & 1.186 (0.984 - 1.423)\\
CR7d\_LR\_c\_all & 23 & 0.006 (-0.002 - 0.011) & 0.046 (0.036 - 0.063) & 0.675 (0.646 - 0.701) & 0.943 (0.752 - 1.153) & 0.032 (0.022 - 0.049) & 1.118 (0.931 - 1.34)\\
\addlinespace
CR7d\_LR\_c\_all & 24 & 0.007 (-0.005 - 0.015) & 0.047 (0.033 - 0.066) & 0.7 (0.668 - 0.73) & 1.13 (0.866 - 1.403) & 0.034 (0.022 - 0.046) & 1.212 (1.008 - 1.479)\\
CR7d\_LR\_c\_all & 25 & 0.006 (-0.007 - 0.013) & 0.047 (0.035 - 0.073) & 0.686 (0.655 - 0.715) & 1.005 (0.779 - 1.245) & 0.034 (0.021 - 0.05) & 1.187 (1.022 - 1.507)\\
CR7d\_LR\_c\_all & 26 & 0.007 (-0.003 - 0.013) & 0.048 (0.037 - 0.076) & 0.677 (0.639 - 0.707) & 0.966 (0.76 - 1.183) & 0.039 (0.026 - 0.058) & 1.165 (0.968 - 1.491)\\
CR7d\_LR\_c\_all & 27 & 0 (-0.007 - 0.007) & 0.039 (0.031 - 0.05) & 0.645 (0.611 - 0.683) & 0.784 (0.631 - 1.056) & 0.039 (0.028 - 0.054) & 1.154 (0.944 - 1.426)\\
CR7d\_LR\_c\_all & 28 & 0.004 (-0.003 - 0.009) & 0.044 (0.035 - 0.054) & 0.654 (0.634 - 0.688) & 0.849 (0.657 - 1.127) & 0.042 (0.031 - 0.058) & 1.029 (0.868 - 1.202)\\
\addlinespace
CR7d\_LR\_c\_all & 29 & 0.008 (0.001 - 0.016) & 0.052 (0.04 - 0.067) & 0.697 (0.658 - 0.725) & 1.075 (0.839 - 1.387) & 0.038 (0.024 - 0.057) & 1.037 (0.874 - 1.26)\\
CR7d\_LR\_c\_all & 30 & 0.011 (0.003 - 0.019) & 0.055 (0.04 - 0.072) & 0.717 (0.678 - 0.76) & 1.236 (0.973 - 1.629) & 0.039 (0.025 - 0.056) & 1.06 (0.875 - 1.261)\\
CR7d\_LRCR\_c\_all & 0 & 0.015 (0.013 - 0.018) & 0.046 (0.041 - 0.05) & 0.745 (0.735 - 0.756) & 0.96 (0.897 - 1.023) & 0.002 (0.001 - 0.003) & 1.086 (1.018 - 1.156)\\
CR7d\_LRCR\_c\_all & 1 & 0.02 (0.019 - 0.024) & 0.056 (0.052 - 0.062) & 0.764 (0.75 - 0.772) & 1.031 (0.946 - 1.106) & 0.002 (0.001 - 0.003) & 1.08 (1.011 - 1.145)\\
CR7d\_LRCR\_c\_all & 2 & 0.023 (0.02 - 0.026) & 0.063 (0.057 - 0.068) & 0.762 (0.749 - 0.773) & 1.079 (0.973 - 1.142) & 0.002 (0.001 - 0.004) & 1.046 (0.99 - 1.102)\\
\addlinespace
CR7d\_LRCR\_c\_all & 3 & 0.025 (0.022 - 0.029) & 0.069 (0.063 - 0.076) & 0.759 (0.747 - 0.774) & 1.068 (0.972 - 1.163) & 0.003 (0.002 - 0.005) & 1.031 (0.985 - 1.089)\\
CR7d\_LRCR\_c\_all & 4 & 0.029 (0.025 - 0.032) & 0.077 (0.069 - 0.084) & 0.768 (0.753 - 0.779) & 1.079 (0.986 - 1.183) & 0.003 (0.002 - 0.006) & 0.989 (0.941 - 1.043)\\
CR7d\_LRCR\_c\_all & 5 & 0.026 (0.023 - 0.029) & 0.072 (0.068 - 0.079) & 0.77 (0.76 - 0.784) & 1.079 (0.99 - 1.177) & 0.004 (0.003 - 0.006) & 0.964 (0.903 - 1.024)\\
CR7d\_LRCR\_c\_all & 6 & 0.022 (0.018 - 0.026) & 0.067 (0.061 - 0.075) & 0.748 (0.735 - 0.763) & 0.983 (0.909 - 1.084) & 0.004 (0.003 - 0.008) & 0.97 (0.904 - 1.019)\\
CR7d\_LRCR\_c\_all & 7 & 0.022 (0.017 - 0.025) & 0.068 (0.062 - 0.074) & 0.75 (0.738 - 0.764) & 1.004 (0.908 - 1.109) & 0.006 (0.004 - 0.01) & 0.988 (0.895 - 1.047)\\
\addlinespace
CR7d\_LRCR\_c\_all & 8 & 0.022 (0.019 - 0.026) & 0.072 (0.066 - 0.08) & 0.747 (0.733 - 0.761) & 1.027 (0.924 - 1.117) & 0.007 (0.004 - 0.01) & 0.959 (0.906 - 1.04)\\
CR7d\_LRCR\_c\_all & 9 & 0.028 (0.023 - 0.03) & 0.086 (0.079 - 0.094) & 0.753 (0.74 - 0.764) & 1.112 (1.008 - 1.194) & 0.01 (0.005 - 0.015) & 0.913 (0.858 - 0.983)\\
CR7d\_LRCR\_c\_all & 10 & 0.024 (0.02 - 0.028) & 0.082 (0.073 - 0.092) & 0.741 (0.727 - 0.756) & 1.07 (0.965 - 1.184) & 0.011 (0.006 - 0.017) & 0.943 (0.865 - 1.028)\\
CR7d\_LRCR\_c\_all & 11 & 0.025 (0.02 - 0.029) & 0.085 (0.077 - 0.093) & 0.745 (0.728 - 0.762) & 1.091 (0.961 - 1.255) & 0.012 (0.008 - 0.02) & 0.926 (0.835 - 0.985)\\
CR7d\_LRCR\_c\_all & 12 & 0.024 (0.019 - 0.029) & 0.087 (0.078 - 0.096) & 0.738 (0.714 - 0.752) & 1.06 (0.916 - 1.193) & 0.013 (0.007 - 0.021) & 0.961 (0.867 - 1.035)\\
\addlinespace
CR7d\_LRCR\_c\_all & 13 & 0.023 (0.017 - 0.027) & 0.087 (0.075 - 0.098) & 0.724 (0.698 - 0.744) & 1.058 (0.857 - 1.2) & 0.013 (0.007 - 0.022) & 0.928 (0.84 - 1.013)\\
CR7d\_LRCR\_c\_all & 14 & 0.026 (0.02 - 0.029) & 0.092 (0.082 - 0.108) & 0.731 (0.712 - 0.745) & 1.145 (0.97 - 1.282) & 0.015 (0.008 - 0.027) & 0.978 (0.895 - 1.055)\\
CR7d\_LRCR\_c\_all & 15 & 0.027 (0.022 - 0.03) & 0.104 (0.091 - 0.124) & 0.732 (0.716 - 0.746) & 1.176 (1.004 - 1.272) & 0.02 (0.014 - 0.036) & 0.963 (0.876 - 1.051)\\
CR7d\_LRCR\_c\_all & 16 & 0.026 (0.022 - 0.03) & 0.101 (0.089 - 0.117) & 0.74 (0.717 - 0.757) & 1.192 (1.025 - 1.351) & 0.019 (0.012 - 0.033) & 1.038 (0.942 - 1.099)\\
CR7d\_LRCR\_c\_all & 17 & 0.022 (0.017 - 0.026) & 0.094 (0.079 - 0.114) & 0.717 (0.697 - 0.734) & 1.05 (0.888 - 1.226) & 0.02 (0.012 - 0.029) & 0.888 (0.819 - 0.968)\\
\addlinespace
CR7d\_LRCR\_c\_all & 18 & 0.019 (0.013 - 0.024) & 0.085 (0.074 - 0.101) & 0.705 (0.679 - 0.738) & 0.996 (0.84 - 1.242) & 0.018 (0.011 - 0.029) & 0.983 (0.896 - 1.089)\\
CR7d\_LRCR\_c\_all & 19 & 0.014 (0.009 - 0.02) & 0.067 (0.058 - 0.08) & 0.707 (0.676 - 0.726) & 1.006 (0.824 - 1.163) & 0.019 (0.012 - 0.025) & 1.037 (0.92 - 1.191)\\
CR7d\_LRCR\_c\_all & 20 & 0.013 (0.007 - 0.017) & 0.066 (0.054 - 0.078) & 0.701 (0.672 - 0.728) & 0.975 (0.844 - 1.238) & 0.019 (0.012 - 0.031) & 1.051 (0.899 - 1.234)\\
CR7d\_LRCR\_c\_all & 21 & 0.009 (0.003 - 0.016) & 0.058 (0.048 - 0.072) & 0.686 (0.656 - 0.719) & 0.938 (0.775 - 1.202) & 0.021 (0.012 - 0.033) & 1.15 (0.934 - 1.317)\\
CR7d\_LRCR\_c\_all & 22 & 0.006 (-0.001 - 0.013) & 0.048 (0.04 - 0.069) & 0.687 (0.658 - 0.716) & 0.989 (0.803 - 1.161) & 0.028 (0.019 - 0.039) & 1.196 (0.974 - 1.429)\\
\addlinespace
CR7d\_LRCR\_c\_all & 23 & 0.006 (-0.001 - 0.011) & 0.048 (0.037 - 0.067) & 0.674 (0.648 - 0.706) & 0.933 (0.775 - 1.165) & 0.033 (0.026 - 0.047) & 1.106 (0.929 - 1.332)\\
CR7d\_LRCR\_c\_all & 24 & 0.008 (-0.004 - 0.015) & 0.049 (0.036 - 0.069) & 0.697 (0.669 - 0.729) & 1.13 (0.893 - 1.413) & 0.032 (0.024 - 0.045) & 1.218 (0.991 - 1.48)\\
CR7d\_LRCR\_c\_all & 25 & 0.007 (-0.006 - 0.014) & 0.048 (0.036 - 0.074) & 0.688 (0.657 - 0.721) & 0.987 (0.818 - 1.288) & 0.033 (0.022 - 0.049) & 1.178 (1.017 - 1.485)\\
CR7d\_LRCR\_c\_all & 26 & 0.008 (-0.003 - 0.013) & 0.052 (0.038 - 0.085) & 0.681 (0.643 - 0.714) & 0.996 (0.758 - 1.208) & 0.041 (0.026 - 0.059) & 1.16 (0.954 - 1.473)\\
CR7d\_LRCR\_c\_all & 27 & 0 (-0.007 - 0.007) & 0.04 (0.031 - 0.05) & 0.646 (0.61 - 0.686) & 0.837 (0.608 - 1.05) & 0.039 (0.023 - 0.058) & 1.142 (0.928 - 1.421)\\
\addlinespace
CR7d\_LRCR\_c\_all & 28 & 0.004 (-0.004 - 0.009) & 0.044 (0.034 - 0.056) & 0.658 (0.632 - 0.687) & 0.866 (0.682 - 1.103) & 0.043 (0.032 - 0.058) & 1.001 (0.861 - 1.2)\\
CR7d\_LRCR\_c\_all & 29 & 0.009 (0.001 - 0.015) & 0.051 (0.04 - 0.065) & 0.698 (0.661 - 0.727) & 1.037 (0.814 - 1.403) & 0.038 (0.022 - 0.055) & 1.032 (0.848 - 1.208)\\
CR7d\_LRCR\_c\_all & 30 & 0.011 (0.003 - 0.02) & 0.053 (0.04 - 0.071) & 0.714 (0.675 - 0.757) & 1.216 (0.93 - 1.59) & 0.04 (0.029 - 0.053) & 1.041 (0.858 - 1.237)\\*
\end{longtable}

\subsubsection{Pooled model evaluation}\label{pooled-model-evaluation}

\begin{figure}
\centering
\includegraphics{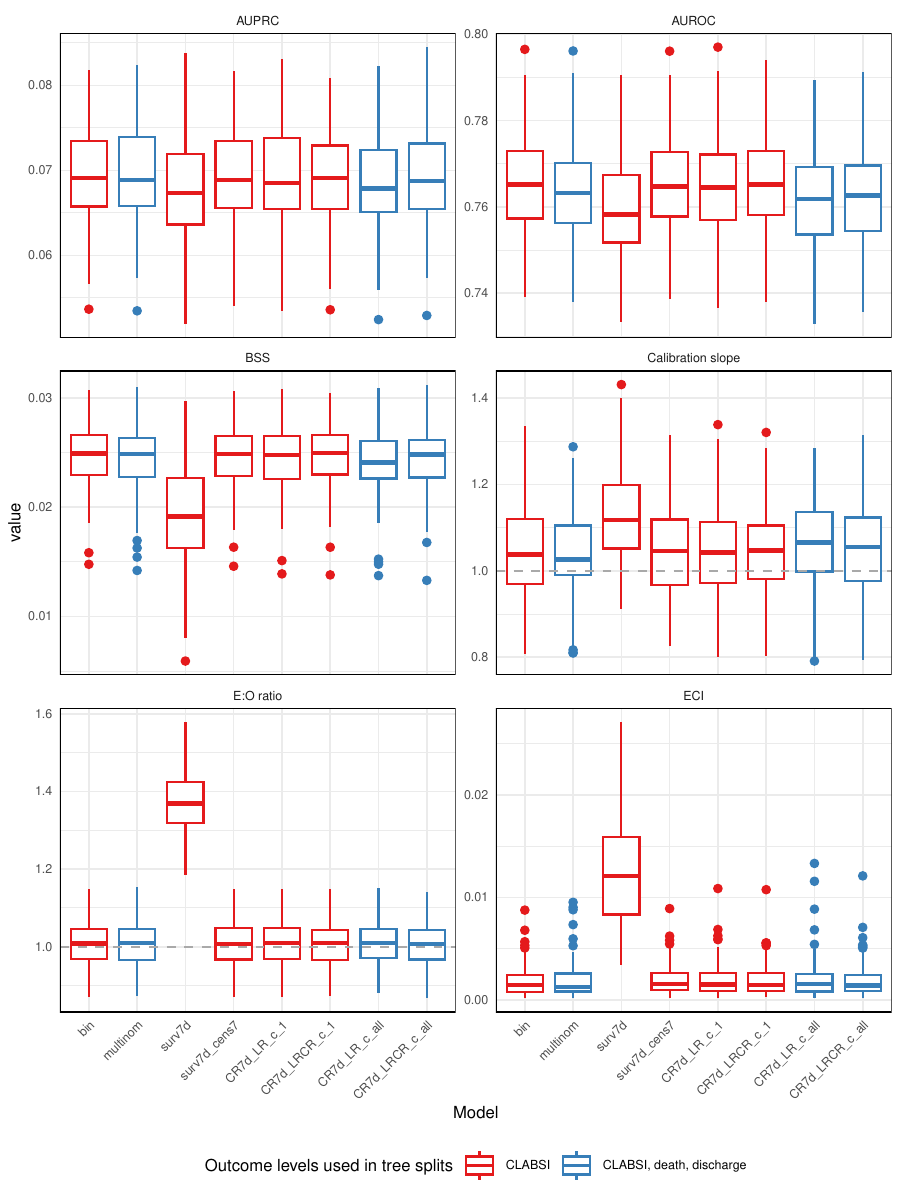}
\caption{\label{fig:pooled-dynamic-cutoff-independent}Prediction performance for dynamic models - pooled metrics}
\end{figure}

In contrast to the time-dependent evaluation, the ``pooled'' evaluation considers all landmarks in the model pooled together (each landmark represents an independent observation). The pooled evaluation metric are presented in Figure \ref{fig:pooled-dynamic-cutoff-independent} The findings are similar to the time-dependent evaluation, with the survival model with competing events censoring at their event time showing poorer performance especially in terms of calibration.

\subsubsection{ROC curves}\label{roc-curves-1}

\begin{figure}
\centering
\includegraphics{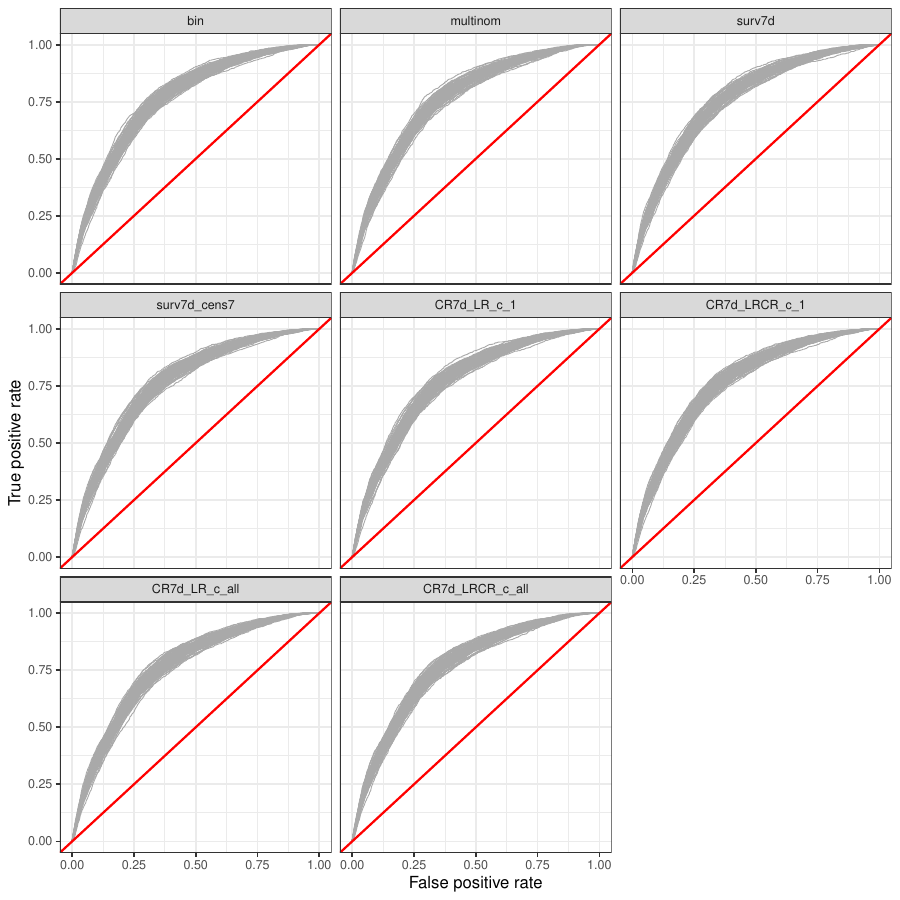}
\caption{\label{fig:ROC-curves-dyn}ROC curves for dynamic models}
\end{figure}

ROC curves are calculated using the ``pooled'' predictions (each landmark represents an independent observation) and presented in Figure \ref{fig:ROC-curves-dyn} for each test set and each model.

\subsubsection{Precision-recall curves}\label{precision-recall-curves-1}

\begin{figure}
\centering
\includegraphics{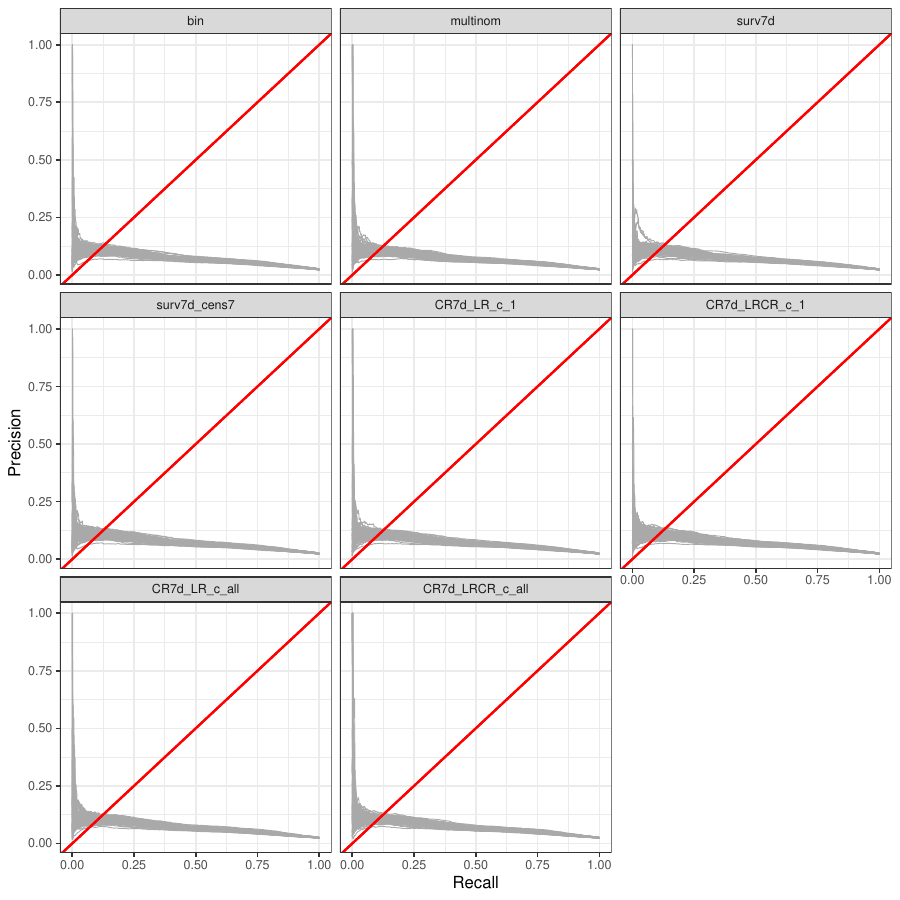}
\caption{\label{fig:PR-curves-dyn}Precision-recall curves for dynamic models}
\end{figure}

Precision-recall curves are calculated using the ``pooled'' predictions (each landmark represents an independent observation) and presented in Figure \ref{fig:PR-curves-dyn} for each test set and each model.

\subsubsection{Calibration curves - deciles}\label{calibration-curves---deciles-1}

\begin{figure}
\centering
\includegraphics{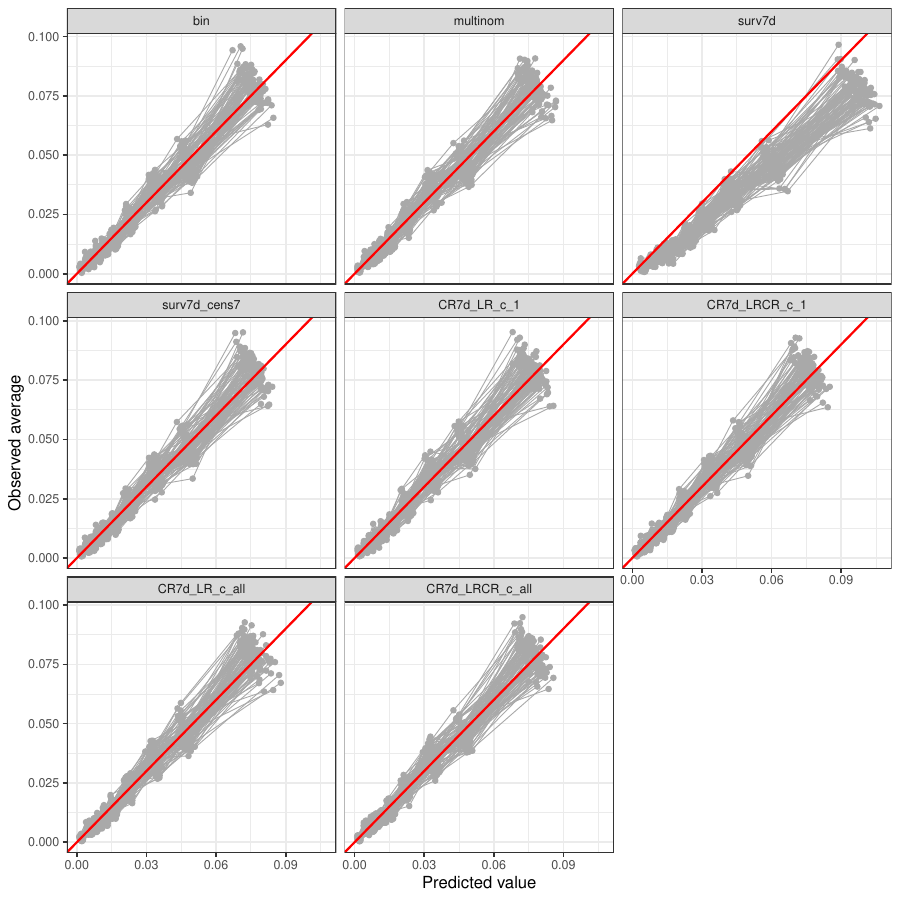}
\caption{\label{fig:calib-deciles-curves-dyn}Calibration curves (deciles) for dynamic models}
\end{figure}

Calibration curves are calculated as for the baseline model, using the ``pooled'' predictions (each landmark represents an independent observation). The survival model with competing risks censored at the time of event shows overestimated predictions. The deciles calibration curves for dynamic models are presented in \ref{fig:calib-deciles-curves-dyn}.

\subsubsection{Calibration curves - splines}\label{calibration-curves---splines-1}

\begin{figure}
\centering
\includegraphics{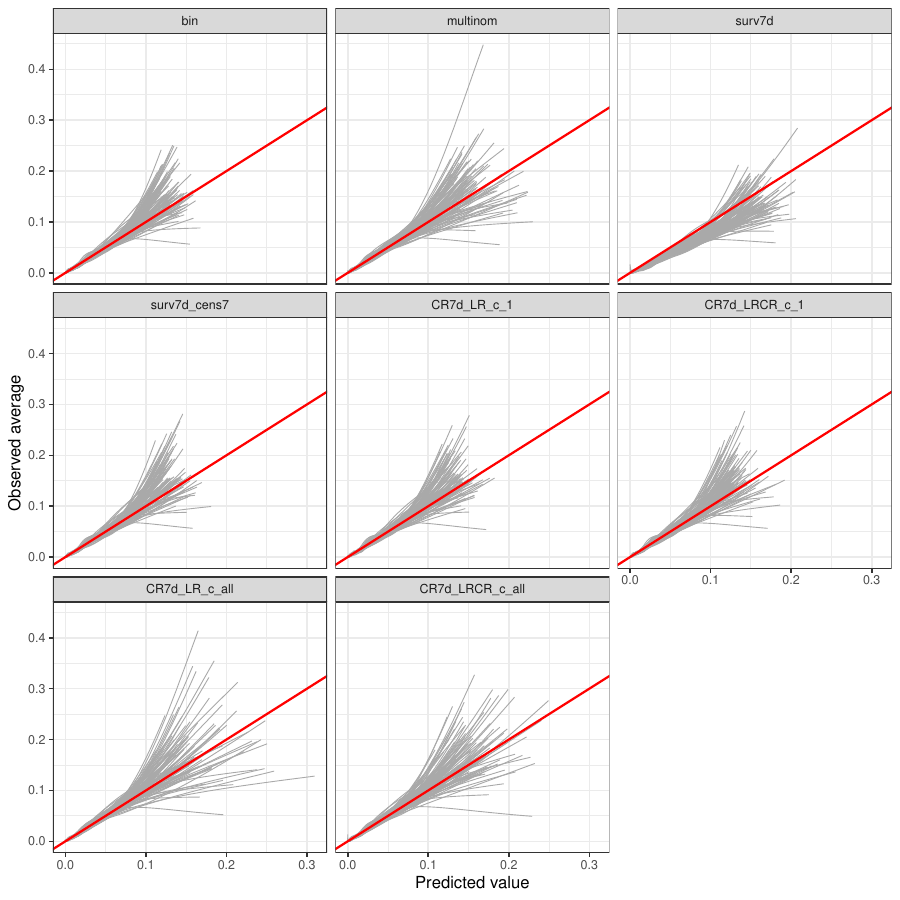}
\caption{\label{fig:calib-splines-curves-dyn}Calibration curves (splines) for dynamic models}
\end{figure}

Calibration curves are calculated as for the baseline model, using the ``pooled'' predictions (each landmark represents an independent observation). The survival model with competing risks censored at the time of event shows overestimated predictions. The calibration curves based on cubic splines are presented in \ref{fig:calib-splines-curves-dyn}.

\subsubsection{Decision curves}\label{decision-curves-1}

Decision curves are calculated as for the baseline model, using the ``pooled'' predictions (each landmark represents an independent observation). The decision curves for dynamic models are presented in Figure \ref{fig:decision-curves-base}.

\begin{figure}
\centering
\includegraphics{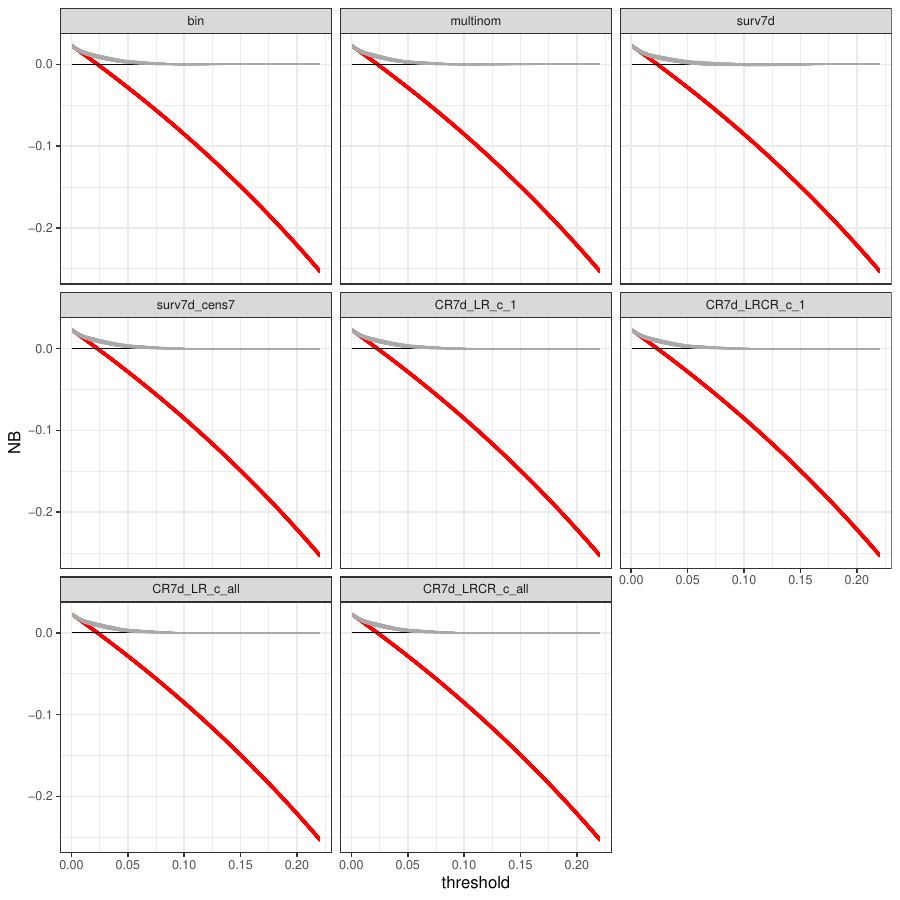}
\caption{\label{fig:decision-curves-dyn}Decision curves for dynamic models}
\end{figure}

\subsubsection{Predictions density curves}\label{predictions-density-curves-1}

\begin{figure}
\centering
\includegraphics{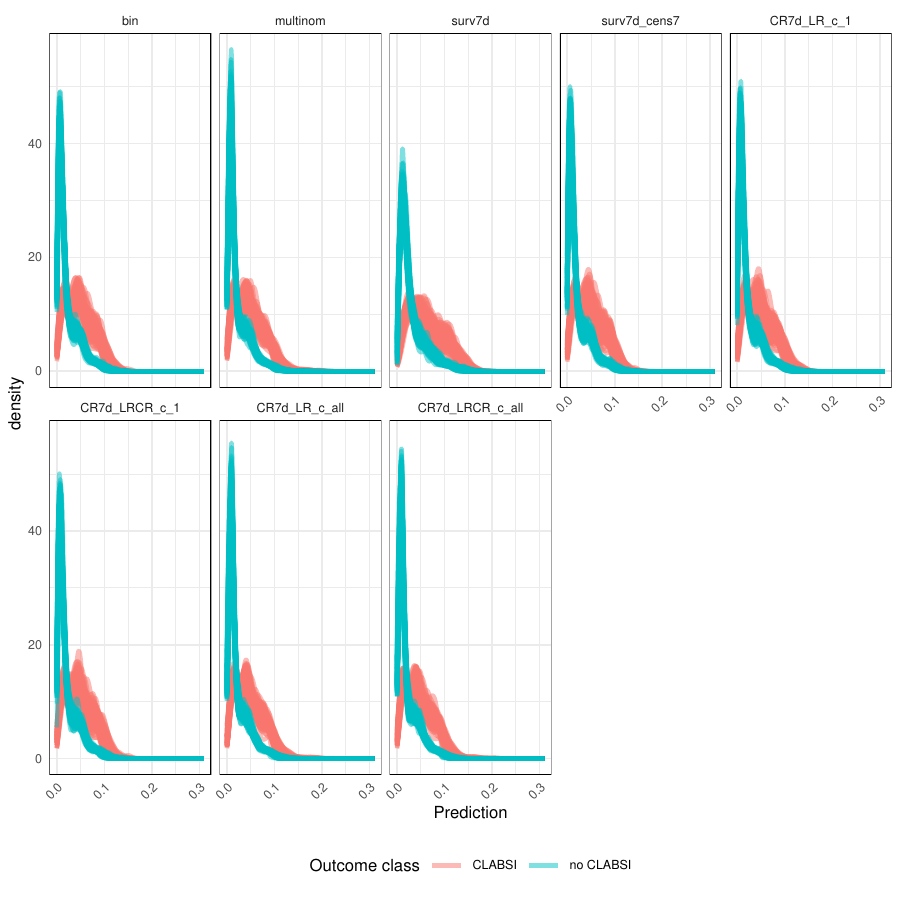}
\caption{\label{fig:pred-density-curves-dyn}Decision curves for dynamic models}
\end{figure}

Prediction density curves for dynamic models using the ``pooled'' predictions (each landmark represents an independent observation) are presented in \ref{fig:pred-density-curves-dyn}. As opposed to the baseline models, there is no notable difference between models.

\subsubsection{Tuned hyperparameters}\label{tuned-hyperparameters-1}

\begin{figure}
\centering
\includegraphics{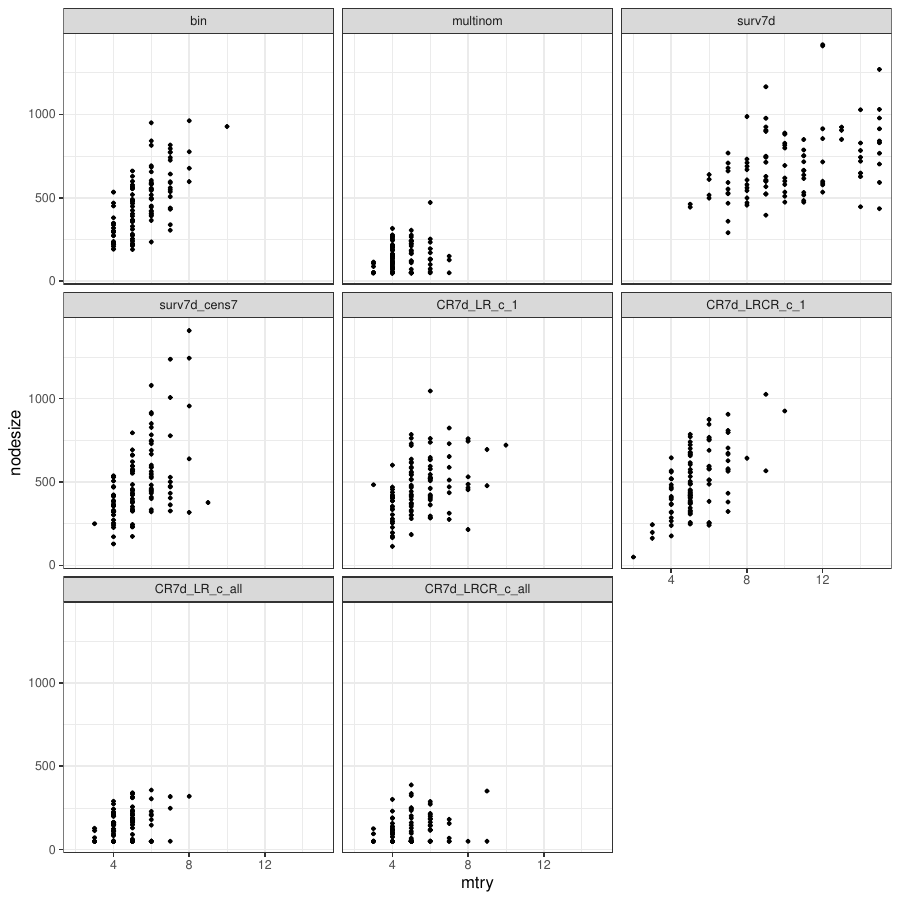}
\caption{\label{fig:hyperparams-dyn-1}Tune hyperparameters (nodesize in function of mtry)}
\end{figure}

\begin{figure}
\centering
\includegraphics{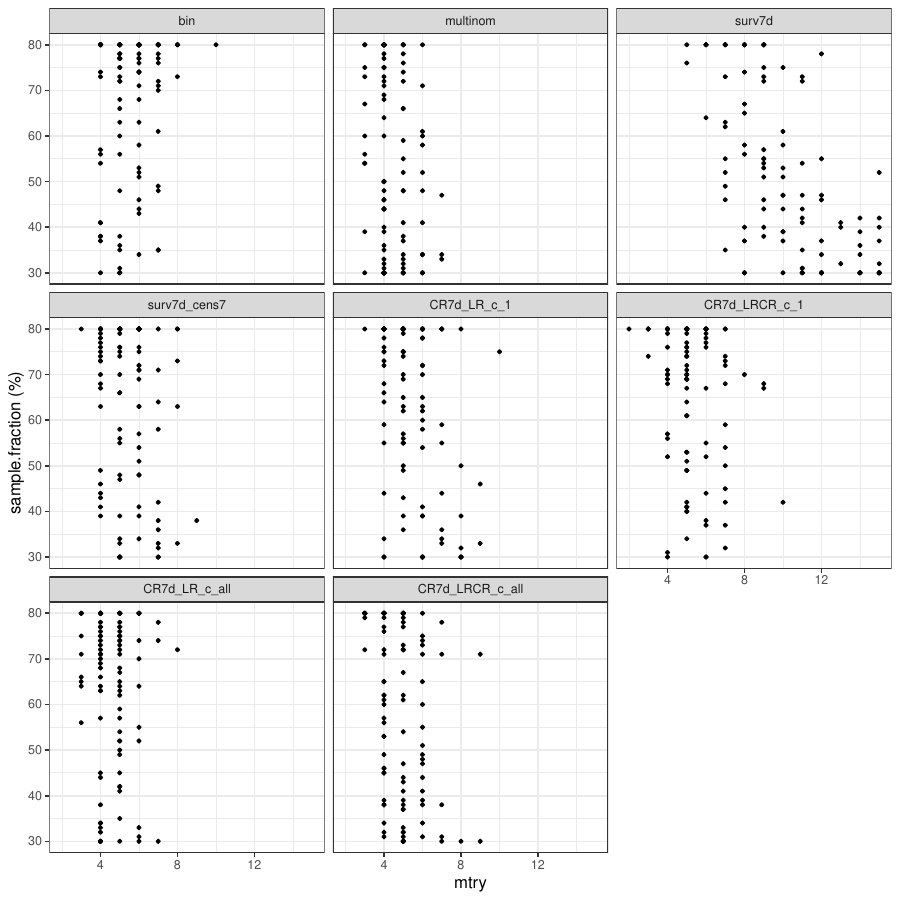}
\caption{\label{fig:hyperparams-dyn-2}Tune hyperparameters (sample.fraction in function of mtry)}
\end{figure}

\begin{figure}
\centering
\includegraphics{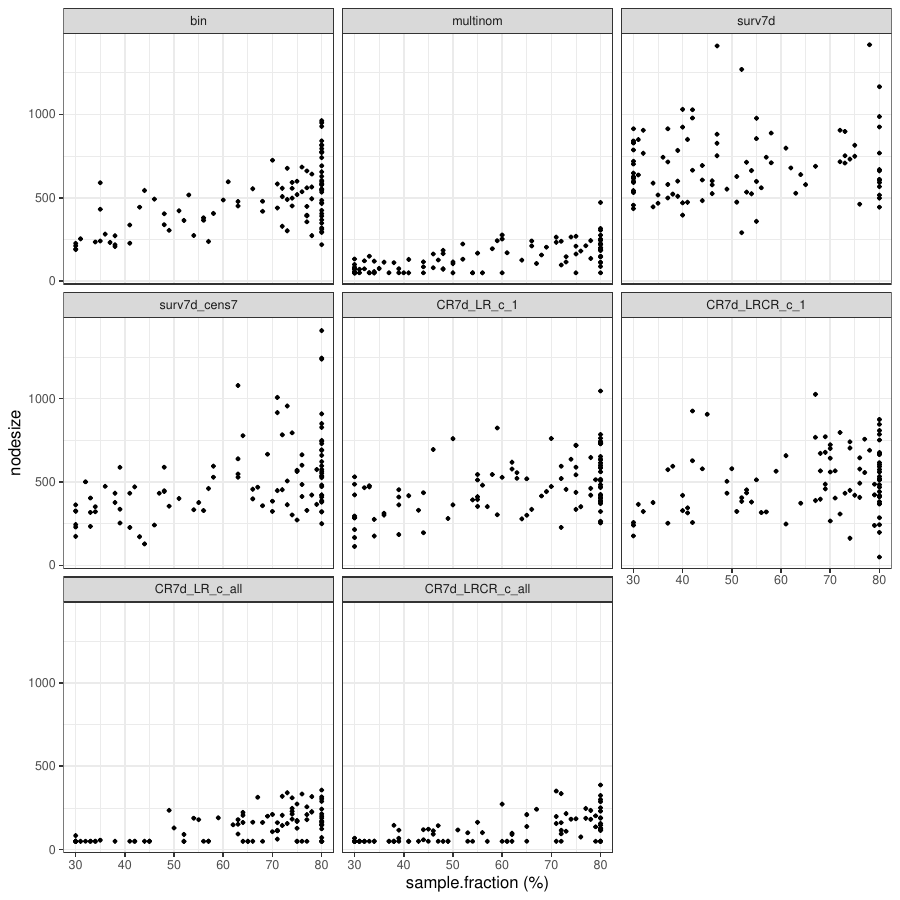}
\caption{\label{fig:hyperparams-dyn-3}Tune hyperparameters (nodesize in function of sample.fraction)}
\end{figure}

The tuned hyperparameters for the dynamic models (mtry, nodesize and sample.fraction) are shown in Figures \ref{fig:hyperparams-dyn-1}, \ref{fig:hyperparams-dyn-2} and \ref{fig:hyperparams-dyn-3}. As for baseline models, models using all events in the outcome definition (multinomial and CR models weighting all causes), have tuned nodesizes lower than the other models. The sample fraction hyperparameter does not show any specific pattern.

\subsubsection{Variable importance}\label{variable-importance-1}

\begin{figure}
\centering
\includegraphics{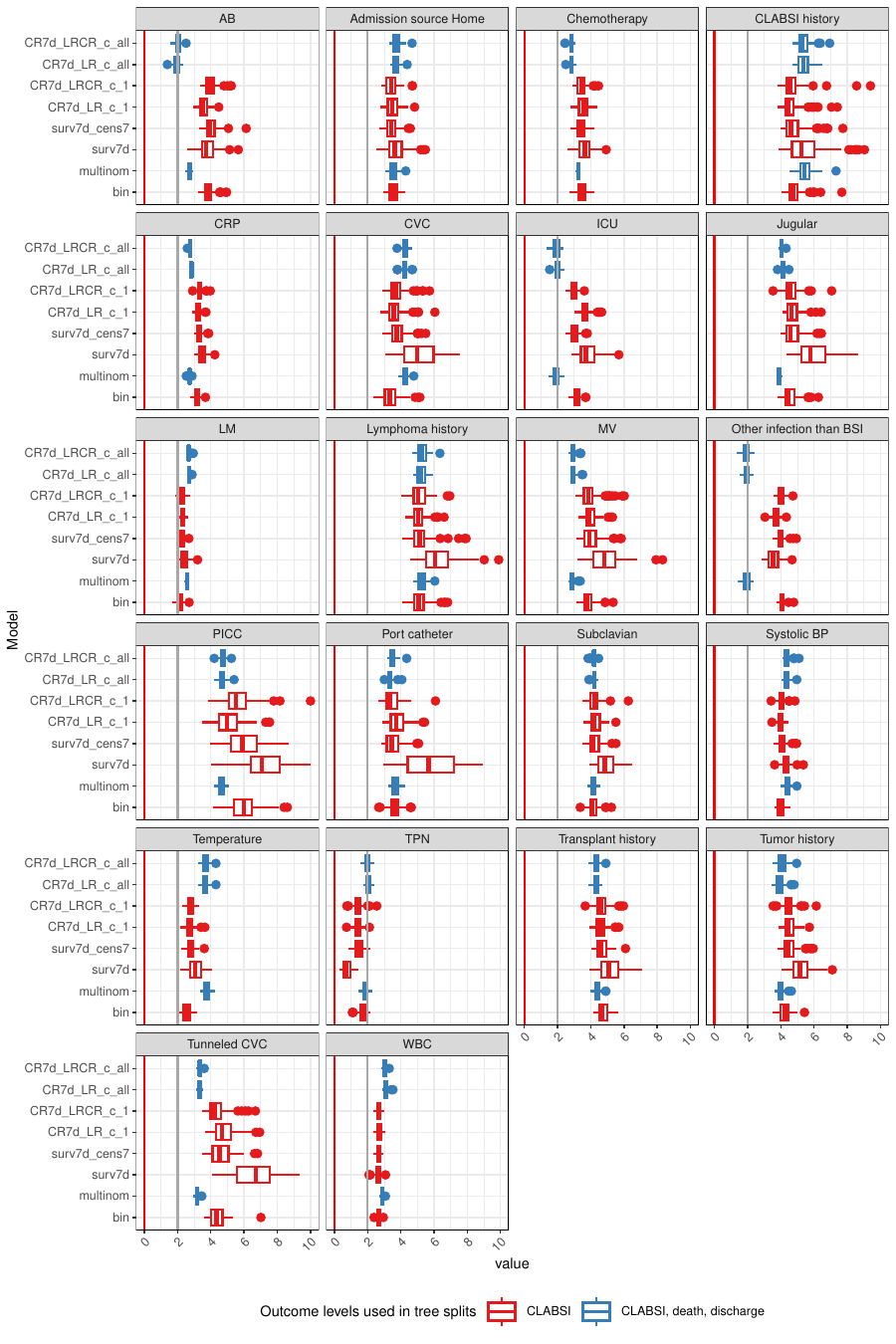}
\caption{\label{fig:var-imp-dyn}ROC curves for dynamic models}
\end{figure}

The minimal depth of the maximal subtree is used as a variable importance metric (Ishwaran et al. 2021), which is the depth in a tree on which the first split is made on a variable \(v\), averaged over all trees in the forest. The lowest possible value is 0 (root node split). The minimal depth of the maximal subtree for dynamic models is presented in Figure \ref{fig:var-imp-dyn}. A guiding line has been added to the plot on value 2, an aleatory choice to guide the focus on most important variables.

Models using multiple levels of the outcome put more weight (splits closer to the root) on variables: ICU, antibacterials, antineoplastic agents (chemotherapy) and CRP, ``other infection than BSI'' and tunneled catheters. The differences in the minimal depth for variables TPN and port-catheter are less notable than for baseline models.

\subsection{Supplementary material 9 - Timings table}\label{supplementary-material-9---timings-table}

The runtimes for baseline and dynamic models are presented in Table \ref{tab:table-timings-all}.

\begin{table}

\caption{\label{tab:table-timings-all}Runtimes for all models}
\centering
\resizebox{\linewidth}{!}{
\begin{tabular}[t]{lllll}
\toprule
Model & Baseline / Dynamic & Tuning & Build final model & Predict\\
\midrule
bin & Baseline & 209.404 (197.433 - 218.468) & 11.641 (10.403 - 12.321) & 0.346 (0.327 - 0.37)\\
multinom & Baseline & 226.332 (208.266 - 236.412) & 7.073 (6.541 - 7.567) & 0.377 (0.345 - 0.4)\\
surv7d & Baseline & 370.972 (355.951 - 385.155) & 9.201 (8.297 - 9.667) & 0.296 (0.288 - 0.306)\\
surv7d\_cens7 & Baseline & 262.443 (249.593 - 274.153) & 7.129 (6.851 - 7.452) & 0.289 (0.278 - 0.307)\\
surv30d & Baseline & 459.341 (446.094 - 476.892) & 10.512 (9.328 - 11.825) & 0.319 (0.309 - 0.328)\\
\addlinespace
surv30d\_cens7 & Baseline & 378.045 (363.976 - 393.637) & 7.893 (7.504 - 8.279) & 0.323 (0.308 - 0.34)\\
CR7d\_LR\_c\_1 & Baseline & 447.95 (431.07 - 472.655) & 8.851 (8.475 - 9.445) & 0.488 (0.465 - 0.509)\\
CR7d\_LRCR\_c\_1 & Baseline & 468.8 (445.506 - 486.273) & 9.578 (8.996 - 11.015) & 0.466 (0.445 - 0.491)\\
CR7d\_LR\_c\_all & Baseline & 401.085 (384.14 - 426.019) & 9.288 (8.962 - 10.001) & 0.481 (0.474 - 0.49)\\
CR7d\_LRCR\_c\_all & Baseline & 426.47 (408.683 - 460.104) & 9.728 (9.195 - 10.307) & 0.514 (0.497 - 0.527)\\
\addlinespace
CR30d\_LR\_c\_1 & Baseline & 567.148 (545.541 - 590.394) & 10.082 (9.295 - 11.126) & 0.518 (0.493 - 0.532)\\
CR30d\_LRCR\_c\_1 & Baseline & 543.608 (526.061 - 578.87) & 10.339 (9.373 - 11.483) & 0.486 (0.467 - 0.513)\\
CR30d\_LR\_c\_all & Baseline & 539.688 (513.044 - 573.47) & 11.179 (10.126 - 13.222) & 0.526 (0.512 - 0.546)\\
CR30d\_LRCR\_c\_all & Baseline & 552.918 (525.582 - 582.852) & 11.528 (10.614 - 13.562) & 0.528 (0.513 - 0.548)\\
bin & Dynamic & 760.216 (736.662 - 786.84) & 209.655 (205.827 - 213.852) & 1.884 (1.778 - 2.019)\\
\addlinespace
multinom & Dynamic & 843.682 (815.876 - 866.967) & 214.542 (209.687 - 219.267) & 2.326 (2.178 - 2.511)\\
surv7d & Dynamic & 935.971 (896.896 - 970.382) & 207.88 (203.285 - 213.398) & 2.25 (2.025 - 2.441)\\
surv7d\_cens7 & Dynamic & 776.992 (747.751 - 804.212) & 205.993 (201.512 - 209.779) & 2.114 (1.768 - 2.278)\\
CR7d\_LR\_c\_1 & Dynamic & 1040.23 (998.582 - 1091.098) & 201.921 (197.449 - 208.597) & 1.969 (1.896 - 2.09)\\
CR7d\_LRCR\_c\_1 & Dynamic & 1045.289 (1004.061 - 1095.314) & 202.667 (197.413 - 206.297) & 2.009 (1.939 - 2.1)\\
\addlinespace
CR7d\_LR\_c\_all & Dynamic & 1039.008 (996.17 - 1091.213) & 207.562 (202.457 - 211.716) & 2.359 (2.161 - 2.767)\\
CR7d\_LRCR\_c\_all & Dynamic & 1057.966 (1004.95 - 1097.65) & 205.315 (201.36 - 211.757) & 2.446 (2.198 - 2.723)\\
\bottomrule
\end{tabular}}
\end{table}

\subsection{Supplementary material 10 - Baseline models on test/train split 001}\label{supplementary-material-10---baseline-models-on-testtrain-split-001}

\begin{figure}
\centering
\includegraphics{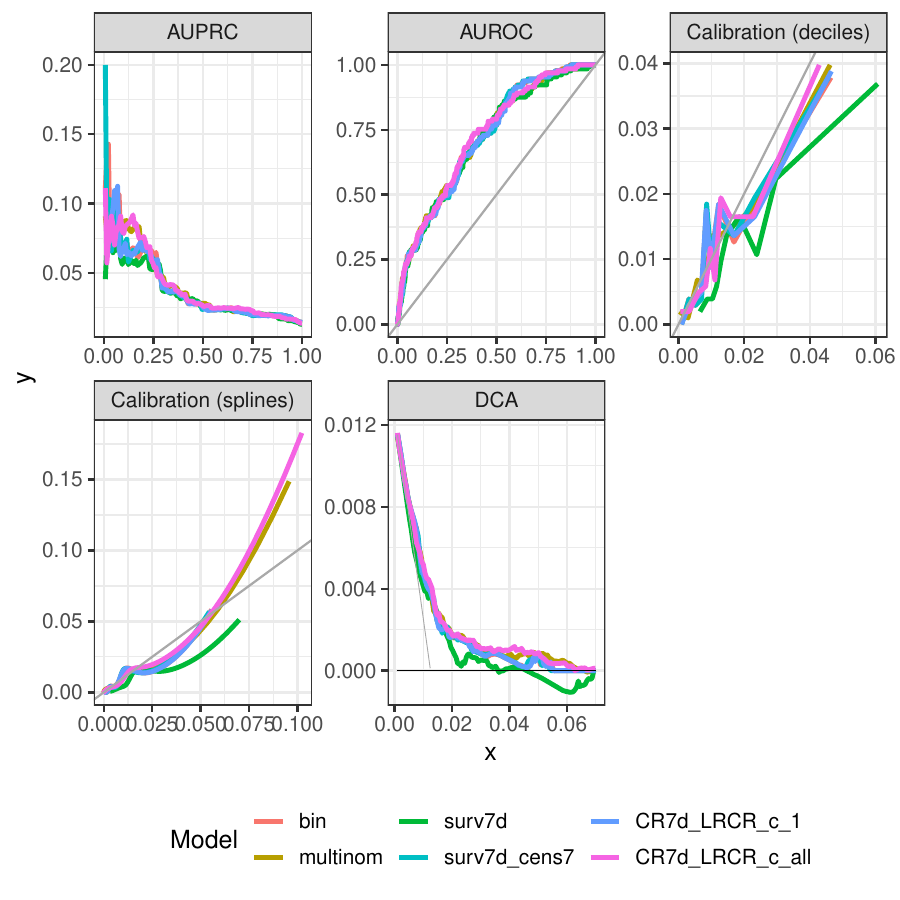}
\caption{\label{fig:test001-curves}All curves. The x and y axis are: Recall and Precision for AUPRC; 1 - Specificity and Sensitivity for AUROC; Predicted probabilities and Observed probabilities for Calibration (deciles) and Calibration (splines); cutoff and Net benefit for DCA}
\end{figure}

\begin{figure}
\centering
\includegraphics{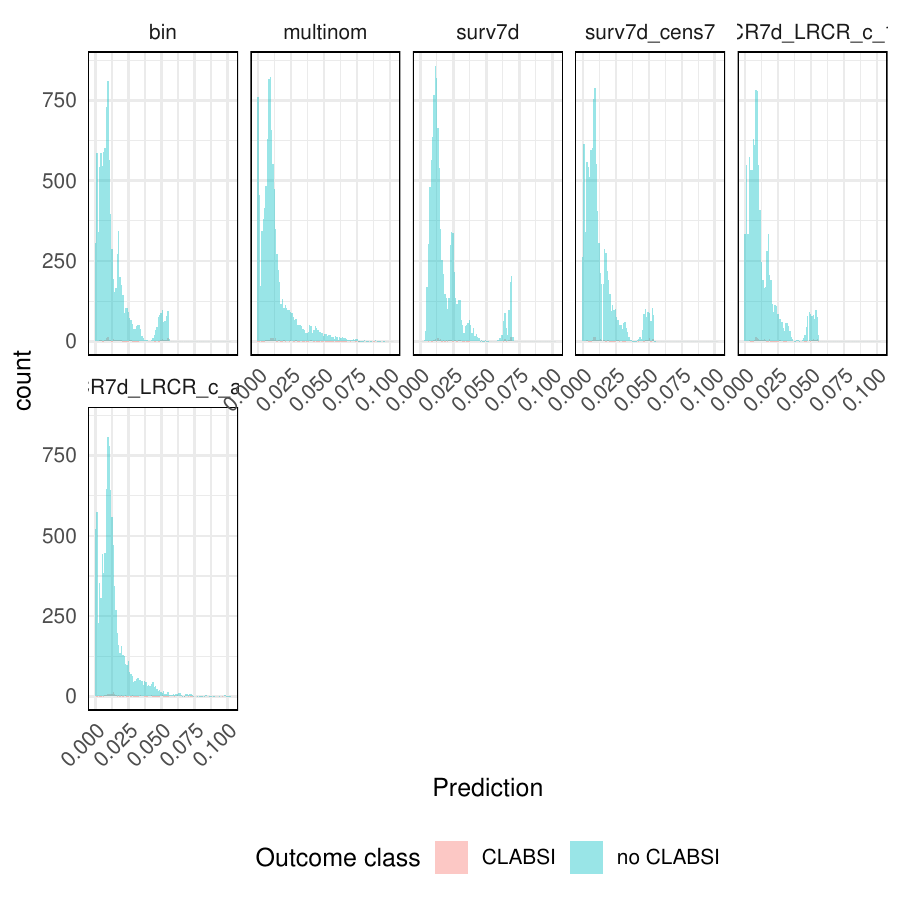}
\caption{\label{fig:test001-pred-hist}Prediction histogram}
\end{figure}

\begin{figure}
\centering
\includegraphics{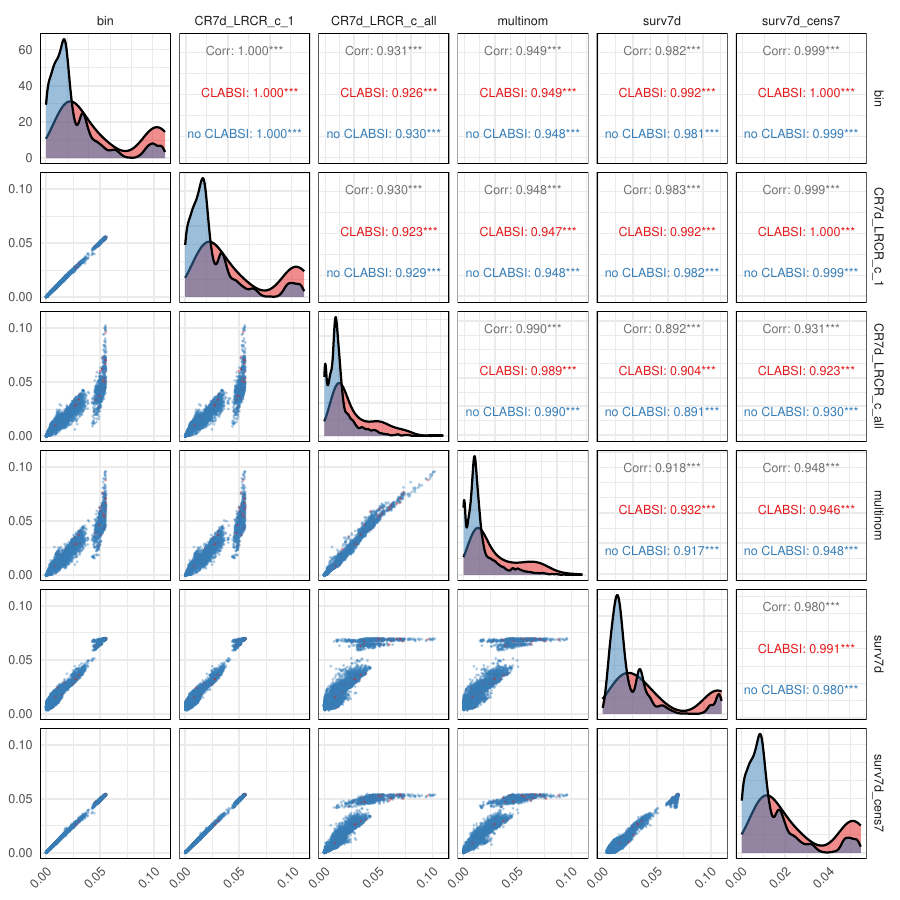}
\caption{\label{fig:test001-pred-plot}Comparison of predictions by model. The lower plots under the diagonal show the predictions of two models plotted against each other. The diagonal contains the the prediction density curves by class (CLABSI vs.~no CLABSI). The upper plots above the diagonal show the correlation (Pearson correlation coefficient) for the predictions of the two models, as well as the correlation within in each class.}
\end{figure}

Additional results on the baseline test set 001 (corresponding to the first data split in the 100 train/test splits) using the ``pooled'' predictions (each landmark represents an independent observation) are presented in Figures \ref{fig:test001-curves}, \ref{fig:test001-pred-hist} and \ref{fig:test001-pred-plot}. To facilitate better visualization, survival and competing risk models with administrative censoring at day 30 (surv30d, surv30d\_cens7, CR30d\_LRCR\_c\_1, CR30d\_LR\_c\_1, CR30d\_LRCR\_c\_all, CR30d\_LR\_c\_all) and competing risk models with administrative censoring at day 7 and logrank split statistic (CR7d\_LR\_c\_1, CR7d\_LR\_c\_all) have been excluded.

\subsection{Supplementary material 11 - Dynamic models on test/train split 001}\label{supplementary-material-11---dynamic-models-on-testtrain-split-001}

Additional results on the dynamic test set 001 (corresponding to the first data split in the 100 train/test splits) using the ``pooled'' predictions (each landmark represents an independent observation) are presented in Figures \ref{fig:test001-curves}, \ref{fig:test001-pred-hist} and \ref{fig:test001-pred-plot}. To facilitate better visualization, competing risk models using logrank split statistic (CR7d\_LR\_c\_1, CR7d\_LR\_c\_all) have been excluded.

\begin{figure}
\centering
\includegraphics{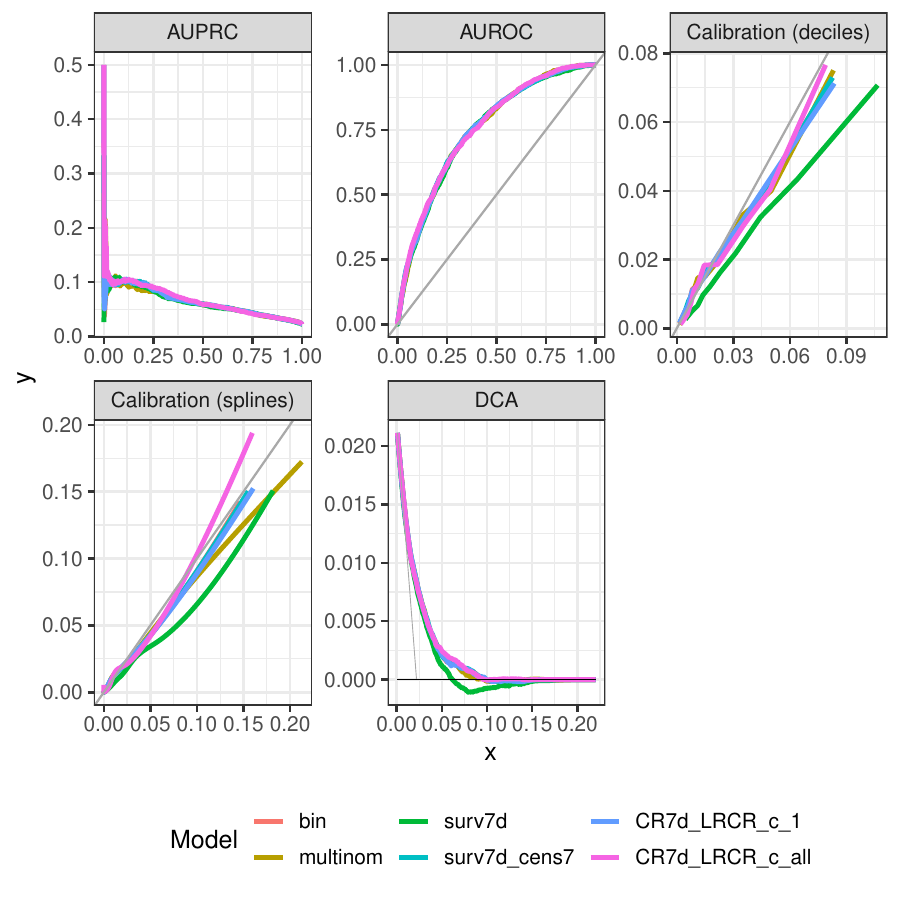}
\caption{\label{fig:test001-curves-dyn}All curves. The x and y axis are: Recall and Precision for AUPRC; 1 - Specificity and Sensitivity for AUROC; Predicted probabilities and Observed probabilities for Calibration (deciles) and Calibration (splines); cutoff and Net benefit for DCA}
\end{figure}

\begin{figure}
\centering
\includegraphics{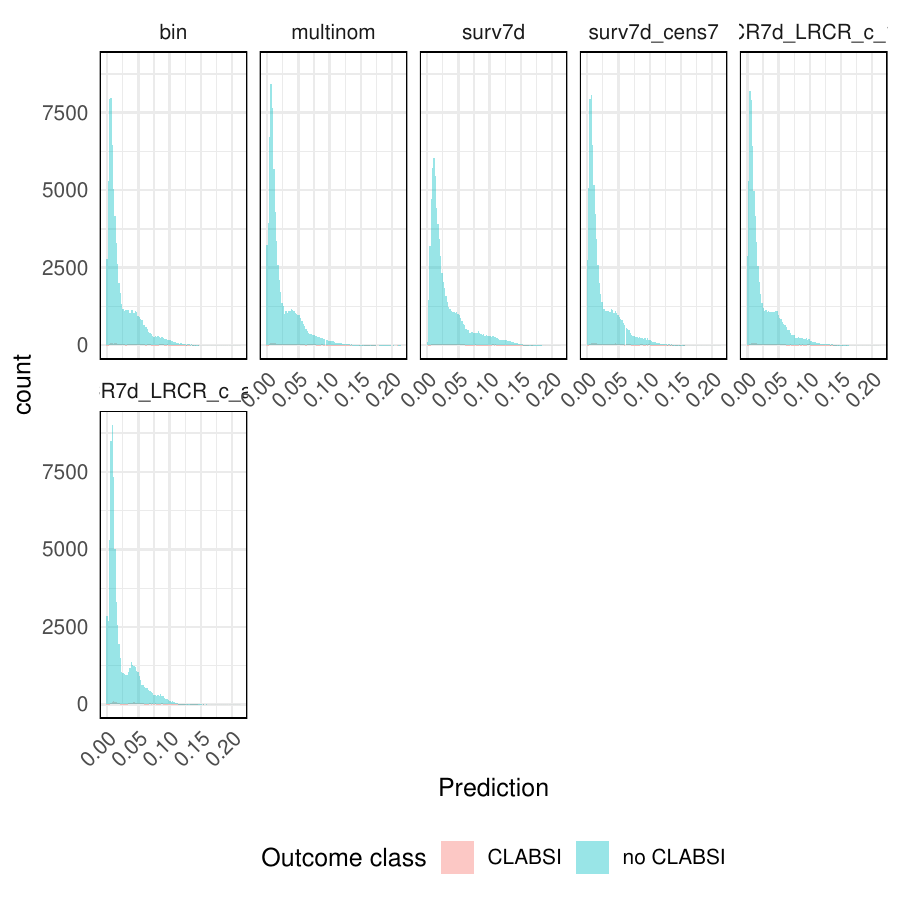}
\caption{\label{fig:test001-pred-hist-dyn}Prediction histogram}
\end{figure}

\begin{figure}
\centering
\includegraphics{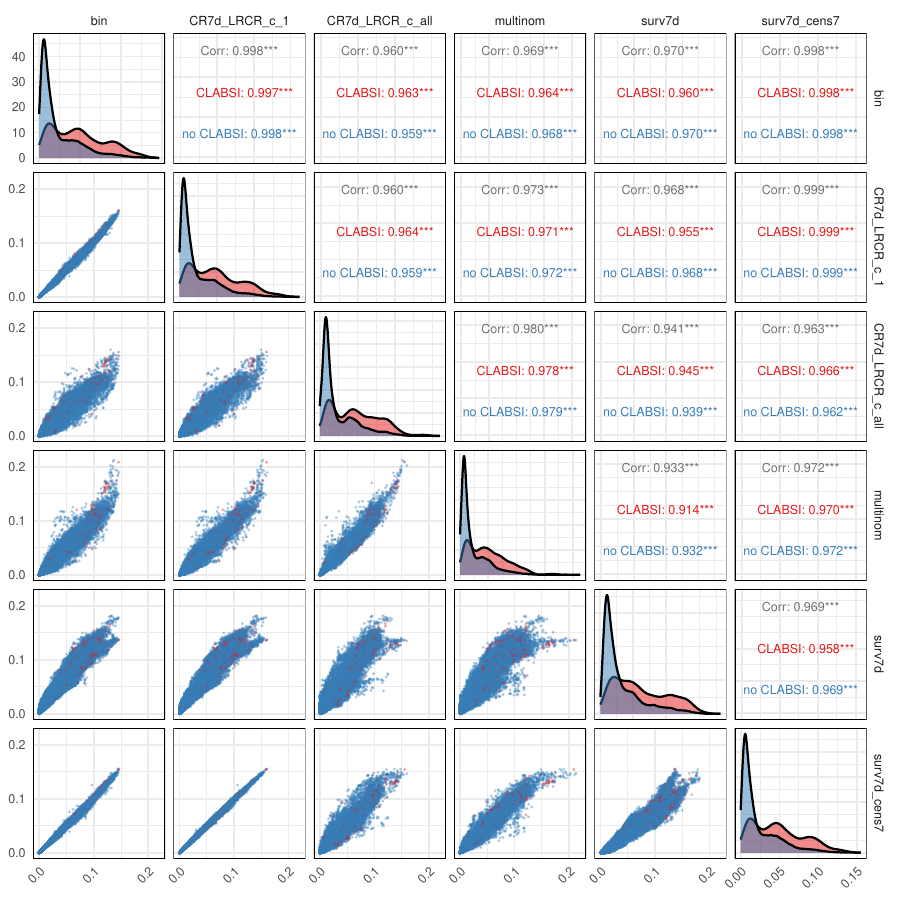}
\caption{\label{fig:test001-pred-plot-dyn}Comparison of predictions by model. The lower plots under the diagonal show the predictions of two models plotted against each other. The diagonal contains the the prediction density curves by class (CLABSI vs.~no CLABSI). The upper plots above the diagonal show the correlation (Pearson correlation coefficient) for the predictions of the two models, as well as the correlation within in each class.}
\end{figure}

\bibliographystyle{unsrt}
\bibliography{references.bib}

\begin{CSLReferences}{1}{0}
\bibitem[\citeproctext]{ref-missForestPredictManual}
Albu, Elena. 2023. \emph{missForestPredict: Missing Value Imputation Using Random Forest for Prediction Settings}. \url{https://github.com/sibipx/missForestPredict}.

\bibitem[\citeproctext]{ref-austin2016introduction}
Austin, Peter C, Douglas S Lee, and Jason P Fine. 2016. {``Introduction to the Analysis of Survival Data in the Presence of Competing Risks.''} \emph{Circulation} 133 (6): 601--9.

\bibitem[\citeproctext]{ref-bischl2017mlrmbo}
Bischl, Bernd, Jakob Richter, Jakob Bossek, Daniel Horn, Janek Thomas, and Michel Lang. 2017. {``mlrMBO: A Modular Framework for Model-Based Optimization of Expensive Black-Box Functions.''} \emph{arXiv Preprint arXiv:1703.03373}.

\bibitem[\citeproctext]{ref-breiman2001random}
Breiman, Leo. 2001. {``Random Forests.''} \emph{Machine Learning} 45: 5--32.

\bibitem[\citeproctext]{ref-deng2022evaluating}
Deng, Hong-Fei, Ming-Wei Sun, Yu Wang, Jun Zeng, Ting Yuan, Ting Li, Di-Huan Li, et al. 2022. {``Evaluating Machine Learning Models for Sepsis Prediction: A Systematic Review of Methodologies.''} \emph{Iscience} 25 (1).

\bibitem[\citeproctext]{ref-sciensano2019CLABSI}
Duysburgh, E. 2019. {``Surveillance Bloedstroom Infecties in Belgische Ziekenhuizen - Protocol 2019.''} \emph{Brussel, België: Sciensano}. \url{https://www.sciensano.be/sites/default/files/bsi_surv_protocol_nl_april2019.pdf}.

\bibitem[\citeproctext]{ref-fine1999proportional}
Fine, Jason P, and Robert J Gray. 1999. {``A Proportional Hazards Model for the Subdistribution of a Competing Risk.''} \emph{Journal of the American Statistical Association} 94 (446): 496--509.

\bibitem[\citeproctext]{ref-fleuren2020machine}
Fleuren, Lucas M, Thomas LT Klausch, Charlotte L Zwager, Linda J Schoonmade, Tingjie Guo, Luca F Roggeveen, Eleonora L Swart, et al. 2020. {``Machine Learning for the Prediction of Sepsis: A Systematic Review and Meta-Analysis of Diagnostic Test Accuracy.''} \emph{Intensive Care Medicine} 46: 383--400.

\bibitem[\citeproctext]{ref-frondelius2023early}
Frondelius, Tuomas, Irina Atkova, Jouko Miettunen, Jordi Rello, Gillian Vesty, Han Shi Jocelyn Chew, and Miia Jansson. 2023. {``Early Prediction of Ventilator-Associated Pneumonia with Machine Learning Models: A Systematic Review and Meta-Analysis of Prediction Model Performance.''} \emph{European Journal of Internal Medicine}.

\bibitem[\citeproctext]{ref-gao2024comparison}
Gao, Shan, Elena Albu, Hein Putter, Pieter Stijnen, Frank Rademakers, Veerle Cossey, Yves Debaveye, Christel Janssens, Ben Van Calster, and Laure Wynants. 2024. {``A Comparison of Regression Models for Static and Dynamic Prediction of a Prognostic Outcome During Admission in Electronic Health Care Records.''} \emph{arXiv Preprint arXiv:2405.01986}.

\bibitem[\citeproctext]{ref-gao2023systematic}
Gao, Shan, Elena Albu, Krizia Tuand, Veerle Cossey, Frank Rademakers, Ben Van Calster, and Laure Wynants. 2023. {``Systematic Review Finds Risk of Bias and Applicability Concerns for Models Predicting Central Line-Associated Bloodstream Infection.''} \emph{Journal of Clinical Epidemiology} 161: 127--39.

\bibitem[\citeproctext]{ref-goldstein2017opportunities}
Goldstein, Benjamin A, Ann Marie Navar, Michael J Pencina, and John PA Ioannidis. 2017. {``Opportunities and Challenges in Developing Risk Prediction Models with Electronic Health Records Data: A Systematic Review.''} \emph{Journal of the American Medical Informatics Association: JAMIA} 24 (1): 198.

\bibitem[\citeproctext]{ref-hodgson2017systematic}
Hodgson, Luke Eliot, Alexander Sarnowski, Paul J Roderick, Borislav D Dimitrov, Richard M Venn, and Lui G Forni. 2017. {``Systematic Review of Prognostic Prediction Models for Acute Kidney Injury (AKI) in General Hospital Populations.''} \emph{BMJ Open} 7 (9): e016591.

\bibitem[\citeproctext]{ref-van2011dynamic}
Houwelingen, Hans van, and Hein Putter. 2011. \emph{Dynamic Prediction in Clinical Survival Analysis}. CRC Press.

\bibitem[\citeproctext]{ref-ishwaran2021randomforestsrc}
Ishwaran, Hemant, Xi Chen, Andy J Minn, Min Lu, Michael S Lauer, and Udaya B Kogalur. 2021. {``randomForestSRC: Minimal Depth Vignette.''} \url{https://www.randomforestsrc.org/articles/minidep.html}.

\bibitem[\citeproctext]{ref-ishwaran2014random}
Ishwaran, Hemant, Thomas A Gerds, Udaya B Kogalur, Richard D Moore, Stephen J Gange, and Bryan M Lau. 2014. {``Random Survival Forests for Competing Risks.''} \emph{Biostatistics} 15 (4): 757--73.

\bibitem[\citeproctext]{ref-ishwaran2008random}
Ishwaran, Hemant, Udaya B Kogalur, Eugene H Blackstone, and Michael S Lauer. 2008. {``Random Survival Forests.''}

\bibitem[\citeproctext]{ref-ishwaran2023package}
Ishwaran, Hemant, Udaya B Kogalur, and Maintainer Udaya B Kogalur. 2023. {``Package {`randomForestSRC'}.''} \emph{Breast} 6 (1): 854.

\bibitem[\citeproctext]{ref-ishwaran2021randomforestsrcSPEED}
Ishwaran, Hemant, Min Lu, and Udaya B Kogalur. 2021. {``randomForestSRC: Speedup Random Forest Analyses Vignette.''}

\bibitem[\citeproctext]{ref-ledger2023multiclass}
Ledger, Ashleigh, Jolien Ceusters, Lil Valentin, Antonia Testa, Caroline Van Holsbeke, Dorella Franchi, Tom Bourne, Wouter Froyman, Dirk Timmerman, and Ben Van Calster. 2023. {``Multiclass Risk Models for Ovarian Malignancy: An Illustration of Prediction Uncertainty Due to the Choice of Algorithm.''} \emph{BMC Medical Research Methodology} 23 (1): 276.

\bibitem[\citeproctext]{ref-li2020consistency}
Li, Yan, Matthew Sperrin, Darren M Ashcroft, and Tjeerd Pieter Van Staa. 2020. {``Consistency of Variety of Machine Learning and Statistical Models in Predicting Clinical Risks of Individual Patients: Longitudinal Cohort Study Using Cardiovascular Disease as Exemplar.''} \emph{Bmj} 371.

\bibitem[\citeproctext]{ref-moor2021early}
Moor, Michael, Bastian Rieck, Max Horn, Catherine R Jutzeler, and Karsten Borgwardt. 2021. {``Early Prediction of Sepsis in the ICU Using Machine Learning: A Systematic Review.''} \emph{Frontiers in Medicine} 8: 607952.

\bibitem[\citeproctext]{ref-pate2019uncertainty}
Pate, Alexander, Richard Emsley, Darren M Ashcroft, Benjamin Brown, and Tjeerd Van Staa. 2019. {``The Uncertainty with Using Risk Prediction Models for Individual Decision Making: An Exemplar Cohort Study Examining the Prediction of Cardiovascular Disease in English Primary Care.''} \emph{BMC Medicine} 17: 1--16.

\bibitem[\citeproctext]{ref-probst2019hyperparameters}
Probst, Philipp, Marvin N Wright, and Anne-Laure Boulesteix. 2019. {``Hyperparameters and Tuning Strategies for Random Forest.''} \emph{Wiley Interdisciplinary Reviews: Data Mining and Knowledge Discovery} 9 (3): e1301.

\bibitem[\citeproctext]{ref-roustant2012dicekriging}
Roustant, Olivier, David Ginsbourger, and Yves Deville. 2012. {``DiceKriging, DiceOptim: Two r Packages for the Analysis of Computer Experiments by Kriging-Based Metamodeling and Optimization.''} \emph{Journal of Statistical Software} 51: 1--55.

\bibitem[\citeproctext]{ref-van2019calibration}
Van Calster, Ben, David J McLernon, Maarten Van Smeden, Laure Wynants, Ewout W Steyerberg, and Topic Group `Evaluating diagnostic tests and prediction models'of the STRATOS initiative Patrick Bossuyt Gary S. Collins Petra Macaskill David J. McLernon Karel GM Moons Ewout W. Steyerberg Ben Van Calster Maarten van Smeden Andrew J. Vickers. 2019. {``Calibration: The Achilles Heel of Predictive Analytics.''} \emph{BMC Medicine} 17: 1--7.

\bibitem[\citeproctext]{ref-van2015spline}
Van Hoorde, Kirsten, Sabine Van Huffel, Dirk Timmerman, Tom Bourne, and Ben Van Calster. 2015. {``A Spline-Based Tool to Assess and Visualize the Calibration of Multiclass Risk Predictions.''} \emph{Journal of Biomedical Informatics} 54: 283--93.

\bibitem[\citeproctext]{ref-vickers10simple}
Vickers, AJ, B Van Calster, and EW Steyerberg. n.d. {``A Simple, Step-by-Step Guide to Interpreting Decision Curve Analysis. Diagn Progn Res. 2019; 3: 18.''} \emph{Journal of Thoracic Disease. All Rights Reserved. Https://Dx. Doi. Org/10.21037/Jtd-21-98 Supplementary}.

\bibitem[\citeproctext]{ref-vickers2006decision}
Vickers, Andrew J, and Elena B Elkin. 2006. {``Decision Curve Analysis: A Novel Method for Evaluating Prediction Models.''} \emph{Medical Decision Making} 26 (6): 565--74.

\bibitem[\citeproctext]{ref-wright2019package}
Wright, Marvin N, Stefan Wager, Philipp Probst, and Maintainer Marvin N Wright. 2019. {``Package {`Ranger'}.''} \emph{Version 0.11} 2.

\bibitem[\citeproctext]{ref-yan2022sepsis}
Yan, Melissa Y, Lise Tuset Gustad, and Øystein Nytrø. 2022. {``Sepsis Prediction, Early Detection, and Identification Using Clinical Text for Machine Learning: A Systematic Review.''} \emph{Journal of the American Medical Informatics Association} 29 (3): 559--75.

\bibitem[\citeproctext]{ref-zhou2021relationship}
Zhou, Qian M, Lu Zhe, Russell J Brooke, Melissa M Hudson, and Yan Yuan. 2021. {``A Relationship Between the Incremental Values of Area Under the ROC Curve and of Area Under the Precision-Recall Curve.''} \emph{Diagnostic and Prognostic Research} 5 (1): 1--15.

\end{CSLReferences}

\end{document}